\documentclass[11pt]{article}

\usepackage[final]{acl}

\usepackage{amsmath}
\usepackage{times}
\usepackage{latexsym}
\usepackage{float}
\usepackage[T1]{fontenc}

\usepackage[utf8]{inputenc}

\usepackage{microtype}

\usepackage{inconsolata}

\usepackage{graphicx}
\usepackage{amssymb}
\usepackage{booktabs}
\usepackage{xcolor}

\usepackage{xspace}
\usepackage[noabbrev,capitalise]{cleveref}

\makeatletter
\newcommand{\ie}{i.e.\@\xspace}
\newcommand{\eg}{e.g.\@\xspace}
\makeatother

\definecolor{mypurple1}{RGB}{143, 94, 255}

\definecolor{revaddgreen}{RGB}{0, 130, 0}
\definecolor{revdelred}{RGB}{200, 0, 0}
\newcommand{\added}[1]{\textcolor{black}{#1}}
\newcommand{\deleted}[1]{\textcolor{black}{#1}}

\newcommand{\myadd}[1]{#1}
\newcommand{\mydel}[1]{}

\newcommand{\cradd}[1]{#1}

\newcommand{\AVg}{IFR\xspace}
\newcommand{\Ve}{IV\xspace}

\usepackage{dsfont}

\title{Same Answer, Different Representations: Hidden instability in VLMs}

\author{
Farooq Ahmad Wani$^1$ \quad 
Alessandro Suglia$^3$ \quad 
Rohit Saxena$^3$ \quad 
Aryo Pradipta Gema$^3$ \\
\textbf{Wai-Chung Kwan}$^3$ \quad 
\textbf{Fazl Barez}$^{4,8}$ \quad 
\textbf{Maria Sofia Bucarelli}$^{2,5,6,7}$ \\ 
\textbf{Fabrizio Silvestri}$^1$ \quad 
\textbf{Pasquale Minervini}$^{3,9}$ \\
\textsuperscript{1}Sapienza University of Rome \qquad
\textsuperscript{2} Université Côte d’Azur\qquad 
\textsuperscript{3}University of Edinburgh \\
\textsuperscript{4}University of Oxford \qquad
\textsuperscript{5}CNRS \qquad
\textsuperscript{6}Inria \qquad
\textsuperscript{7}I3S \qquad
\textsuperscript{8}Martian \qquad \textsuperscript{9}Miniml.AI
\\
\texttt{farooqahmad.wani@uniroma1.it, p.minervini@ed.ac.uk }\\
}

\begin{document}
\maketitle

\begin{abstract}
The robustness of Vision Language Models (VLMs) is commonly assessed through output-level invariance, implicitly assuming that stable predictions reflect stable multimodal processing.
In this work, we argue that this assumption is insufficient.
We introduce a representation-aware and frequency-aware evaluation framework that measures internal embedding drift, spectral sensitivity, and structural smoothness (spatial consistency of vision tokens), alongside standard label-based metrics.
Applying this framework to modern VLMs across the SEEDBench, MMMU, and POPE datasets reveals three distinct failure modes.
\mydel{First, models frequently preserve predicted answers while undergoing substantial internal representation drift; for perturbations such as text overlays, this drift approaches the magnitude of inter-image variability, indicating that representations move to regions typically occupied by unrelated inputs despite unchanged outputs.} \myadd{First, models frequently preserve predicted answers while undergoing substantial internal representation drift; for text overlays this drift can approach inter-image variability in our smallest model, indicating that representations move toward regions typically occupied by unrelated inputs despite unchanged outputs, with the perturbation ordering consistent across families.}
Second, robustness does not improve with scale; larger models achieve higher accuracy but exhibit equal or greater sensitivity, consistent with sharper yet more fragile decision boundaries.
%
%
Third, we find that perturbations affect tasks differently: they harm reasoning when they disrupt how models combine coarse and fine visual cues, but on the hallucination benchmarks, they can reduce false positives by making models generate more conservative answers.
\end{abstract}

\section{Introduction}
Vision Language Models (VLMs) are increasingly deployed in real-world applications that require multimodal perception and reasoning, enabling systems to answer questions, follow instructions, and understand diverse visual inputs~\citep{qwen3vl2024, deitke2024molmo}.
%
As these systems are used more widely, it becomes critical to understand whether they can \emph{reliably maintain their predictions when input images undergo minor transformations that preserve semantic interpretation}.
%

%
Most existing robustness evaluations for VLMs report \emph{output-level stability}, which measures whether a model gives the same answer to slightly perturbed versions of the same input image~\citep{pmlr-v162-fang22a}.
However, recent studies suggest that output-level stability masks hidden instability, particularly in overparameterised models, where correct predictions can persist despite significant differences in the latent representations induced by minor semantically-invariant transformations~\citep{chandhok2025responsewideshut,liu2025gleam}.
In large language models (LLMs), research shows that the hidden representations can differ substantially in the presence of such transformations, even when output tokens remain the same~\citep{wang2025latent,khancal,repstability}---we term this phenomenon \emph{representation drift}.
This representation drift is concerning because it masks decision boundary instability that could lead to failures in downstream applications or brittle behaviour under additional perturbations.
In this work, we aim to answer: \emph{Does output-level robustness in VLMs similarly mask representation drift?}
%

%
VLMs present a unique challenge compared to LLMs: predictions emerge from the interaction of multiple components (\ie, vision encoders, connector layers, and language model backbones), making internal instability more complex and harder to diagnose \citep{Karamcheti}.
%
%
To study this systematically, we examine two failure modes: \emph{reasoning stability}---whether models maintain correct logical deductions under 
minor semantically-invariant perturbations---and \emph{hallucination dynamics}
---whether models correctly verify object existence when images are altered.
We evaluate reasoning stability using SEED-Bench~\citep{li2024seedbench} and MMMU \citep{yue2023mmmu}, and hallucination dynamics using POPE~\citep{li2023evaluating}.
We evaluate multiple VLMs of different scales within the Qwen3-VL, Gemma and LLaVA families on these benchmarks under multiple perturbation types: geometric transformations (\ie, scaling, rotation, translation), spatial modifications (\ie, cropping, padding), and text overlays.
Although these perturbations are defined in the spatial domain, many systematically alter the image's frequency content
after the resizing and interpolation steps required by VLM preprocessing.
%
%

%
We identify several phenomena:
\emph{(i)} A substantial disconnect between output and representation stability; 37.6\% (prediction flipped) of images are affected by at least one perturbation type, and in some cases, the model preserves its predicted answer under perturbation despite a substantial representation drift.
\emph{(ii)} Text overlays are particularly disruptive (19.2\% answer flip rate), while geometric perturbations cause smaller but measurable changes (6--8\%).
\emph{(iii)} \mydel{Most perturbations produce representation drifts comparable in magnitude to inter-image variability, suggesting that such perturbations move internal representations to regions that are typically occupied by different inputs despite unchanged outputs.} \myadd{Text overlays---the most disruptive perturbation---can produce representation drift approaching inter-image variability (most pronounced in the smallest model), moving internal representations toward regions typically occupied by different inputs despite unchanged outputs; the perturbation ordering is consistent across families.}
\emph{(iv)} Model scale does not improve robustness; larger models achieve higher accuracy but exhibit comparable or greater representation drift under perturbations.
\emph{(v)} On reasoning tasks, perturbations lead to more random errors. In contrast, on hallucination tasks, perturbations cause the model to make more conservative predictions, reducing false positives.
%

These observations suggest that assessing VLM robustness requires more than checking output invariance.
Our evaluation framework provides a more complete diagnosis of where and why VLMs fail under perturbations, measuring their robustness beyond output consistency.
\section{Related Work}
\paragraph{Robustness and Internal Consistency in VLMs.}
Robustness in vision-language models has been studied primarily through the lens of task accuracy and output consistency. Extensive work characterised VLM failure modes under adversarial attacks \citep{zhao2023evaluate}, geometric transformations \citep{ishmam2025visual}, and hallucination triggers \citep{Guan_2024_CVPR, li2023evaluating}.
However, these evaluations typically treat the model as a black box, assuming that output stability implies robust processing.
In parallel, research on LLMs challenged this assumption, showing that models can preserve identical outputs while undergoing substantial changes in hidden states and confidence margins~\citep{wang2025latent, nishida-etal-2025-instability}.
Such \emph{latent instability}---where internal representations shift substantially despite unchanged outputs---has been linked to calibration failures and instability in chain-of-thought reasoning.
We bridge these two lines of inquiry: unlike prior VLM benchmarks that focus exclusively on output correctness \citep{yue2023mmmu, zhang2024vlmevalkit}, we analyse how internal representations evolve under perturbations \added{ across multiple context and answer embedding regimes}.
%

%
\paragraph{Spectral Analysis and Structural Drift.}
A large body of work defines robustness through the frequency domain.
%
%
%
Early analyses demonstrated that CNNs and Vision Transformers (ViTs) exhibit a strong reliance on low-frequency features \deleted{ (the slowly varying components that encode coarse shapes, overall colour gradients, and global structure)}, yet can still fail when high-frequency details \deleted{ (\eg, sharp edges, fine textures, and rapid intensity changes)} are perturbed~\citep{yin2019fourier,shao2021vitrobust}.
Furthermore, discretisation steps in ViTs, such as ``patchification'', have been shown to introduce aliasing artefacts that exacerbate spectral sensitivity~\citep{Abraham_2025_CVPR, NEURIPS2021_2b3bf3ee}.
Our work formalises the intuition that ``benign'' natural perturbations (\eg, rotation, scaling) are actually sources of \emph{induced spectral drift}---systematic changes to the frequency-domain structure of the image. 
\deleted{In the frequency domain, images are characterised by both \emph{magnitude} (how much energy exists at each frequency) and \emph{phase} (where that frequency content is spatially located).}
Phase carries most structural information: edges, boundaries, and spatial relationships are encoded primarily in phase rather than magnitude~\citep{oppenheim1981importance}\added{, so rotation and scaling can preserve overall magnitudes while disrupting phase alignment, mislocating visual features even when the image appears semantically unchanged}.
%
%
We extend frequency analysis from pure vision backbones to the full VLM pipeline, demonstrating that robustness failures
are better explained by \emph{cross-frequency misalignment} \added{ across low- and high-frequency bands}
rather than simple low-frequency reliance.
To quantify the structural impact of these shifts, we adopt \emph{Dirichlet energy}, traditionally used to measure spatial smoothness in graph neural networks \citep{NEURIPS2021_b6417f11}, and apply it here to track the structural reorganisation of vision tokens under perturbation.

\paragraph{Evaluation Protocols: Generation vs. Scoring.}
VLM evaluation protocols often rely on free-form text generation, which complicates robustness analysis due to decoding stochasticity and formatting ambiguity.
We adopt a multiple-choice scoring protocol based on the log-likelihood of answer options, which is popular in VLM~\citep{li2024seedbench} and LLM~\citep{eval-harness} evaluation benchmarks.
This approach allows us to track \emph{margin dynamics}, the confidence gap between the correct answer and the top distractor.
%
%
By combining decision-level margins with representation-level metrics, our framework provides a diagnostic view of robustness, distinguishing between cases where the model is robust (\emph{stable internal state}) versus cases where it is merely fortuitous (\emph{drifted state but preserved prediction}).

\section{Experimental Setup}
\label{methodology and experiment}
Our evaluation framework is designed to decouple \emph{decision instability} (output changes) from \emph{representation instability} (representation drift).
We conduct experiments on SEEDBench~\citep{li2024seedbench}, MMMU~\citep{yue2023mmmu}, and POPE~\citep{li2023evaluating}, restricting our scope to multiple-choice and binary decision subsets.
This setting is particularly well-suited for robustness analysis because
\emph{(i)} the semantic task remains fixed across perturbations,
\emph{(ii)} decisions are discrete and directly comparable, and
\emph{(iii)} confidence margins can be measured reliably.
A detailed description of the datasets, models, perturbation examples, and prompt templates is provided in \cref{app:exp_details}.
\subsection{Scoring Protocol and Margin Dynamics}
%
%
We use a log-likelihood-based evaluation protocol that ranks answer options by their conditional probabilities under the model.
More formally, let $x$ denote the visual input.
For SEEDBench and POPE, $x$ consists of a single image, while for MMMU, $x$ represents a sequence of images, denoted as $x = \{I_1, I_2, \dots\}$.
Given a question $q$ and a set of answer options $\mathcal{A}$, we construct a prompt $p(q, \mathcal{A})$ and compute the conditional log-likelihood for each option $o \in \mathcal{A}$:
\begin{equation}
s(o \mid x) = \log P_\theta(o \mid x, p(q,\mathcal{A})),
\end{equation}
where $\theta$ represents the frozen VLM parameters.
The predicted label is $\hat{y}(x) = \arg\max_{o \in \mathcal{A}} s(o \mid x)$.
%
%
This allows us to track \emph{margin dynamics}, \ie the confidence gap between the \mydel{correct answer}\myadd{predicted answer (the model's $\arg\max$ option)} and the top distractor.
Specifically, we define the margin as the log-probability gap between the predicted option $o^*$ and its strongest competitor:
\begin{equation} \label{eq:margin}
\text{margin}(x) = s(o^* \mid x) - \max_{o \neq o^*} { s(o \mid x)}.
\end{equation}
Intuitively, a positive but shrinking margin indicates latent instability even when the output label remains unchanged.
\subsection{Robustness Metrics}
We evaluate robustness under six families of \emph{natural} perturbations $v'$.
The perturbation families include translation (cyclic shifts), padding/cropping, scaling, rotation, and text overlays.
For text overlays, we evaluate three variants with identical geometry (boxes) but different content: semantic directives, \ie, adversarial text (\texttt{TextOverlay}), random strings (\texttt{RandomText}), and empty boxes (\texttt{BoxOverlay}).
Parameter ranges and specific values are detailed in \cref{app:naturalpertubations}.
We quantify stability using two metrics: \emph{Instance Flip Rate} (\AVg) and \emph{Image Vulnerability} (\Ve).
\paragraph{Instance Flip Rate.}
\AVg measures the failure frequency across all perturbation trials.
Let $\mathcal{I}_{v'}$ be the set of all perturbed pairs $(x, x')$ for perturbation type $v'$:
\begin{equation*}
\mathrm{\AVg}_{v'} = \frac{1}{|\mathcal{I}_{v'}|} \sum_{(x,x') \in \mathcal{I}_{v'}} \mathds{1}\{\hat{y}(x') \neq \hat{y}(x)\}.
\end{equation*}
\paragraph{Image Vulnerability.}
Measures the fraction of unique images that are vulnerable to \emph{at least one} instance of perturbation $v'$.
Let $\mathcal{X}$ be the set of unique base samples:
\begin{equation*}
\mathrm{\Ve}_{v'} = \frac{1}{|\mathcal{X}|} \sum_{x \in \mathcal{X}} \mathds{1}\{\exists x' \in v'(x) : \hat{y}(x') \neq \hat{y}(x)\}.
\end{equation*}
\subsection{Representation-Aware Analysis}
%
We employ four diagnostic tools to measure internal drift, namely \emph{Embedding Stability}, \emph{Structural Smoothness} (also referred to as \emph{Dirichlet Energy}), \emph{Perturbation and Control Drift}, and \emph{Drift-to-Prior}.
\added{These signals are designed as \emph{complementary diagnostic indicators}: embedding stability captures global representational displacement, Dirichlet energy captures local structural reorganisation, Cohen's $d$ contextualises drift against inter-image variability, and drift-to-prior isolates language-prior reliance. Hidden instability is revealed when these signals disagree (\eg, an unchanged output combined with large embedding drift); modest individual correlations between them are therefore expected, not analytical limitations.}
\paragraph{Embedding Stability.}
%
%
We measure the cosine distance and $\ell_2$ norm between the base embedding $e(x)$ and perturbed embedding $e(x')$ at five extraction points that vary by prompt type (open-ended vs.\ multiple-choice question (MCQ)) and token position (context vs. answer)---isolating where in the pipeline the visual grounding erodes.
%
%
%
Specifically, we extract hidden states from the final transformer layer (before the output projection) at five positions: \texttt{ctx\_open} and \texttt{ctx\_mcq} capture the last context token before generation begins, under open-ended and MCQ prompts, respectively; \texttt{ans\_open} and \texttt{ans\_mcq} capture the mean-pooled embedding of generated answer tokens under each prompt type; \texttt{ans\_mcq\_free} captures the embedding when the model generates freely but is evaluated against the MCQ options.
This design separates the effects of prompt conditioning from answer generation, revealing whether drift originates in visual-context encoding or in the answer-production stage.
See \cref{app:regimes} for more details.
\paragraph{Structural Smoothness (Dirichlet Energy).}
%
%
%
For coherent visual understanding, adjacent image patches should encode semantically related features; a patch showing part of a dog's ear should have a similar representation to a neighbouring patch showing more of the same ear.
If perturbations fragment this local coherence---causing neighbouring patches to encode unrelated features---downstream reasoning may be disrupted even when global representations remain stable.
We quantify this local structure using Dirichlet energy, a measure of variation between adjacent nodes in a graph~\citep{NIPS2001_f106b7f9}\added{; it is the natural graph-theoretic measure of variation over a grid, widely used in spectral graph theory and manifold learning~\citep{shuman2013emerging,zhu2003semi}}.
Concretely, we treat vision tokens as nodes in a grid graph, where each node corresponds to an image patch.
Let $\mathbf{z}_{i,j} \in \mathbb{R}^d$ be the embedding of the patch at coordinates $(i,j)$, and let $Z = \{\mathbf{z}_{i,j}\}$ denote the full grid of patch embeddings.
The Dirichlet energy $E(Z)$ of $Z$ measures the total variation between adjacent patches:
\begin{equation}
E(Z)= \sum_{(i,j) \sim (i',j')} ||\mathbf{z}_{i,j} - \mathbf{z}_{i',j'}||_2^2.
\end{equation}
We report the energy gap $\Delta E = E(Z{_{\text{pert}}}) - E(Z_{\text{base}})$. A positive $\Delta E$ implies the perturbation has introduced high-frequency structural noise, while a negative $\Delta E$ implies over-smoothing.

\paragraph{Perturbation Drift vs. Control Drift.}
To contextualise perturbation-induced representation drift, we compare it against a control baseline.
For each base image, we compute distances between its embedding across regimes and the embeddings of randomly sampled other images, yielding a control-drift distribution.
We quantify separation using Cohen’s $d$ \citep{cohen1988statistical}\added{~\citep{sawilowsky2009new}}.
Large negative values indicate that perturbation-induced drift is substantially smaller than inter-image variability, while values closer to zero indicate non-local displacement in representation space (detailed in \cref{app:controldrift}).
\paragraph{Drift-to-Prior (POPE).}
To distinguish between visual hallucinations and language bias, we evaluate model predictions on blank images, computing the prior score $S_{\text{prior}} = P(\text{``Yes''} \mid q, \text{blank})$ to measure whether perturbations shift predictions toward this language-only baseline, effectively forcing the model to abandon visual evidence in favour of its base language prior (\cref{sec:exp_pope}).

\section{Results}
\label{sec:embedding_drift}
We now analyse the effect of natural visual perturbations on both model outputs and internal representations.
Our goal is to answer two questions:
\emph{(i)} how often do predictions change under meaning-preserving transformations, and
\emph{(ii)} what happens internally when predictions do \emph{not} change.
Unless stated otherwise, results in this section are reported for a single representative model, \texttt{Qwen3-VL-2B Instruct} and SEEDBench dataset, while the same evaluation protocol is applied consistently across datasets (SEEDBench, MMMU, and POPE) and architectural families; additional results can be found in \cref{app:scaling} and \cref{appendix_POPE}.
%
%

%
\subsection{Label Instability Under Natural Perturbations}
We first report results on SEEDBench, which we use as a primary illustrative benchmark due to its diverse visual reasoning tasks and multiple-choice structure.
Each experiment is conducted over \emph{four independent runs}, each using a disjoint random subset of 3,500 images.
In total, this corresponds to approximately 14,000 evaluated samples, and all reported statistics are averaged across runs.
%

%
The average base accuracy on unperturbed images is $61.7\%$.
\cref{tab:fliprates} reports the instance flip rate (\AVg) and image-level flip probability (\Ve) for each perturbation type.
Even simple geometric perturbations induce non-trivial instability.
For translation and pad/crop, approximately $16$--$17\%$ of images exhibit at least one perturbation instance that flips the predicted answer.
Rotation further increases instability, while text overlays are the most disruptive: they exhibit the highest instance flip rate (\AVg $\approx 19.2\%$) and image-level flip probability (\Ve $\approx 23.9\%$).
When considering the union of all perturbations, over one third of images ($\mathrm{\Ve} \approx 37.6\%$) experience at least one decision flip.
This suggests that robustness failures are not isolated corner cases but arise across a broad range of natural transformations.
%

%
\added{Since these perturbations preserve semantics,} the observed flips reflect sensitivity to representational changes rather than semantic corruption.
%

%
\paragraph{Revisiting Overlays: Semantics vs. Occlusions.}
Decomposing overlays reveals a clear ordering.
\texttt{TextOverlay} (semantic, \eg, ``Answer is X'') is consistently the most disruptive variant, followed by \texttt{RandomText}, while \texttt{BoxOverlay} is comparatively benign.
This indicates that overlay-induced failures cannot be attributed to occlusion alone:
\emph{semantic or instruction-like visual content} substantially amplifies decision instability beyond what masking (BoxOverlay) or edge injection (RandomText) explains.
%
%
\added{The semantic-vs.\ random gap (19.2\% vs.\ 6.4\%) shows the disruption is driven by the language backbone interpreting the overlay \emph{as an instruction}; the random-vs.\ empty gap (6.4\% vs.\ 4.3\%) isolates visual edge injection. We therefore interpret \texttt{TextOverlay} as \emph{pixel-space prompt injection}, not visual misalignment.}
\subsection{Correctness Transitions: Perturbations Can Fix and Break Predictions}
To understand whether perturbations systematically degrade performance or move predictions bidirectionally, 
we distinguish four transition types: R$\to$W (correct $\to$ wrong), W$\to$R (wrong $\to$ correct), R$\to$R (stays correct), and W$\to$W (stays wrong).
\cref{tab:gt_transitions} reports transition counts for each perturbation family.
All perturbations exhibit both R$\to$W and W$\to$R transitions, showing they induce movement across decision boundaries in both directions rather than acting as uniform noise.
\mydel{For \texttt{TextOverlay}, W$\to$R transitions (881) are comparable to R$\to$W (685), indicating overlays can correct wrong predictions despite increasing instability.} \myadd{The \emph{geometric} perturbations carry no textual instruction, so their bidirectionality (Translation 694/618; Rotation 390/262) directly evidences decision-boundary instability. \texttt{TextOverlay} differs: its overlays exclude the model's own prediction, so the W$\to$R$\,>\,$R$\to$W asymmetry (881 vs.\ 685) is by design and reflects fragility-gated instruction compliance, not uniform degradation (Appendix~\ref{app:naturalpertubations}).} \cradd{The model in fact resists most overlays: base-correct images shown an explicit wrong-letter overlay flip only $10.7\%$ of the time, and base-incorrect images are corrected at only $22\%$---below the $33\%$ that uniform compliance with the correct-pointing overlay would give---so overlay-induced flips concentrate in fragile-grounding cases rather than reflecting blanket instruction-following.}
\mydel{This bidirectionality suggests non-trivial representational shifts rather than systematic degradation.}

\subsection{Embedding Invariance: The Answer Can Stay While the Model Moves}

Decision stability alone can give a misleading impression of robustness.
To probe internal behaviour, we analyse representation drift across multiple context- and answer-conditioned representation regimes (\cref{app:regimes}).
\cref{tab:embinv} reports the mean cosine similarity and L2 distance between base and perturbed embeddings, averaged across perturbations and runs.

\paragraph{Pattern 1: TextOverlay induces the largest drift.}
\texttt{TextOverlay} produces substantially larger drift than geometric perturbations across all regimes. In \texttt{ctx\_open}, cosine similarity drops to $0.866$ with L2 distance exceeding $1000$, compared to $\sim0.99$ for translation and pad/crop.

\paragraph{Pattern 2: MCQ conditioning stabilises context representations.}
Context embeddings under MCQ prompts (\texttt{ctx\_mcq}) are more stable than under open prompts (\texttt{ctx\_open}), indicating explicit option conditioning constrains the representation space. However, this stability does not propagate to answer embeddings (\texttt{ans\_mcq}, \texttt{ans\_mcq\_free}), which remain susceptible to drift.
Predictions may remain unchanged while representations undergo substantial movement, highlighting a gap between output-level robustness and internal stability that motivates the control-drift analysis next. \cradd{As \texttt{ctx\_open} and \texttt{ctx\_mcq} are read at the final context token, before any answer is generated, they are unaffected by changes in the generated answer text; we therefore base our representation-drift conclusions on these context probes and on the option-constrained \texttt{ans\_mcq}, and treat the freely generated \texttt{ans\_open}/\texttt{ans\_mcq\_free} as additionally carrying a generation-variability component.}

\subsection{Drift Versus Control Drift}

\added{To gauge whether drift is local or comparable to inter-image variation, we compare against a control baseline of distances to randomly sampled other images, quantified via Cohen's $d$.}
Tables~\ref{tab:drift_ans_mcq_free_cos} (and~\ref{tab:drift_ans_mcq_free_l2} in the Appendix) summarise results for the \texttt{ans\_mcq\_free} regime, which exhibits the strongest effects.

\begin{figure*}[t]
\centering
\includegraphics[width=0.48\linewidth]{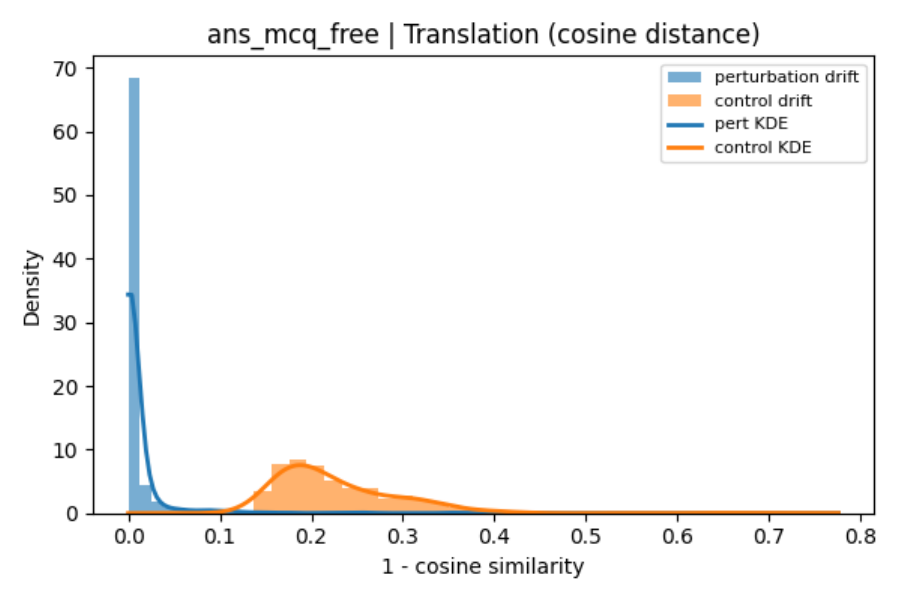}\hfill
\includegraphics[width=0.48\linewidth]{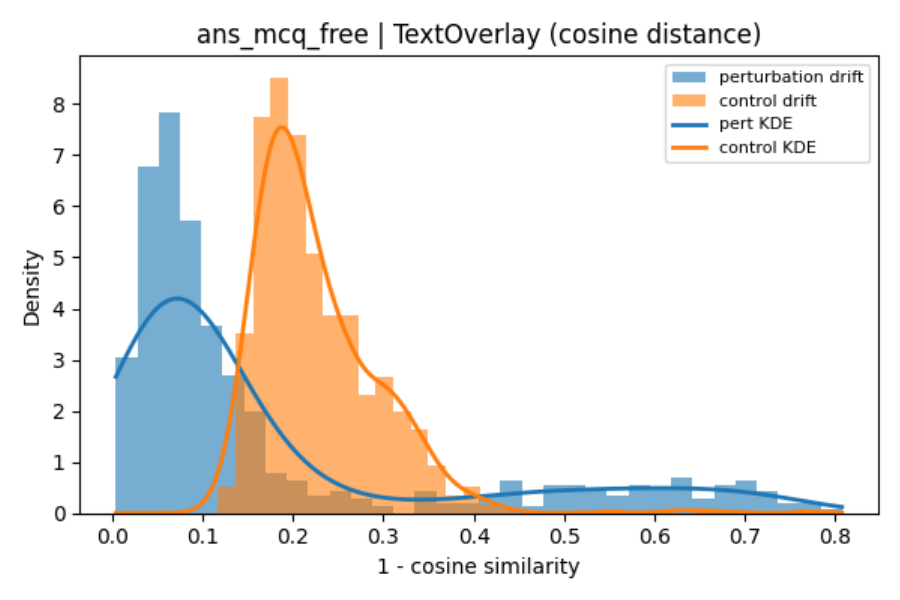}
\caption{
Cosine distance ($1-\cos$), Drift versus control drift for the \texttt{ans\_mcq\_free} embedding under Translation and Textoverlay perturbation.
Blue shows perturbation-induced drift relative to the base image; orange shows control drift (base image versus randomly sampled other images).
Left: Translation. Right: Textoverlay.
Unlike geometric perturbations, the Textoverlay perturbation-induced distribution does not remain well separated from control drift, indicating that the representation no longer stays local in embedding space.
}
\label{fig:drift_ans_mcq_free_translation}
\end{figure*}

Translation, pad/crop, and scale produce drift well separated from control drift (large negative Cohen's $d$), indicating local deformations, while \texttt{TextOverlay} produces drift substantially overlapping with control drift (Figure \ref{fig:drift_ans_mcq_free_translation}, Figure~\ref{fig:drift_ans_mcq_free_textoverlay} (Appendix)  and Table \ref{tab:drift_ans_mcq_free_cos}). Mean perturbation drift ($\approx 0.177$) approaches control mean ($\approx 0.228$), yielding Cohen's $d \approx -0.34$, implying text overlays move representations toward regions of unrelated images. \added{The random-pair control is deliberately conservative: tighter controls (\eg, within-category pairs) would yield smaller control distances and make this overlap even more pronounced, so our analysis is a lower bound on severity.}
Even with unchanged predictions, representations may no longer reside in neighbourhoods of the original image\added{, motivating the spectral analysis next}.



\begin{table}[t]
\centering
\resizebox{\linewidth}{!}{%
\small
\begin{tabular}{lcc}
\toprule
{\bf Perturbation} & {\bf $\mathrm{\AVg}$ (flip-rate)} & {\bf $\mathrm{\Ve}$ (vulnerability)} \\
\midrule
Translation & 0.062 & 0.168 \\
Pad/Crop    & 0.065 & 0.169 \\
Scale       & 0.079 & 0.079 \\
Scale+Pad   & 0.080 & 0.100 \\
Rotation    & 0.122 & 0.166 \\
TextOverlay(semantic) & 0.192 & 0.239 \\
TextOverlay(random) & 0.064 & 0.086 \\
TextOverlay(empty) & 0.043 & 0.044 \\
\midrule
Any (union) & 0.079 & 0.376 \\
\bottomrule
\end{tabular}}
\caption{Label invariance statistics on SEEDBench samples using log-likelihood MCQ scoring. \cradd{All values are means over four independent runs; the cross-run standard deviation is $\le 0.005$ for the flip rate ($\mathrm{\AVg}$) and $\le 0.008$ for vulnerability ($\mathrm{\Ve}$), so cross-scale differences far exceed run-to-run noise.}}
\label{tab:fliprates}
\end{table}

\begin{table}[t]
\centering
\resizebox{\linewidth}{!}{%
\small
\begin{tabular}{lccccc}
\toprule
{\bf Perturbation} & {\bf R$\to$W} & {\bf W$\to$R} & {\bf R$\to$R} & {\bf W$\to$W} & {\bf GT inst.} \\
\midrule
Translation  & 694 & 618 & 16538 & 10094 & 27944 \\
Pad/Crop     & 725 & 642 & 16507 & 10070 & 27944 \\
Scale        & 95  & 106 & 2059  & 1233  & 3493  \\
Scale+Pad    & 232 & 181 & 4076  & 2497  & 6986  \\
Rotation     & 390 & 262 & 3918  & 2416  & 6986  \\
TextOverlay  & 685 & 881 & 5738  & 3126  & 10430 \\
\midrule
Any          & 2821 & 2690 & 48836 & 29436 & 83783 \\
\bottomrule
\end{tabular} }
\caption{
Prediction correctness transitions under natural perturbations.
R$\to$W: base-correct predictions that become wrong under perturbation; W$\to$R: base-wrong predictions that become correct.
Counts are reported at the perturbation-instance level for 3500 base samples.
}
\label{tab:gt_transitions}
\end{table}

\begin{table*}[t]
\centering
\resizebox{\linewidth}{!}{%
\small
\begin{tabular}{lccccc}
\toprule
{\bf Type} &
{\bf ctx\_open (cos / L2)} &
{\bf ctx\_mcq (cos / L2)} &
{\bf ans\_open (cos / L2)} &
{\bf ans\_mcq (cos / L2)} &
{\bf ans\_mcq\_free (cos / L2)} \\
\midrule
Translation & 0.989 / 226 & 0.995 / 104 & 0.980 / 141 & 0.994 / 82 & 0.991 / 65 \\
Pad/Crop    & 0.988 / 245 & 0.995 / 113 & 0.979 / 153 & 0.993 / 88 & 0.991 / 70 \\
Scale       & 0.984 / 283 & 0.993 / 132 & 0.977 / 171 & 0.991 / 103 & 0.990 / 80 \\
Scale+Pad   & 0.981 / 299 & 0.992 / 145 & 0.974 / 184 & 0.990 / 113 & 0.988 / 89 \\
Rotation    & 0.963 / 458 & 0.976 / 262 & 0.959 / 276 & 0.983 / 182 & 0.980 / 142 \\
TextOverlay & 0.866 / 1027& 0.918 / 551 & 0.922 / 465 & 0.963 / 310 & 0.820 / 484 \\
\midrule
Any         & 0.970 / 360 & 0.984 / 181 & 0.970 / 201 & 0.988 / 124 & 0.968 / 128 \\
\bottomrule
\end{tabular} }
\caption{Embedding invariance summary (mean similarity to base) across five representation regimes.}
\label{tab:embinv}
\end{table*}


\begin{table}[t]
\centering
\small
\begin{tabular}{lcc}
\toprule
Perturbation & Drift (\% of control) & Cohen's $d$ \\
\midrule
Translation  & $4.1 \pm 14.8$  & $-3.88$ \\
Pad/Crop     & $4.6 \pm 16.7$  & $-3.77$ \\
Scale        & $4.8 \pm 21.9$  & $-3.50$ \\
Scale+Pad    & $5.7 \pm 18.7$  & $-3.63$ \\
Rotation     & $8.8 \pm 18.5$  & $-3.52$ \\
TextOverlay  & $77.7 \pm 88.6$ & $-0.34$ \\
\bottomrule
\end{tabular}
\caption{
Drift versus control drift for the \texttt{ans\_mcq\_free} embedding measured using cosine distance ($1-\cos$).
\textbf{Drift} is expressed as a percentage of mean control drift ($\mu_{\text{ctrl}} = 0.228$), i.e., the typical distance between unrelated images; values are reported as mean $\pm$ std.
Cohen's $d$ quantifies separation between distributions (values near $0$ indicate overlap).
Geometric perturbations induce drift below 10\% of inter-image variability, while TextOverlay approaches 78\%, indicating near-complete displacement to unrelated regions.
}
\label{tab:drift_ans_mcq_free_cos}
\end{table}

\subsection{Vision-Token Smoothness and Dirichlet Energy}
\label{sec:dirichlet}
\added{Embedding drift captures \emph{where} representations move; Dirichlet energy captures \emph{how} local structure reorganises.}

We 
compute $\Delta E_{\text{dir}} = E_{\text{dir}}(x') - E_{\text{dir}}(x)$\added{ on the 2D grid of vision tokens} for each perturbation. Different perturbations induce distinct patterns: Translation shows small positive shifts ($10.34 \pm 67.49$), TextOverlay
negative shifts ($-33.87 \pm 60.14$), and Rotation
the strongest effect ($-72.73 \pm 99.95$). Flip-inducing instances show larger absolute deviations, indicating spatial reorganisation correlates with decision failures.
Complete analyses are in Appendix~\ref{appendix_dirichlet}.


\section{Scaling Behaviour and Cross-Dataset Robustness}
\label{sec:scaling}

\added{We evaluate whether the previous findings persist across model scale, datasets, and architectural families, using the same protocol (\cref{methodology and experiment}).}

\subsection{Model Scaling on SEEDBench}
\label{subsec:scaling_seedbench}

\added{We evaluate \texttt{Qwen3-VL-2B/4B/8B/32B} (Instruct) on the same SEEDBench subset to test whether robustness improves monotonically with capacity.}

\begin{figure*}[t]
\centering
\includegraphics[width=0.48\linewidth]{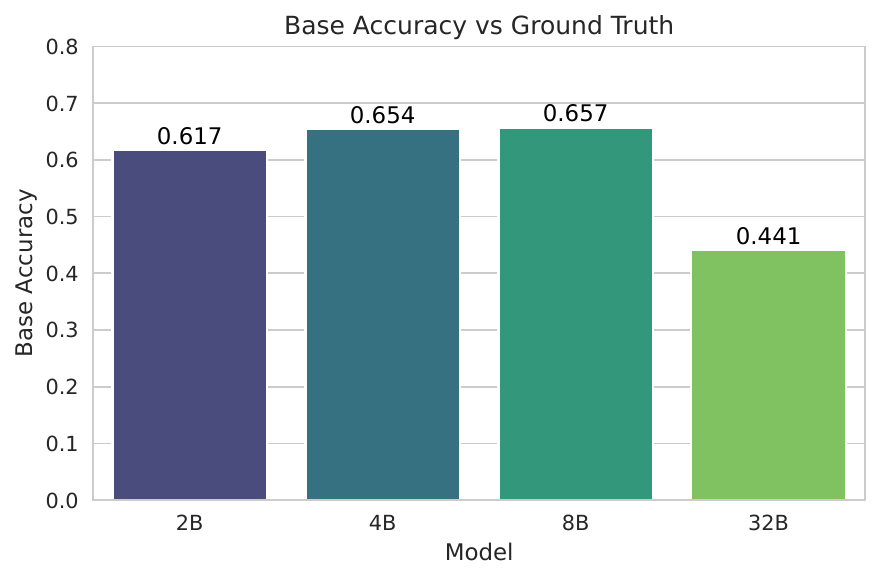}\hfill
\includegraphics[width=0.48\linewidth]{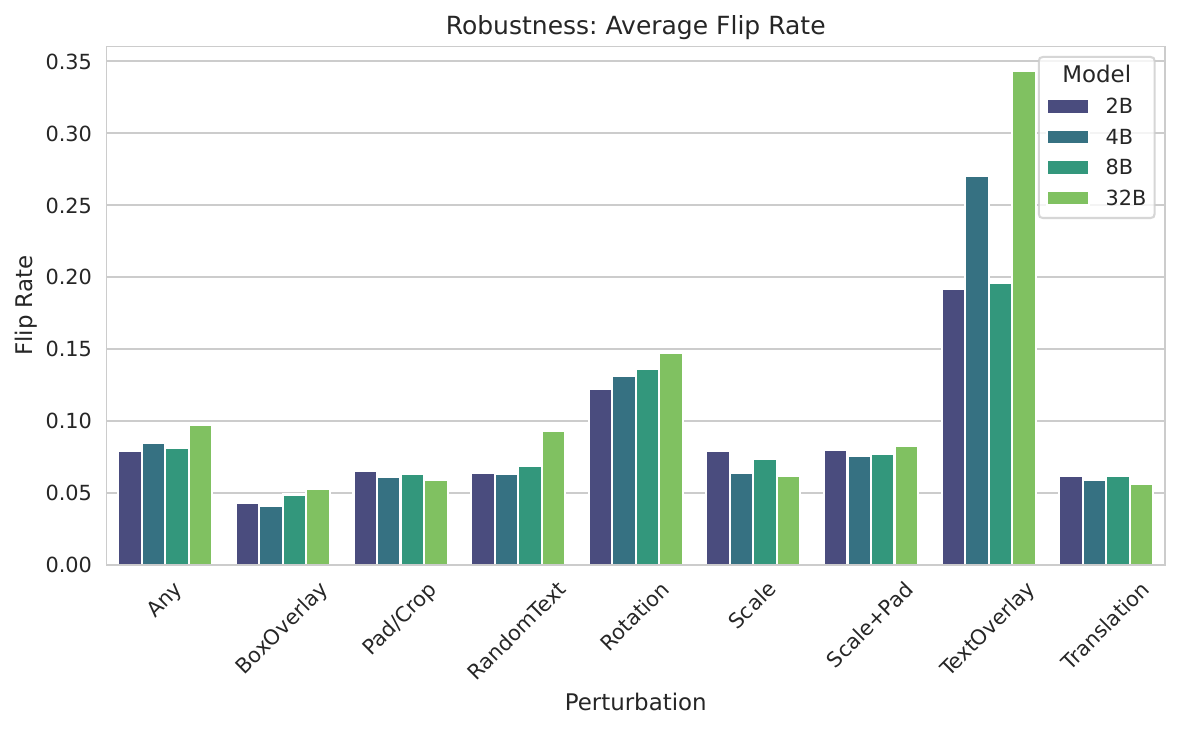}
\caption{
\textbf{Qwen3-VL (Instruct) scaling on SEEDBench.}
Left: base accuracy versus ground truth.
Right: average flip rate under natural perturbations (lower is better).
}
\label{fig:seedbench_scaling_overview}
\end{figure*}

\mydel{Base accuracy improves from 2B to 8B, but slightly degrades at 32B due to the restriction of answers to fixed multiple-choice options under log-likelihood scoring.} \myadd{Base accuracy improves from 2B to 8B; at 32B it is lower under our log-likelihood MCQ scoring~\citep{zheng2024robust}. As all scales use FP16, this is not a precision artifact, and we do not rely on 32B: the decoupling is already clear among scales with no accuracy drop (2B$\to$4B; Gemma-3 4B$\to$12B).}
More importantly, relative stability does not scale linearly with accuracy; larger models often exhibit comparable or higher flip rates under natural perturbations (Figure~\ref{fig:seedbench_scaling_overview}).
This decoupling is most pronounced for semantic text overlays, where \mydel{the largest models exhibit the highest instability}\myadd{larger models exhibit higher instability (the 4B flip rate exceeds the 2B despite higher accuracy)}.

Figure~\ref{fig:seedbench_scaling_transitions} in Appendix \ref{appendix_scaling_seedbench} decomposes flip behaviour into correctness transitions.
As model size increases, generally, both error injection (R$\rightarrow$W) and correction (W$\rightarrow$R) rates rise under perturbations.
This indicates that larger models develop sharper but more fragile decision boundaries, rather than uniformly improved robustness.

\paragraph{Cross-Dataset Validation on MMMU.}
To assess whether scaling-related robustness trends generalise beyond SEEDBench, we repeat the analysis on the MMMU benchmark, which features multi-disciplinary reasoning and more complex visual–textual dependencies.
Despite substantial differences in task structure and difficulty between SEEDBench and MMMU, the qualitative robustness trends remain consistent. All the results are reported in Appendix \ref{appendix_MMU}, in particular see Figure~\ref{fig:mmmu_scaling_overview} and \ref{fig:mmmu_scaling_transitions}.
\mydel{Accuracy improves with scale, while robustness under natural perturbations fails to improve monotonically.}\myadd{Accuracy improves with scale, yet every larger Qwen3-VL model is \emph{less} robust than the 2B (Figure~\ref{fig:mmmu_scaling_overview}).}
\mydel{Semantic text overlays again induce the largest instability, and correctness transitions exhibit the same bidirectional behaviour observed on SEEDBench.}
This consistency suggests that the observed robustness failures are not dataset-specific artifacts.
\paragraph{Cross-Architecture Robustness: LLaVA-OneVision Models.}
We extend our analysis to the LLaVA-OneVision  \cite{li2024llava} architecture, 
specifically,\texttt{llava-onevision-qwen2-0.5b} and \texttt{llava-onevision-qwen2-7b} using the same perturbation suite and protocol.
\added{Its} robustness behaviour closely mirrors that of Qwen3-VL
\added{~(Appendix~\ref{appendix_LLaVa})}.

\added{\paragraph{Cross-Family Validation on Gemma-3.}
We additionally evaluate Gemma-3~\citep{gemma3_2025} (\texttt{4B-IT}, \texttt{12B-IT}) on SEEDBench and MMMU. All findings replicate, and Gemma-3 mirrors the Qwen3-VL scale-decoupling: on MMMU the 12B model is substantially more sensitive than the 4B (TextOverlay $\mathrm{\AVg}$: $20.2\%\!\to\!31.8\%$; union $\mathrm{\Ve}$: $44.3\%\!\to\!54.0\%$), and on SEEDBench \mydel{drift also grows with scale (TextOverlay \texttt{ctx\_open} cos: $0.957\!\to\!0.918$)}\myadd{both label instability and representation drift grow with scale (TextOverlay flip $0.203\!\to\!0.316$; \texttt{ctx\_open} cos $0.957\!\to\!0.918$)}. Full tables are in Appendix~\ref{appendix_gemma3}.}

\added{Taken together, across three architecturally distinct families (Qwen3-VL, LLaVA-OneVision, Gemma-3) and eight model scales ($0.5$B--$32$B), robustness consistently fails to track accuracy gains, indicating a structural property of modern VLMs rather than a family-specific artifact\myadd{ (\cref{tab:consolidated})}.}

\begin{table}[t]
\centering
\small
\resizebox{\columnwidth}{!}{%
\begin{tabular}{lccc}
\toprule
\textbf{Model (SEEDBench)} & \textbf{Base acc} & \textbf{TextOverlay $\mathrm{\AVg}$} & \textbf{Union $\mathrm{\Ve}$} \\
\midrule
Qwen3-VL-2B          & 61.7 & 0.192 & 0.376 \\
Qwen3-VL-4B          & 65.4 & 0.270 & 0.476 \\
Qwen3-VL-8B          & 65.7 & 0.196 & 0.397 \\
Qwen3-VL-32B$^\dagger$ & 44.1 & 0.343 & 0.547 \\
LLaVA-OV-0.5B        & 33.0 & 0.076 & 0.199 \\
LLaVA-OV-7B          & 55.9 & 0.087 & 0.255 \\
Gemma-3-4B           & 47.0 & 0.203 & 0.471 \\
Gemma-3-12B          & 53.0 & 0.316 & 0.671 \\
\bottomrule
\end{tabular}}
\caption{Consolidated SEEDBench robustness across three families and eight scales: base accuracy (\%), TextOverlay instance-flip rate ($\mathrm{\AVg}$), and union image vulnerability ($\mathrm{\Ve}$). Within every family, base accuracy rises with scale while flip rate and vulnerability do \emph{not} decrease (Qwen $2\text{B}\!\to\!4\text{B}$; LLaVA $0.5\text{B}\!\to\!7\text{B}$; Gemma $4\text{B}\!\to\!12\text{B}$). $^\dagger$\,32B base accuracy is lower under log-likelihood MCQ scoring (all models are evaluated at FP16); it is reported for transparency, and the scale--robustness decoupling does not depend on it.}
\label{tab:consolidated}
\end{table}

\section{Frequency Dynamics and Spectral Drift}
\label{sec:freq_analysis}
\added{We probe the role of frequency content in VLM decisions by testing two hypotheses:}
(H1) \textit{Low-frequency dominance}, suggesting VLMs rely primarily on global shapes and should be robust to high-frequency noise \citep{yin2019fourier,H11,H12,H13};
(H2) \textit{Cross-frequency sensitivity}, suggesting that decision boundaries depend on the alignment of both low- and high-frequency components.
\added{We run three controlled experiments on SEEDBench --- random band-limited noise, progressive frequency ablation, and frequency-constrained PGD attacks~\citep{madry2018towards} --- detailed in Appendix~\ref{app:freq_suite}.}

Our analysis strongly supports H2 (Cross-frequency sensitivity). First, random noise injection reveals that high-frequency perturbations are just as effective at inducing flips as low-frequency ones (Figure~\ref{fig:freq_rand_noise_flip} in Appendix), contradicting the low-frequency dominance hypothesis. Second, frequency ablations show that margins degrade smoothly rather than abruptly, indicating that VLMs do not rely on a single "truth" band but require spectral coherence across components (Figure~\ref{fig:freq_ablation_margin}). 
These findings suggest that "benign" natural perturbations (like rotation or text overlay) induce failures not by destroying semantic content, but by causing \emph{spectral and phase drift} that decouples the visual encoding from the model's reasoning priors.

\added{\paragraph{Generalisation Beyond Benign Perturbations.} We further confirm this with frequency-constrained PGD~\citep{madry2018towards} at $\epsilon\!=\!8/255$ (Appendix~\ref{subsec:freq_pgd}): attack success rates exceed $79\%$ in pixel-space, low-frequency, and high-frequency regimes alike (Table~\ref{tab:freq_pgd}). The same cross-frequency sensitivity that drives benign-perturbation failures thus also exposes models to deliberate frequency-localised attacks, indicating that our findings extend beyond ``natural'' perturbations.}

\section{Does Drift Cause Hallucination? (POPE Analysis)}
\label{sec:exp_pope}
\added{We analyse POPE (Adversarial) to relate visual instability to hallucination via object existence verification.}
\added{Using a blank-image baseline as ``skeptic prior'', the model predicts ``No'' for nearly all samples without visual input.}
\added{For samples where the clean model hallucinates object presence (Figure~\ref{fig:pope_error_dynamics} in Appendix), perturbations shift margins toward the negative spectrum (avg.\ flip margin $\approx -0.96$).}
\added{This \emph{drift-to-prior} shows hallucinations here arise from fragile visual features rather than language bias: perturbations suppress these features, reducing false positives at the cost of recall.}
Detailed discussion is in Appendix \ref{appendix_POPE}.

\section{Conclusion} \label{sec:discussion_conclusion}

We presented a systematic analysis of robustness in VLMs 
%
%
across SEEDBench, MMMU, and POPE\added{, covering natural, spectral, and adversarial perturbations}.\mydel{ Our findings challenge the assumption that high base accuracy or increased model scale reliably confer robustness.}
Across all experiments, a consistent pattern emerges that perturbations induce substantial internal drift captured by embedding displacement and Dirichlet energy changes, well before any observable prediction flip
\mydel{, exposing a shared vulnerability in multimodal alignment that manifests differently by task geometry}.\mydel{ In reasoning tasks (SEEDBench, MMMU), high-frequency corruption degrades fine-grained discrimination}
%
%
\added{The scale-decoupling pattern (\cref{sec:scaling}) suggests larger capacity sharpens decision boundaries without stabilising representations\mydel{; text overlays are especially disruptive via the pixel-space prompt-injection effect~\citep{li2023-text_as_image}}.}
Our framework unifies natural, semantic, and adversarial failures\mydel{, showing robust VLMs require representation stability, structural smoothness, and asymmetric error dynamics beyond accuracy}.

\section*{Limitations}

Our analysis focuses on a specific set of vision-language models and benchmarks, which may not fully capture the diversity of architectures and task distributions in deployment. While we examine multiple perturbation families, the parameter ranges and specific transformations were chosen to reflect natural variations rather than exhaustive adversarial exploration—more extreme perturbations or targeted attacks may reveal different failure modes---as originally showcased in \citet{goodfellow2014explaining}. \myadd{We also note that a fixed pixel-magnitude perturbation (\eg, a $\pm 16$px shift) corresponds to different fractions of the effective input across families, which use different preprocessing resolutions; we therefore report cross-family conclusions as ordinal (the ranking of perturbations and probes) rather than absolute-magnitude comparisons.}

The frequency-domain analysis provides interpretable insights but relies on specific metrics (Dirichlet energy, spectral norms) that may not capture all aspects of representational drift. Alternative measures of trajectory stability or decision margin erosion could complement our findings. Additionally, our hook points target the final LLM layer, which is typically the target layer used for probing and model editing~\cite{nikandrou-etal-2024-enhancing}. However, this choice may ignore the nuances that characterise the early fusion dynamics within vision encoders or cross-attention mechanisms.

Our task-dependent observations (\eg, high-frequency noise reducing POPE hallucinations) suggest that robustness cannot be evaluated uniformly across all applications. A perturbation that improves calibration in one context may degrade performance in another, complicating the development of universal robustness interventions. Finally, while we document the scale-robustness decoupling, we do not propose architectural modifications or training objectives to address it
\added{; our findings nonetheless suggest concrete directions, such as representation-aware training objectives that penalise embedding drift under augmentation and Dirichlet-energy regularisation to enforce spatial coherence in vision tokens, both of which we leave for future work}.

\added{Our analysis is intentionally diagnostic rather than mechanistic. Our five probe regimes (\texttt{ctx\_open}, \texttt{ctx\_mcq}, \texttt{ans\_open}, \texttt{ans\_mcq}, \texttt{ans\_mcq\_free}) provide partial localisation by separating context- vs.\ answer-conditioned and open- vs.\ MCQ-conditioned representations, but they do not constitute a layer- or head-level causal attribution of where representation drift originates within the VLM pipeline (vision encoder, connector, or specific LLM layers). A full modular attribution---for example, layer-wise probing of embedding drift, attention-map entropy across heads, or gradient-based sensitivity to estimate local Lipschitz constants around each input---would provide mechanistic depth, and we view this as a natural extension of the present work that we plan to pursue in future work.}

\makeatletter
\ifacl@anonymize
\else
\section*{Acknowledgements}

This work was partially supported by two Sapienza projects; SEEDS (Sustainable Ecosystems in Evolving Digital Societies, PROGETTI DIPARTIMENTALI DI ATENEO, Sapienza, 231\_009\_23\_RAT\_SILVE) and EVOLVE (SAP\_RICERCA\_2025\_EVOLVE\_SILVES\_F\_01).
Aryo Pradipta Gema was supported by the United Kingdom Research and Innovation (grant EP/S02431X/1), UKRI Centre for Doctoral Training in Biomedical AI at the University of Edinburgh, School of Informatics.
Maria Sofia Bucarelli acknowledges support from the French government National Research Agency (ANR) through the UCA JEDI (ANR-15-IDEX-0001), through the EUR DS4H (ANR-17-EURE-0004), and through the 3IA Cote d'Azur Investments in the project with the reference number ANR-23-IACL-0001.
Rohit Saxena was supported by the Engineering and Physical Sciences Research Council (EPSRC) through the AI Hub in Generative Models (grant number EP/Y028805/1).
Wai-Chung Kwan was funded by the Huawei–Edinburgh Joint Research Laboratory Grant.
\fi
\makeatother

\bibliography{custom}

\clearpage

\appendix

\section{Detailed Experimental Setup}
\label{app:exp_details}
\paragraph{System configuration and reproducibility.}
All experiments are conducted with a fixed random seed (seed = 0). Unless otherwise specified, evaluations are performed on a single NVIDIA A100 GPU (80GB memory) using FP16 precision.
For Qwen3-VL-32B-Instruct, evaluation is distributed across four NVIDIA A100 GPUs (80GB each). Log-likelihood scoring is performed with a batch size of 4.

\subsection {Models}
We evaluate the following vision language models in a zero-shot setting:
\begin{itemize}
  \item Qwen3-VL (Instruct): 2B, 4B, 8B, 32B
  \item LLaVA-OneVision: 0.5B, 7B
  \item Gemma-3 (Instruct): 4B, 12B
\end{itemize}
All models are evaluated without fine-tuning.

\subsection {Datasets and Sample Counts}

\begin{table}[h]
\resizebox{\linewidth}{!}{%
\centering
\begin{tabular}{lccc}
\toprule
Dataset & Split & Base Samples & Task \\
\midrule
SEEDBench & Full & 14000 & MCQ \\
MMMU & Validation & 847 & Multi-image MCQ \\
POPE & Adversarial & 3000 & Yes/No \\
\bottomrule
\end{tabular}}
\caption{Datasets and evaluation splits used in the paper.}
\end{table}

Each base sample is evaluated under multiple perturbation instances \ref{app:naturalpertubations}.
\subsection{Prompt Templates (Prompt Regimes)}
\label{app:prompts}

We use chat-style multimodal prompts where images are embedded in the user message content.
The textual instruction depends on the representation probes regime ~\ref{app:regimes}. In all regimes, the image(s) are provided as part of the user
content, followed by the text prompt.

\paragraph{\texttt{ans\_mcq} regime }
This regime explicitly lists the options and instructs the model to select from them:

{\small
\begin{verbatim}
User:
  [IMAGE]
  Question: <question>
  Options:
  A. <option A>
  B. <option B>
  C. <option C>
  D. <option D>
  Please select the correct answer from 
  the options above.
Assistant:
  <option text>
\end{verbatim}
}

\paragraph{\texttt{ans\_open} regime }
This regime does not surface the options and requests a concise answer:
{\small
\begin{verbatim}
User:
  [IMAGE]
  Question: <question>
  Please answer the question concisely.

\end{verbatim}
}

\paragraph{\texttt{ans\_mcq\_free} regime }

This regime provides the options for context but requests a free-form answer:
{\small
 \begin{verbatim}
User:
  [IMAGE]
  Question: <question>
  Options (for your reference; answer freely):
  A. <option A>
  B. <option B>
  C. <option C>
  D. <option D>
  Provide the best answer in your own words.
\end{verbatim}
}

\paragraph{Multi-image MMMU formatting.}
For MMMU, questions may include markers such as \texttt{<image 1>}, \texttt{<image 2>}, etc.
We interleave images into the user message at the referenced positions; any remaining images are
appended afterward. The instruction line is still determined by the prompt regime above.

\paragraph{POPE (Yes/No).}
For POPE, the user message contains the image and a binary question.
The assistant response is restricted to a single token (\texttt{Yes} or \texttt{No}).

{\small
\begin{verbatim}
User:
  [IMAGE]
  Question: <question>
  Answer with exactly one word: Yes or No.

Assistant:
  Yes / No
\end{verbatim}
}
\subsection {Visual Perturbation Examples}
\begin{figure}[h]
    \centering
    \includegraphics[width=0.48\linewidth]{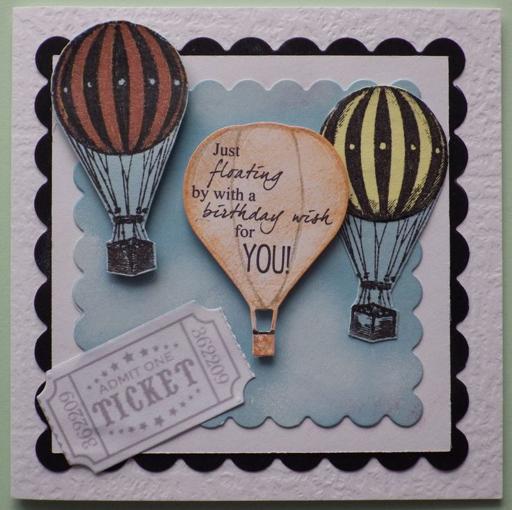}
    \hfill
    \includegraphics[width=0.48\linewidth]{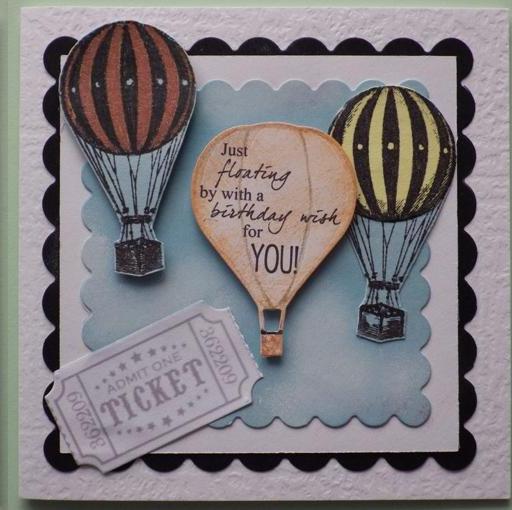}
    \\[1ex] 
    \includegraphics[width=0.48\linewidth]{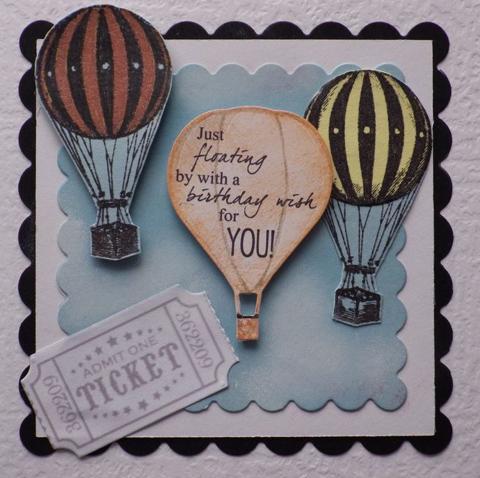}
    \hfill
    \includegraphics[width=0.48\linewidth]{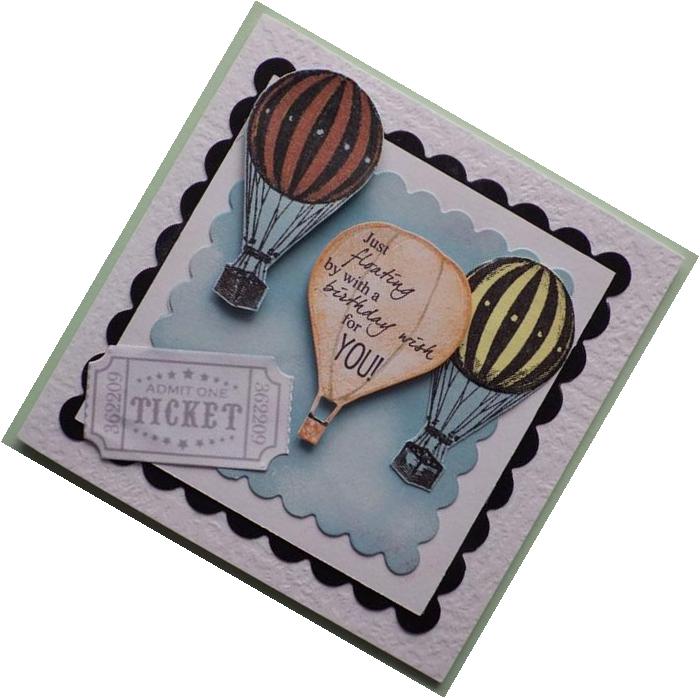}
    
    \caption{Visual perturbation examples applied to a SEEDBench sample used in our robustness evaluation. The figure displays the \textbf{original base image (Top-Left)} alongside three geometric transformations: \textbf{translation (Top-Right)}, \textbf{padding and cropping (Bottom-Left)}, and \textbf{$-30^{\circ}$ rotation (Bottom-Right)}. These perturbations serve to test the model's structural consistency under spatial variation without altering semantic content.}
    \label{fig:pertubationexample}
\end{figure}

We ensure that all applied perturbations remain \mydel{semantically benign}\myadd{content-preserving for a human observer under mild transforms}, meaning a human observer would still easily recognise the original content. As shown in Figure~\ref{fig:pertubationexample}, while operations like rotation ($-30^{\circ}$) or padding significantly shift pixel distributions, the core visual evidence required for reasoning remains intact. \myadd{Spatial-relation questions are a partial exception, since rotation or translation can alter the reference frame; we therefore report a flip-rate breakdown by SEEDBench evaluation dimension as a check in future work, and separately verify (Appendix~\ref{appendix_scaling_seedbench}) that photometric corruptions which cannot alter spatial layout still induce comparable flips.}

\subsection{Data Pre-processing}
We utilise the official evaluation splits for SEEDBench (Image), MMMU (Validation), and POPE (Adversarial).
For MMMU, we process multi-image examples by applying the perturbation $v'$ to all image components $\{I_k\}$ within the sequence $x$ simultaneously. 

\subsection{Perturbation Hyperparameters}
\label{app:naturalpertubations}
We define the perturbation families with the following parameter sweeps:
\begin{itemize}
    \item \textbf{Translation (cyclic):} horizontal wrap-around (cyclic) shifts by $\Delta x \in \{-16,-12,\ldots,16\}\setminus\{0\}$ pixels.
    \item \textbf{Pad/Crop:} symmetric padding or cropping by $n \in \{-16,-12,\ldots,16\}\setminus\{0\}$ pixels.
    \item \textbf{Scale:} rescaling by a factor $\alpha$ (default $\alpha=0.9$) followed by resizing back to the original resolution.
    \item \textbf{Scale+Pad:} rescaling followed by padding with a uniform background.
    \item \textbf{Rotation:} in-plane rotation by $\pm 30^\circ$ with interpolation.
    \item \textbf{Text Overlays (three variants).}
    To disentangle occlusion and edge injection from \emph{semantic steering}, we evaluate three overlay families with identical geometry but different content:
    \begin{itemize}
        \item \textbf{Semantic overlay (\texttt{TextOverlay})}: short directive phrases such as ``Answer is A/B/C/D'' rendered on the image.
        \item \textbf{Random-text overlay (\texttt{RandomText})}: same-sized text region filled with random strings of comparable length and ink density.
        \item \textbf{Empty-box overlay (\texttt{BoxOverlay})}: the same box region rendered without text, controlling for occlusion and shape.
    \end{itemize}
\end{itemize}

Each perturbation type is applied multiple times per image, producing a set of perturbed variants $x' \in v'(x)$ for each base image $x$, here $v'$ is perturbation type .

Unless explicitly stated otherwise, \texttt{TextOverlay} refers to the semantic overlay variant.
We report \texttt{RandomText} and \texttt{BoxOverlay} separately when analyzing overlay-specific effects.

\myadd{\paragraph{Overlay letter assignment.} For each image, \texttt{TextOverlay} renders ``Answer is $L$'' for up to three option letters $L$, chosen as the options \emph{other} than the model's own base prediction. Each image therefore receives at most three overlays, each naming an option the model did not select, and the correct option can appear as an overlay only for images the model initially answered incorrectly. This design underlies the asymmetric $\mathrm{W}\to\mathrm{R} > \mathrm{R}\to\mathrm{W}$ transition counts in \cref{sec:embedding_drift}: overlays can correct an already-incorrect prediction but can only mislead an already-correct one. Consistent with this, base-correct images shown an explicit ``wrong-letter'' overlay flip only $10.7\%$ of the time, and base-incorrect images are corrected at $22\%$---below the $33\%$ that uniform compliance with the correct-pointing overlay would produce---indicating that the effect is fragility-gated rather than blanket instruction-following.}

\myadd{\subsection{Supplementary: Photometric Corruptions on GQA}}
\myadd{To separate representational instability from any change in ground truth under geometric transforms, we additionally evaluate \texttt{Qwen3-VL-2B-Instruct} on 100 GQA images under a broader corruption suite. Photometric corruptions that cannot move objects or alter spatial relations still flip predictions: Gaussian blur ($8.0\%$), JPEG-70 compression ($7.4\%$), motion blur ($4.3\%$), and Gaussian noise ($2.9\%$) flip at rates comparable to the geometric perturbations, whereas brightness and contrast shifts are nearly inert ($\le 1.8\%$). Because these corruptions preserve object identity and spatial layout, the resulting flips cannot reflect ground-truth changes, so the instability we report is representational. This is a supplementary consistency check on a different dataset and a broader perturbation set; its numbers are not directly comparable to the main SEEDBench tables.}

\subsection{Representation Probes}
\label{app:regimes}
We extract hidden states $h \in \mathbb{R}^d$ from the final layer of the LLM backbone at five specific hook points:
\begin{itemize}
\item \textbf{\texttt{ctx\_open}:} The last token of the visual context under an open-ended prompt.
\item \textbf{\texttt{ctx\_mcq}:} The last token of the visual context under the specific MCQ prompt template.
\item \textbf{\texttt{ans\_open}:} The mean-pooled embedding of the generated answer tokens (greedy decoding).
\item \textbf{\texttt{ans\_mcq}:}  The mean-pooled embedding of the generated answer conditioned by MCQ.
\item \textbf{\texttt{ans\_mcq\_free}:} The mean-pooled embedding of the generated answer when the model is forced to answer freely, constrained to the option set.
\end{itemize}

\subsection{Control Drift Baseline}
\label{app:controldrift}
To normalise embedding distances, we compute Cohen's $d$ between the perturbation drift distribution $D_{pert} = \{ \|e(x) - e(x')\| \}$ and a control distribution $D_{ctrl} = \{ \|e(x_i) - e(x_j)\| \}$ derived from 1,000 random pairs of unrelated images~\cref{tab:drift_ans_mcq_free_l2}.

\section{Drift versus Control Drift, L2 results}
\label{appendix_drift}
\begin{figure*}[t!]
\centering
\includegraphics[width=0.48\linewidth]{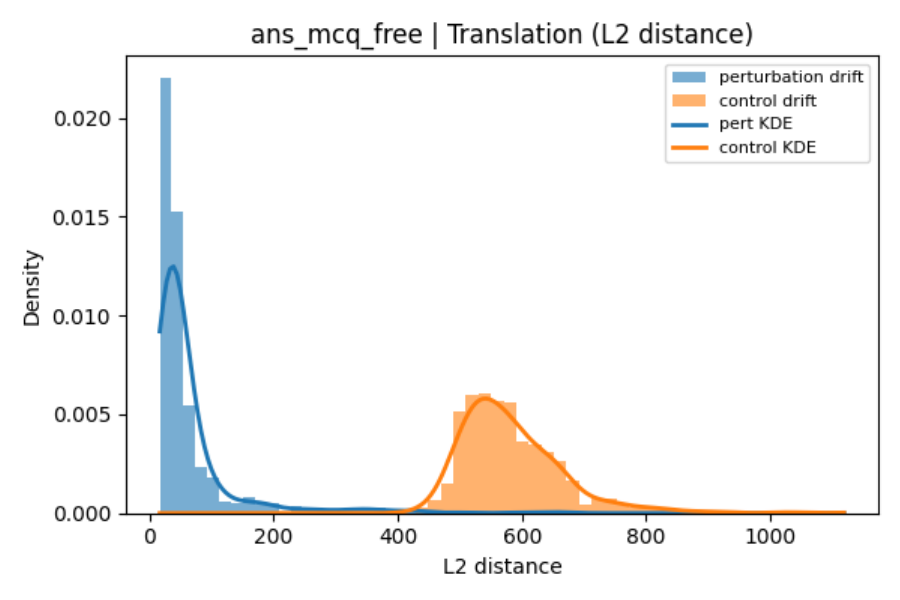}
\includegraphics[width=0.48\linewidth]{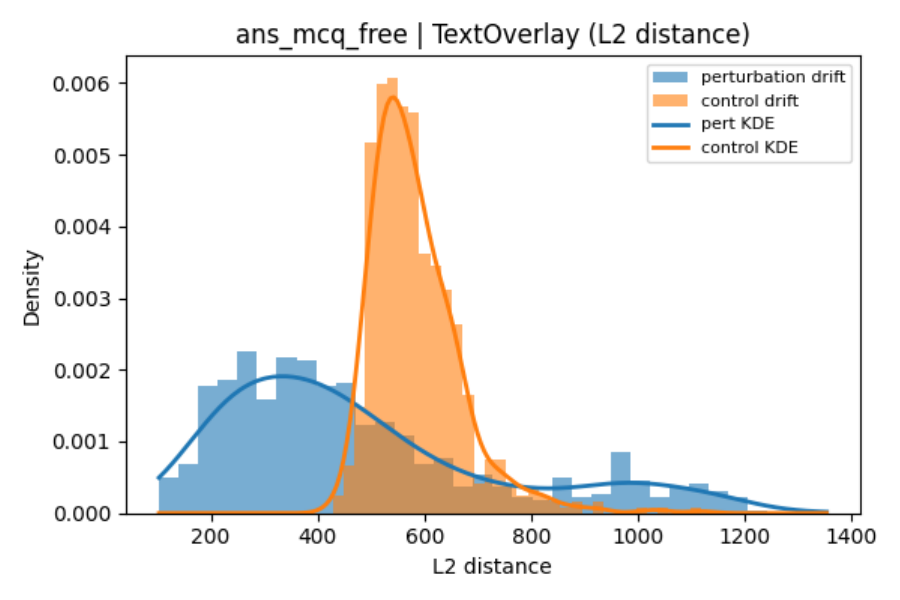}
\caption{
L2 distance, Drift versus control drift for the \texttt{ans\_mcq\_free} embedding under Translation and text overlay perturbations.
Blue shows perturbation-induced drift relative to the base image; orange shows control drift (base image versus randomly sampled other images).
Left: Translation. Right: Textoverlay.
Unlike geometric perturbations, the perturbation-induced distribution does not remain well separated from control drift, indicating that the representation no longer stays local in embedding space.
}
\label{fig:drift_ans_mcq_free_textoverlay}
\end{figure*}

\section{Model Scaling }
\label{app:scaling}
\subsection{SeedBench Dataset}
\label{appendix_scaling_seedbench}
Here we report some additional results on model scaling for SeedBench dataset.

\begin{figure*}[t]
\centering
\includegraphics[width=0.48\linewidth]{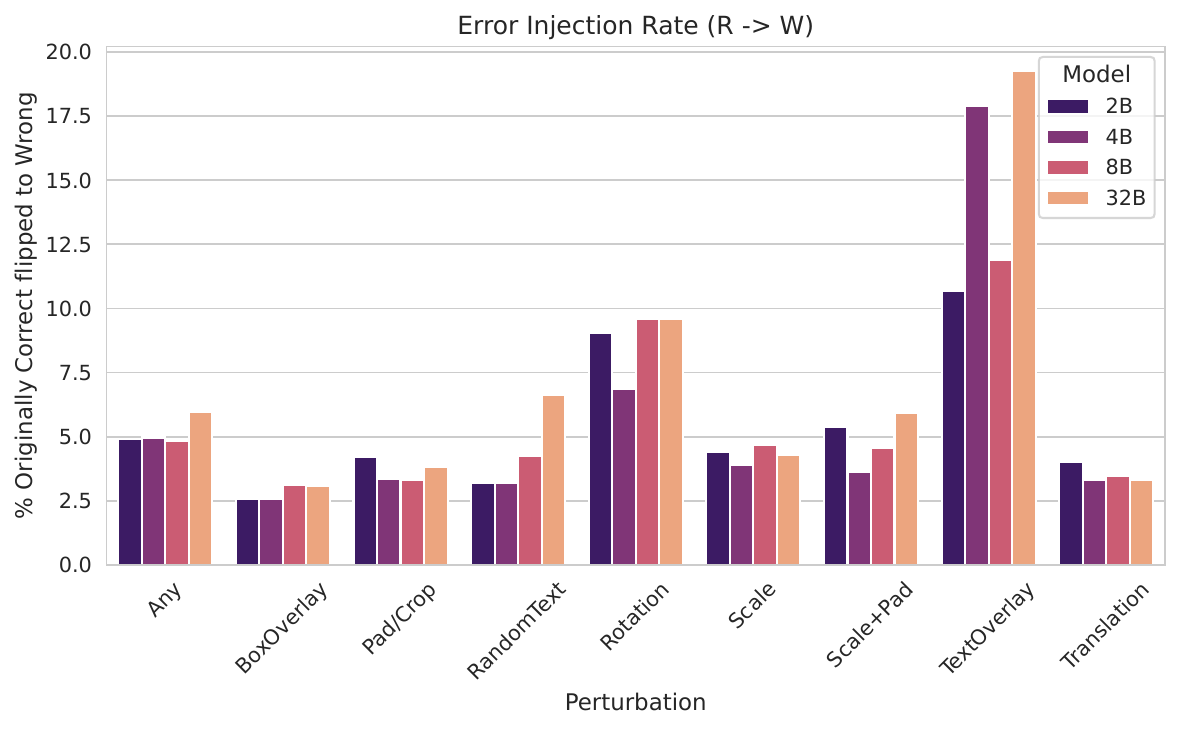}\hfill
\includegraphics[width=0.48\linewidth]{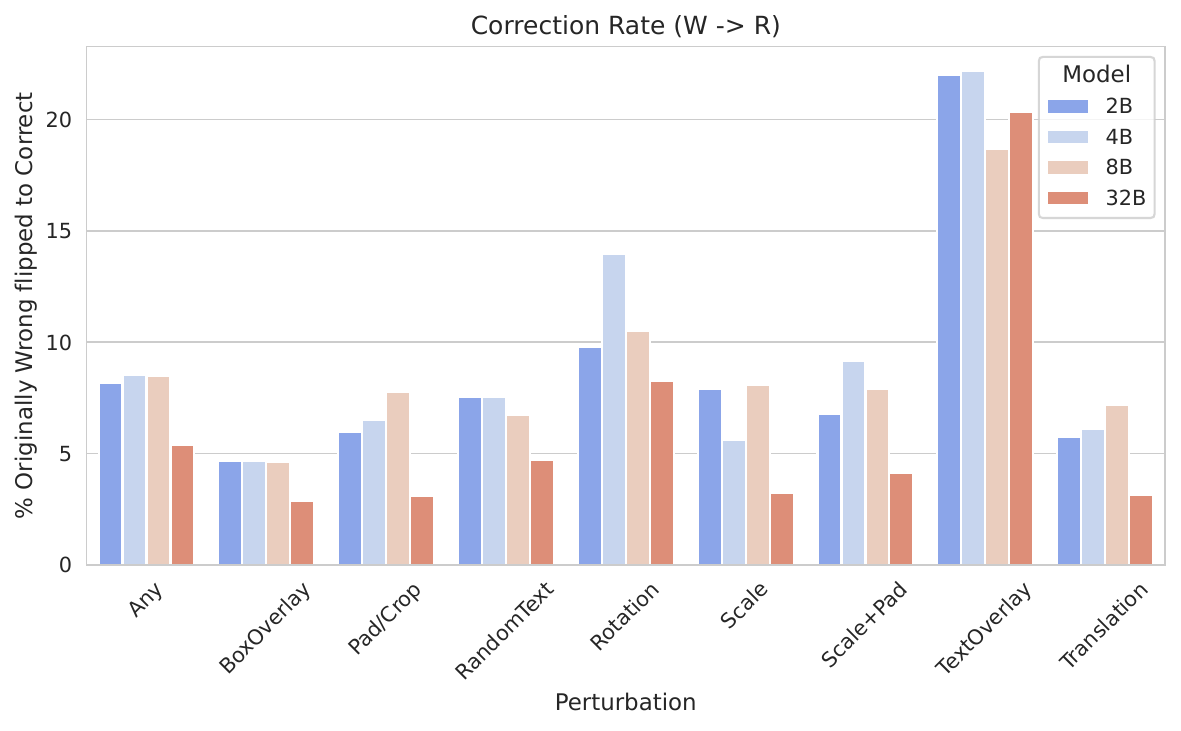} 
\includegraphics[width=0.48\linewidth]{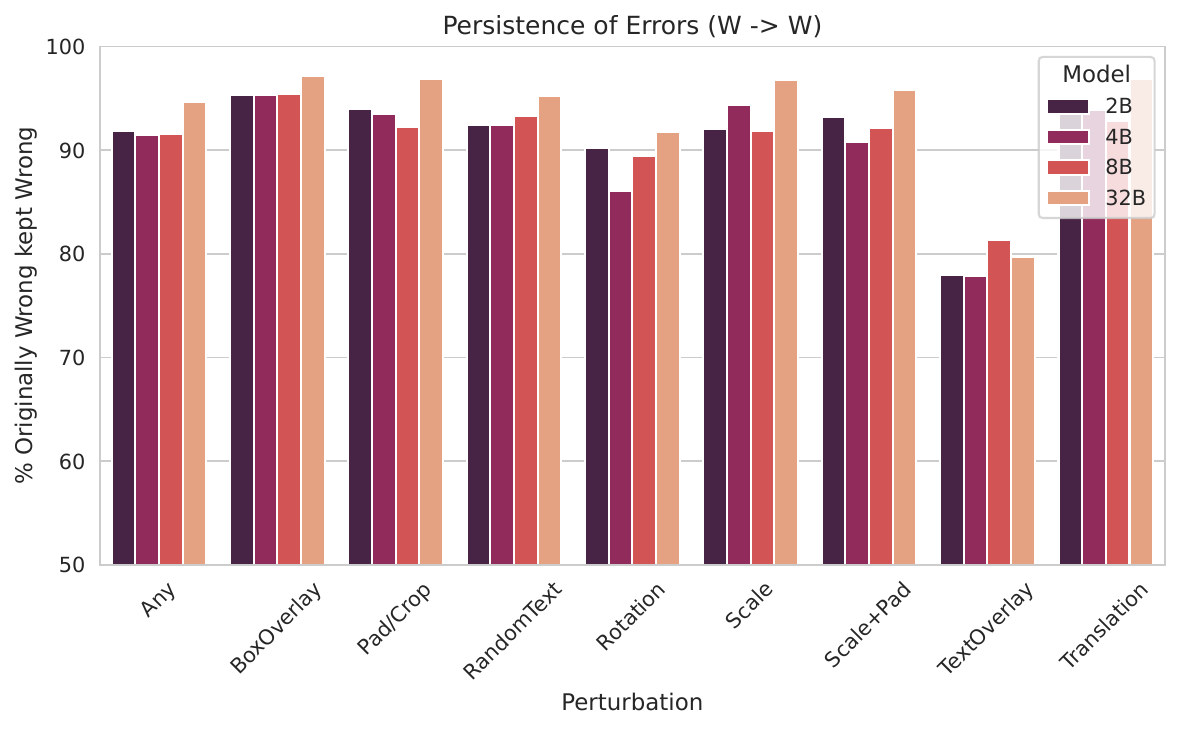}\hfill
\includegraphics[width=0.48\linewidth]{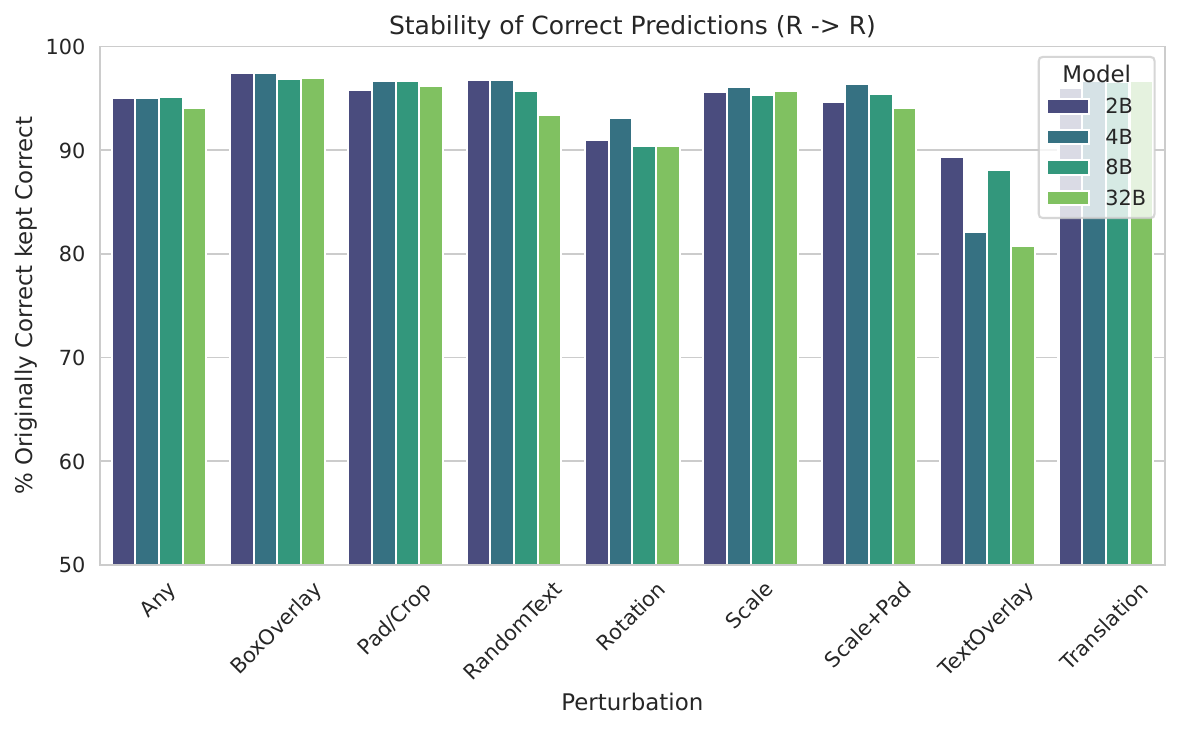}
\caption{
Correctness transition statistics under natural perturbations for different Qwen3-VL model scales on SEEDBench.
}
\label{fig:seedbench_scaling_transitions}
\end{figure*}

Figure~\ref{fig:seedbench_scaling_transitions} shows that as model size increases, both error injection (R$\rightarrow$W) and correction (W$\rightarrow$R) rates rise, indicating larger models develop sharper but more fragile decision boundaries rather than uniformly improved robustness.

\subsection{Cross-Dataset Validation on MMMU}
\label{appendix_MMU}

\begin{figure*}[t]
\centering
\includegraphics[width=0.48\linewidth]{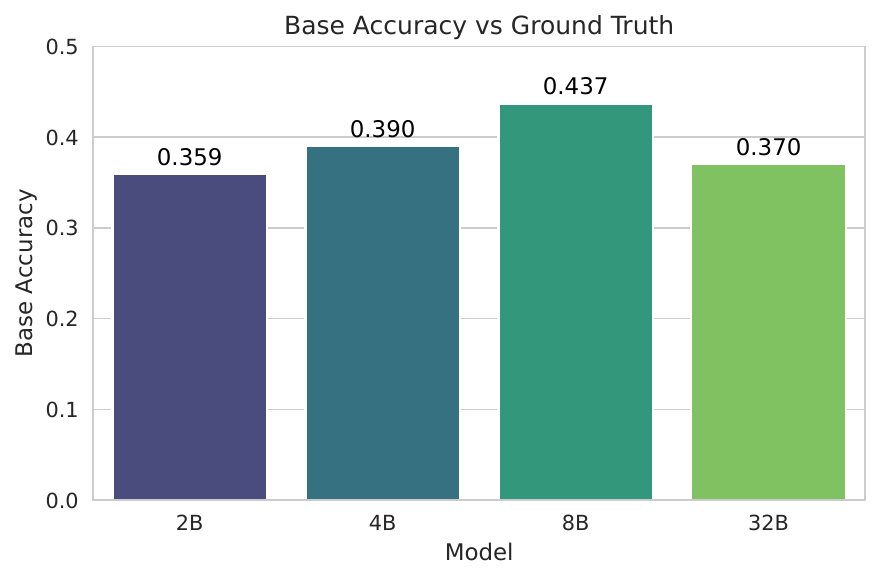}\hfill
\includegraphics[width=0.48\linewidth]{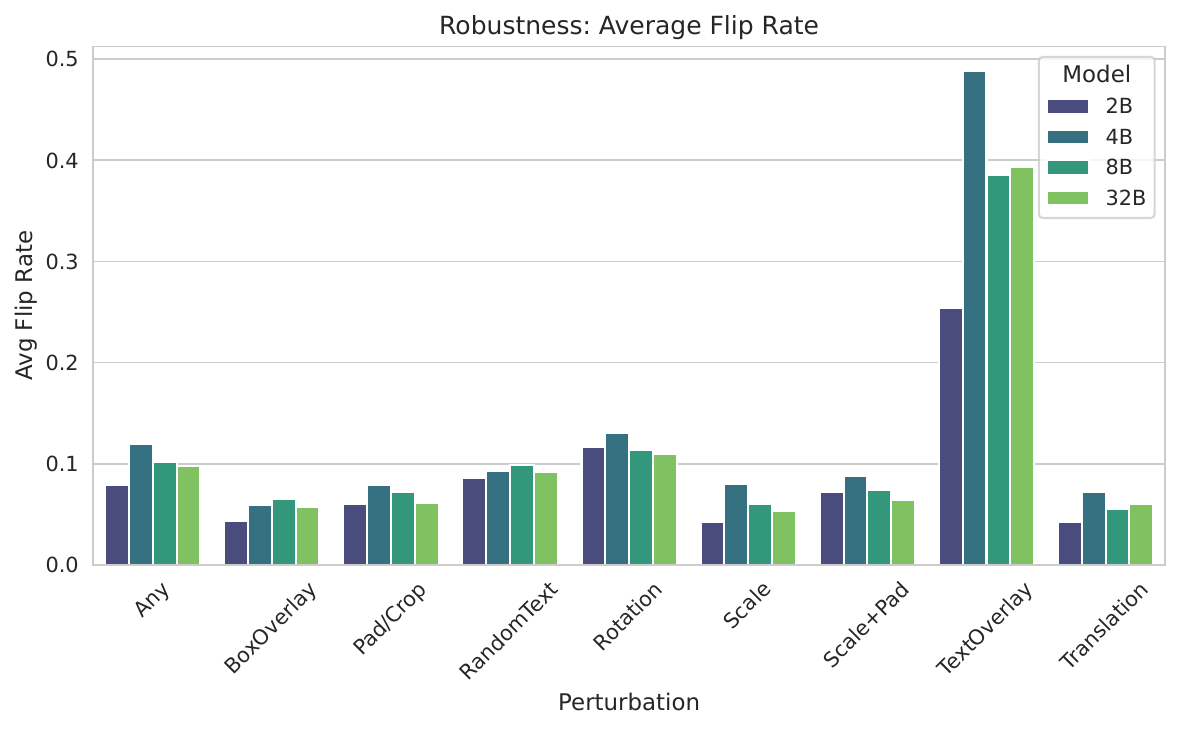}
\caption{
\textbf{Qwen3-VL (Instruct) scaling on MMMU.}
Left: base accuracy versus ground truth.
Right: average flip rate under natural perturbations.
}
\label{fig:mmmu_scaling_overview}
\end{figure*}

\begin{figure*}[t]
\centering
\includegraphics[width=0.48\linewidth]{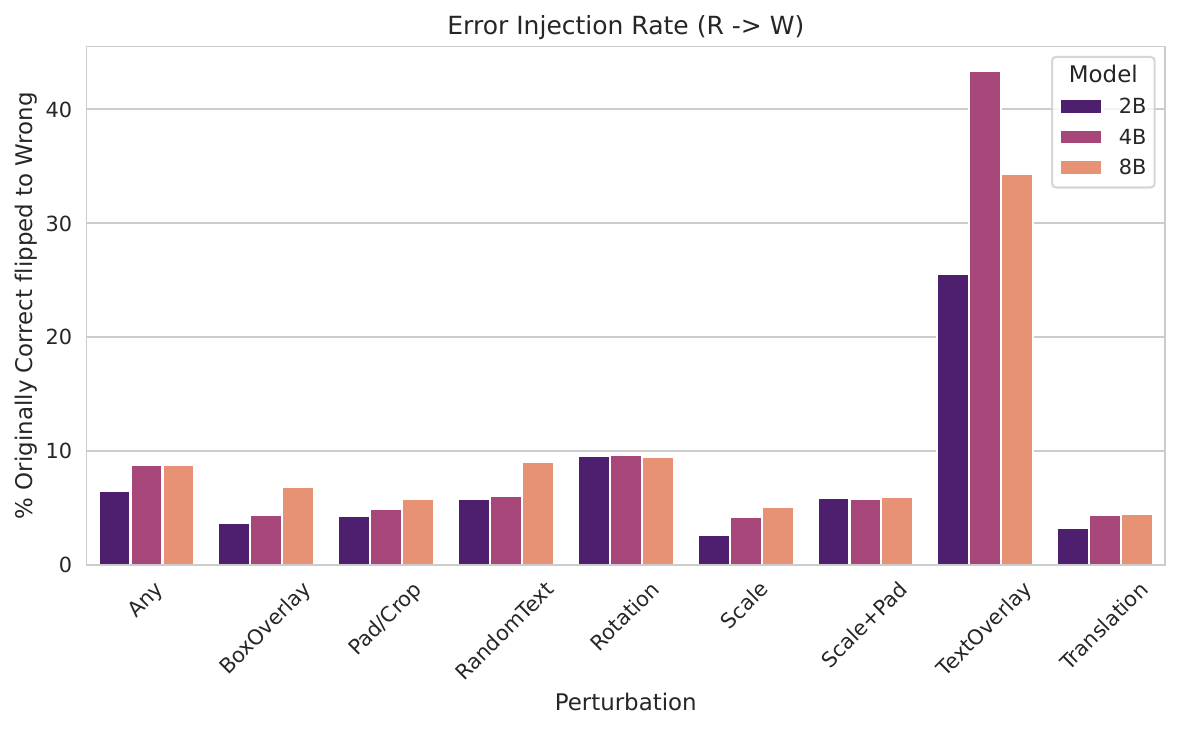} \hfill
\includegraphics[width=0.48\linewidth]{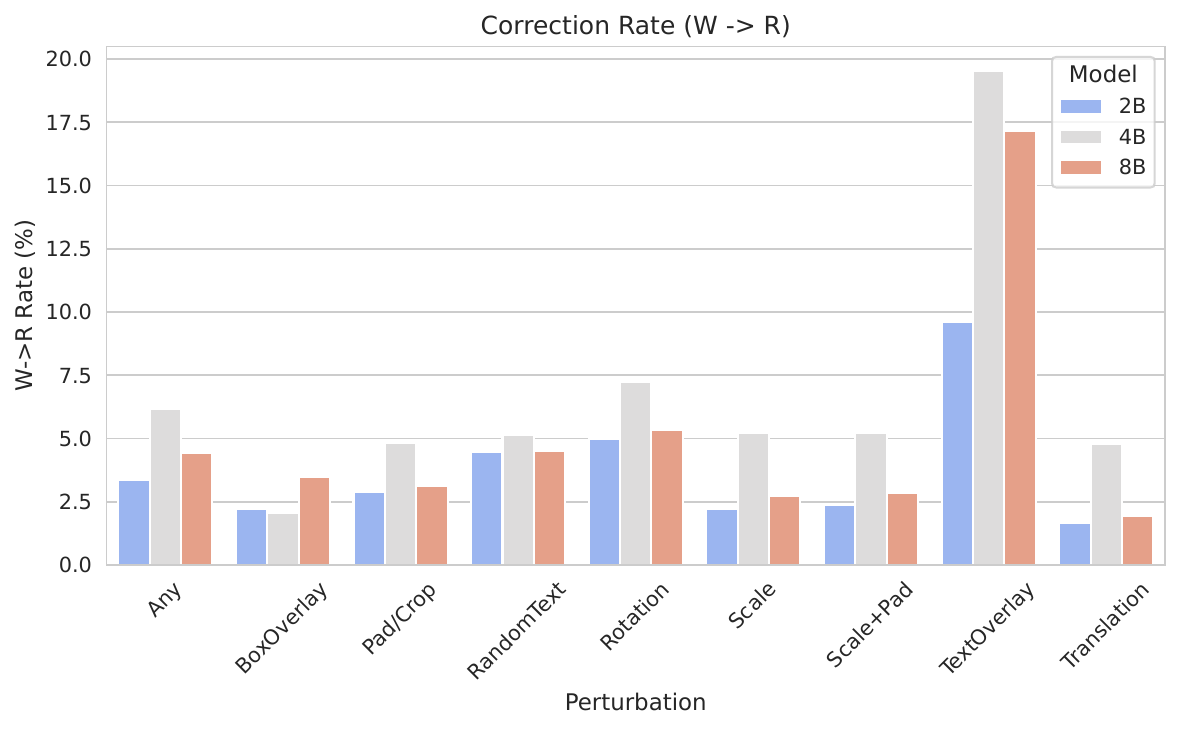}
\includegraphics[width=0.48\linewidth]{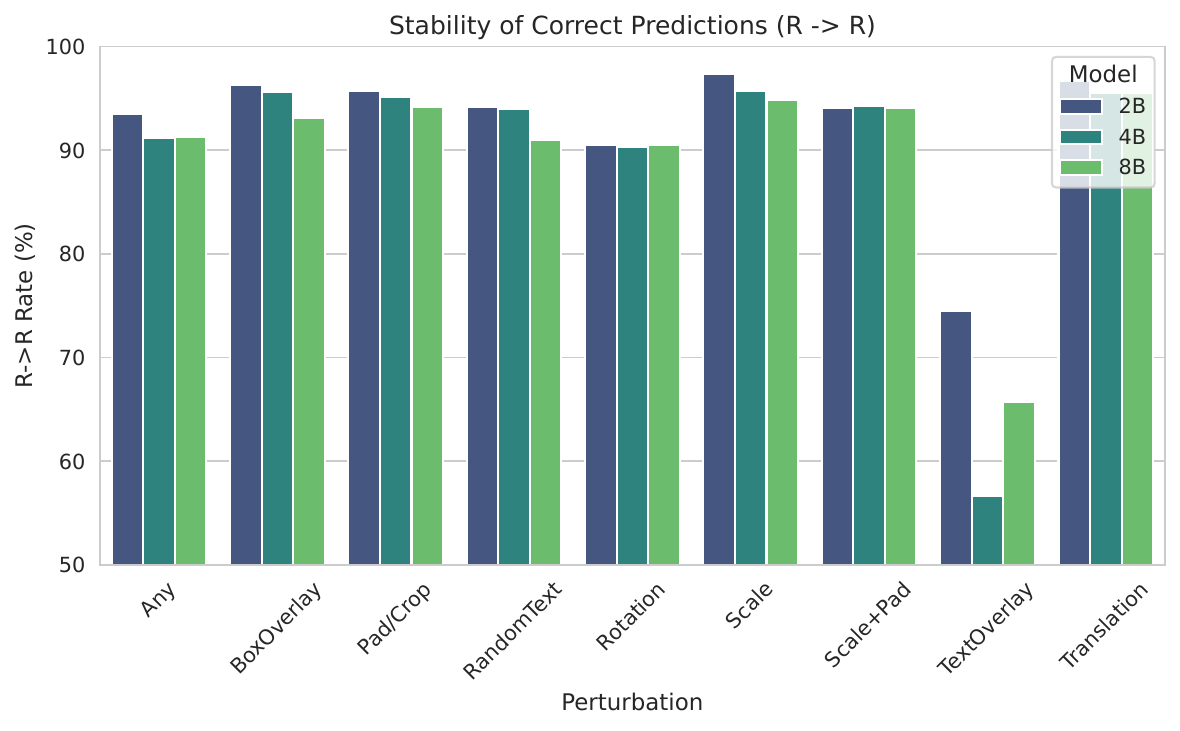}\hfill
\includegraphics[width=0.48\linewidth]{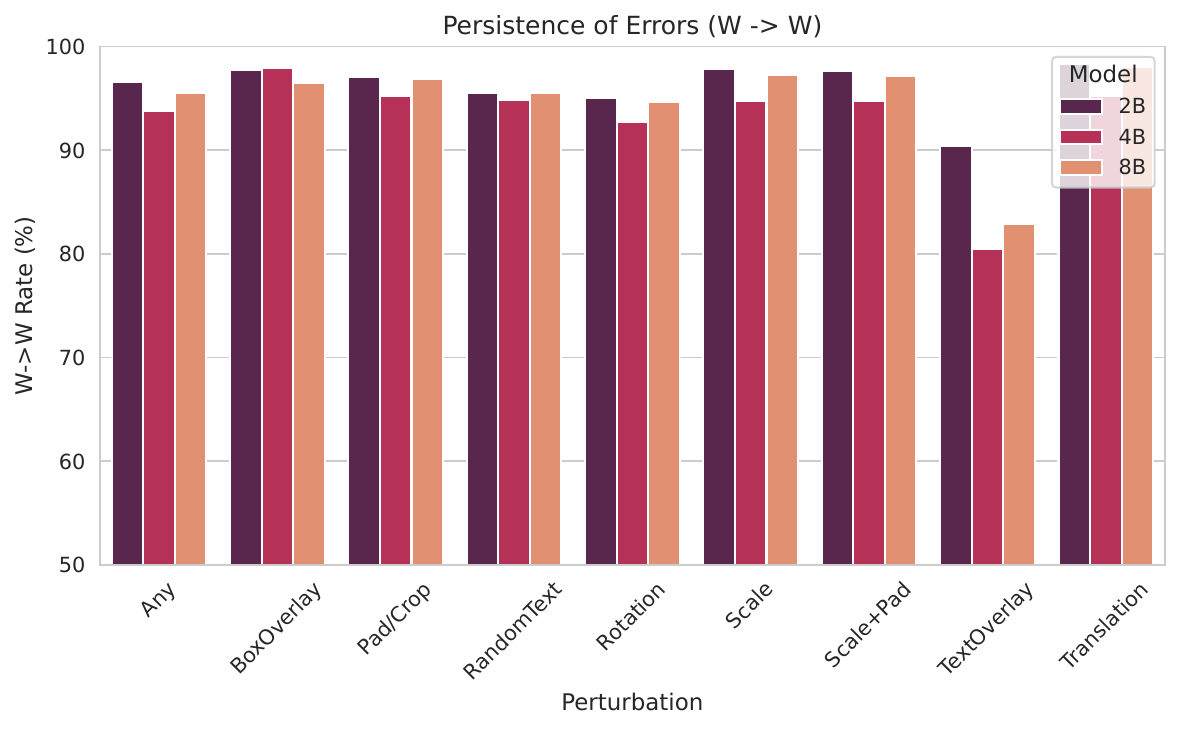}
\caption{
Correctness transition statistics under natural perturbations for different Qwen3-VL model scales on MMMU.
}
\label{fig:mmmu_scaling_transitions}
\end{figure*}

To assess whether scaling-related robustness trends generalise beyond SEEDBench, we repeat the analysis on the MMMU benchmark, which features multi-discipline reasoning and more complex visual–textual dependencies.
Despite substantial differences in task structure and difficulty between SEEDBench and MMMU, the qualitative robustness trends remain consistent (Figure~\ref{fig:mmmu_scaling_overview} and \ref{fig:mmmu_scaling_transitions}).

\myadd{On MMMU (847 validation samples), \emph{every} larger Qwen3-VL model is less robust than the 2B: the TextOverlay instance-flip rate rises from $0.254$ (2B) to $0.488$ (4B), $0.385$ (8B), and $0.393$ (32B), and union vulnerability from $0.453$ to $0.705$, $0.612$, and $0.616$ respectively, while base accuracy generally increases ($0.359\!\to\!0.437$ up to 8B; 32B lower under log-likelihood MCQ scoring). LLaVA-OneVision shows the same trend across its two scales ($0.5$B$\to$7B: TextOverlay flip $0.086\!\to\!0.126$, union $0.249\!\to\!0.333$; base accuracy $0.285\!\to\!0.379$), and Gemma-3 likewise (4B$\to$12B; \cref{tab:gemma_mmmu_fliprates}). The decoupling therefore holds on MMMU as well, and if anything more cleanly, across all three families.}

\begin{table}[h]
\centering
\small
\begin{tabular}{lcccc}
\toprule
Perturbation &
$\mu_{\text{pert}}$ &
$\sigma_{\text{pert}}$ &
$\mu_{\text{ctrl}}$ &
Cohen's $d$ \\
\midrule
Translation  & 64.42  & 78.78  & 584.19 & $-6.3522$ \\
Pad/Crop     & 70.98  & 84.18  & 584.19 & $-6.0757$ \\
Scale        & 77.47  & 104.29 & 584.19 & $-5.3325$ \\
Scale+Pad    & 91.12  & 103.06 & 584.19 & $-5.2256$ \\
Rotation     & 140.57 & 98.37  & 584.19 & $-4.8316$ \\
TextOverlay  & 483.77 & 266.40 & 584.19 & $-0.5080$ \\
\bottomrule
\end{tabular}
\caption{
Drift versus control drift for the \texttt{ans\_mcq\_free} embedding measured using L2 distance.
TextOverlay exhibits substantially reduced separation from control drift, consistent with non-local displacement in representation space.
}
\label{tab:drift_ans_mcq_free_l2}
\end{table}

\subsection{Cross-Architecture Robustness: LLaVA-OneVision Models}
\label{appendix_LLaVa}
To assess whether the observed robustness phenomena are specific to the Qwen3-VL family, we extend our analysis to the LLaVA-OneVision architecture.
We evaluate \texttt{llava-onevision-qwen2-0.5b} and \texttt{llava-onevision-qwen2-7b} using the same perturbation suite and evaluation protocol.

\begin{figure*}[t]
\centering
\includegraphics[width=0.48\linewidth]{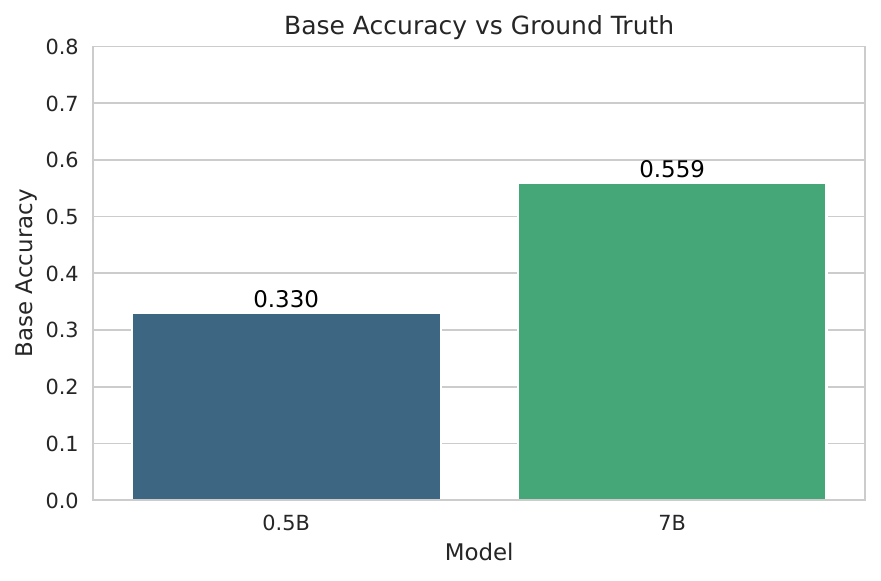} \hfill
\includegraphics[width=0.48\linewidth]{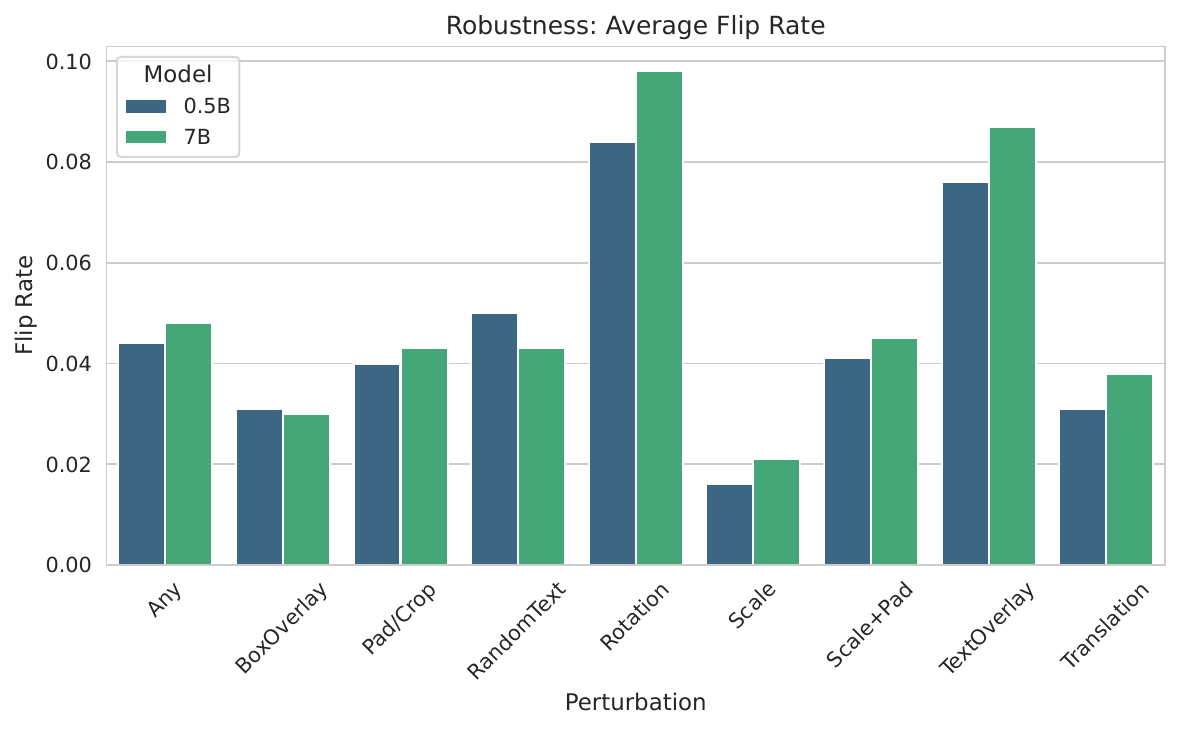}
\caption{
\textbf{LLaVA-OneVision scaling on SEEDBench.}
left: base accuracy versus ground truth.
Right: average flip rate under natural perturbations.
}
\label{fig:llava_seedbench_scaling}
\end{figure*}

\begin{figure*}[t]
\centering
\includegraphics[width=0.48\linewidth]{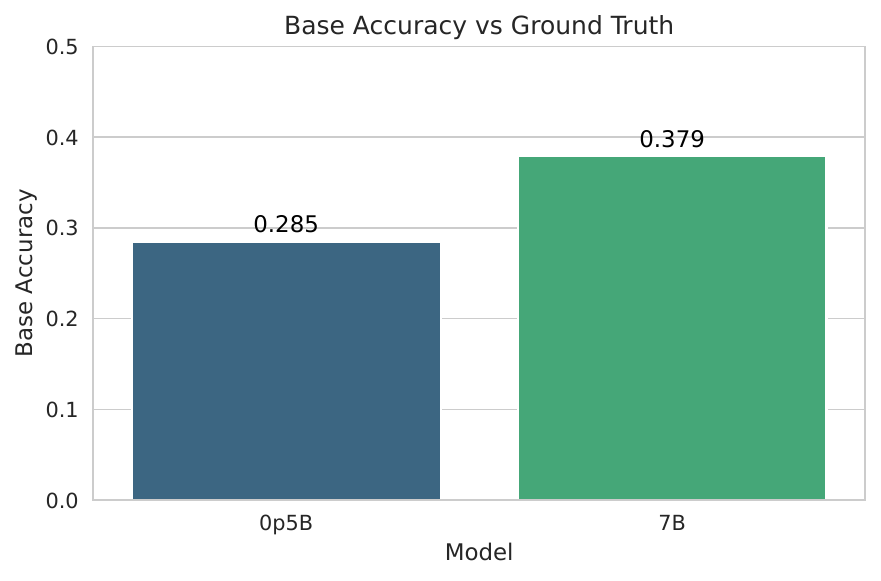} \hfill
\includegraphics[width=0.48\linewidth]{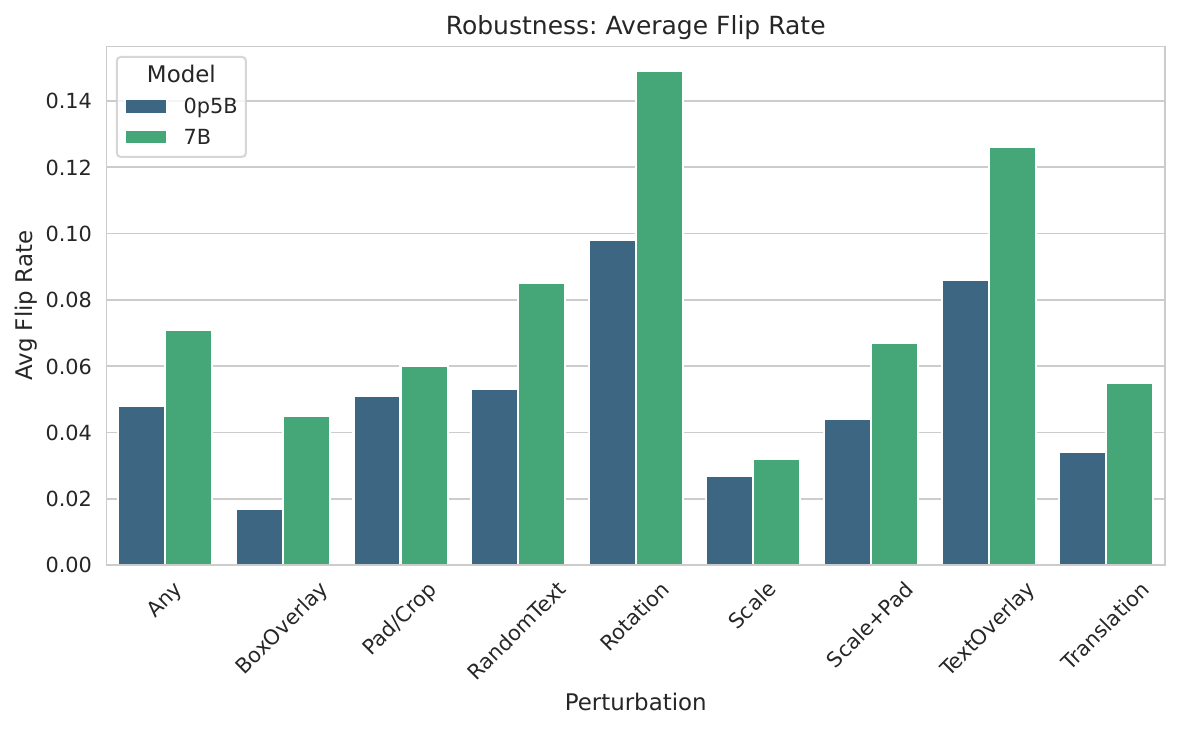}
\caption{
\textbf{LLaVA-OneVision scaling on MMMU.}
Left: base accuracy versus ground truth.
Right: average flip rate under natural perturbations.
}
\label{fig:llava_mmmu_scaling}
\end{figure*}

Although LLaVA-OneVision employs a different vision backbone and multimodal fusion strategy, its robustness behaviour closely mirrors that of Qwen3-VL (Figure~\ref{fig:llava_seedbench_scaling} and \ref{fig:llava_mmmu_scaling}).
Accuracy improves with scale, while robustness remains comparable or degrades, particularly under semantic text overlays.
Correctness transitions again show increased error injection and correction at larger scales, indicating sharper yet more fragile decision boundaries.
These results demonstrate that robustness failures persist across architectural families.

\added{\subsection{Cross-Family Validation on Gemma-3}}
\added{\label{appendix_gemma3}}
\added{Following the breadth of experimental framework across model families, we additionally evaluated the Gemma-3 family~\citep{gemma3_2025} (Gemma-3-4B-IT and Gemma-3-12B-IT) on SEEDBench and MMMU using the identical evaluation protocol described in Section~\ref{methodology and experiment}. For brevity, we report Gemma-3-4B-IT on SEEDBench in detail (3,500 samples, $\sim$600 for drift analysis); Gemma-3-12B-IT and MMMU results follow the same qualitative trends. The full result tables are presented below.}

\added{Tables~\ref{tab:gemma_fliprates}--\ref{tab:gemma_drift_ans_open_cos} confirm that all key findings reported on Qwen3-VL and LLaVA-OneVision generalise to this third, architecturally distinct family: (i)~TextOverlay remains the most disruptive perturbation; (ii)~the three-way overlay decomposition preserves the semantic > random > empty ordering; (iii)~at least one perturbation flips the prediction for $47.1\%$ of images, indicating the phenomenon is at least as severe as in Qwen3-VL-2B ($37.6\%$); (iv)~W$\to$R transitions are comparable to or exceed R$\to$W transitions, confirming bidirectional movement across decision boundaries; and (v)~MCQ-conditioned probes (\texttt{ctx\_mcq}, \texttt{ans\_mcq}) are consistently more stable than open-ended ones (\texttt{ctx\_open}, \texttt{ans\_open}), so our MCQ-based main-text findings represent a conservative lower bound on the instability observed under open-ended generation.}

\begin{table}[h]
\added{\centering
\small
\begin{tabular}{lcc}
\toprule
\textbf{Perturbation} & \textbf{$\mathrm{\AVg}$ (flip-rate)} & \textbf{$\mathrm{\Ve}$ (vulnerability)} \\
\midrule
Translation         & 0.097 & 0.229 \\
Pad/Crop            & 0.102 & 0.249 \\
Scale               & 0.071 & 0.071 \\
Scale+Pad           & 0.114 & 0.151 \\
Rotation            & 0.173 & 0.234 \\
TextOverlay         & 0.203 & 0.327 \\
RandomText          & 0.124 & 0.180 \\
BoxOverlay          & 0.084 & 0.098 \\
\midrule
Any (union)         & \textbf{0.116} & \textbf{0.471} \\
\bottomrule
\end{tabular}
\caption{Label invariance statistics for \texttt{Gemma-3-4B-IT} on SEEDBench (log-likelihood MCQ scoring).}
\label{tab:gemma_fliprates}}
\end{table}

\begin{table}[h]
\added{\centering
\small
\begin{tabular}{lcccc}
\toprule
\textbf{Perturbation} & \textbf{R$\to$W} & \textbf{W$\to$R} & \textbf{R$\to$R} & \textbf{W$\to$W} \\
\midrule
Translation     & 616  & 647  & 5880  & 20801 \\
Pad/Crop        & 636  & 675  & 5860  & 20773 \\
Scale           & 60   & 63   & 752   & 2618 \\
Scale+Pad       & 159  & 209  & 1465  & 5153 \\
Rotation        & 261  & 300  & 1363  & 5062 \\
TextOverlay     & 402  & 601  & 2016  & 7411 \\
RandomText      & 257  & 348  & 2161  & 7664 \\
BoxOverlay      & 186  & 193  & 2232  & 7819 \\
\midrule
Any             & \textbf{2577} & \textbf{3036} & \textbf{21729} & \textbf{77301} \\
\bottomrule
\end{tabular}
\caption{Prediction correctness transitions under perturbations for \texttt{Gemma-3-4B-IT} on SEEDBench (instance level).}
\label{tab:gemma_gt_transitions}}
\end{table}

\begin{table*}[t]
\added{\centering
\resizebox{\linewidth}{!}{%
\small
\begin{tabular}{lccccc}
\toprule
\textbf{Type} &
\textbf{ctx\_open (cos / L2)} &
\textbf{ctx\_mcq (cos / L2)} &
\textbf{ans\_open (cos / L2)} &
\textbf{ans\_mcq (cos / L2)} &
\textbf{ans\_mcq\_free (cos / L2)} \\
\midrule
Translation  & 0.981 / 61.37  & 0.997 / 24.34 & 0.955 / 52.28 & 0.986 / 28.52 & 0.976 / 28.35 \\
Pad/Crop     & 0.978 / 66.31  & 0.995 / 29.21 & 0.953 / 54.49 & 0.986 / 30.94 & 0.974 / 29.69 \\
Scale        & 0.988 / 48.32  & 0.997 / 20.48 & 0.965 / 44.08 & 0.992 / 20.53 & 0.982 / 24.03 \\
Scale+Pad    & 0.974 / 72.46  & 0.995 / 32.59 & 0.947 / 59.25 & 0.985 / 33.34 & 0.971 / 31.97 \\
Rotation     & 0.949 / 104.00 & 0.984 / 54.84 & 0.923 / 78.22 & 0.966 / 53.14 & 0.955 / 41.72 \\
TextOverlay  & 0.957 / 102.22 & 0.968 / 81.12 & 0.901 / 93.01 & 0.947 / 69.44 & 0.958 / 41.55 \\
RandomText   & 0.972 / 74.55  & 0.990 / 43.48 & 0.939 / 65.70 & 0.976 / 42.70 & 0.966 / 35.19 \\
BoxOverlay   & 0.987 / 49.75  & 0.995 / 30.15 & 0.958 / 49.86 & 0.984 / 32.19 & 0.979 / 25.84 \\
\midrule
Any          & \textbf{0.975 / 70.06} & \textbf{0.992 / 36.23} & \textbf{0.945 / 59.94} & \textbf{0.980 / 36.71} & \textbf{0.971 / 31.44} \\
\bottomrule
\end{tabular}}
\caption{Embedding invariance (mean similarity to base) across five representation regimes for \texttt{Gemma-3-4B-IT} on SEEDBench.}
\label{tab:gemma_embinv}}
\end{table*}

\begin{table}[h]
\added{\centering
\small
\begin{tabular}{lccc}
\toprule
\textbf{Perturbation} & \textbf{Pert. Drift} & \textbf{Ctrl. Drift} & \textbf{Cohen's $d$} \\
\midrule
Translation  & 0.024 & 0.229 & $-8.09$ \\
Pad/Crop     & 0.026 & 0.229 & $-7.91$ \\
Scale        & 0.018 & 0.229 & $-8.04$ \\
Scale+Pad    & 0.029 & 0.229 & $-7.20$ \\
Rotation     & 0.045 & 0.229 & $-5.89$ \\
TextOverlay  & 0.042 & 0.229 & $-6.88$ \\
RandomText   & 0.034 & 0.229 & $-6.65$ \\
BoxOverlay   & 0.020 & 0.229 & $-7.73$ \\
\bottomrule
\end{tabular}
\caption{Drift vs.\ control drift for \texttt{Gemma-3-4B-IT}, \texttt{ans\_mcq\_free} regime (cosine distance, $1-\cos$).}
\label{tab:gemma_drift_ans_mcq_free_cos}}
\end{table}

\begin{table}[h]
\added{\centering
\small
\begin{tabular}{lccc}
\toprule
\textbf{Perturbation} & \textbf{Pert. Drift} & \textbf{Ctrl. Drift} & \textbf{Cohen's $d$} \\
\midrule
Translation  & 0.045 & 0.394 & $-4.06$ \\
Pad/Crop     & 0.047 & 0.394 & $-4.04$ \\
Scale        & 0.035 & 0.394 & $-4.20$ \\
Scale+Pad    & 0.053 & 0.394 & $-3.64$ \\
Rotation     & 0.077 & 0.394 & $-3.48$ \\
TextOverlay  & 0.099 & 0.394 & $-2.98$ \\
RandomText   & 0.061 & 0.394 & $-3.64$ \\
BoxOverlay   & 0.042 & 0.394 & $-3.87$ \\
\bottomrule
\end{tabular}
\caption{Drift vs.\ control drift for \texttt{Gemma-3-4B-IT}, \texttt{ans\_open} regime (cosine distance, $1-\cos$). Open-ended drift (this table) is consistently larger than MCQ-conditioned drift (Table~\ref{tab:gemma_drift_ans_mcq_free_cos}), supporting that MCQ-conditioned analyses are a conservative lower bound.}
\label{tab:gemma_drift_ans_open_cos}}
\end{table}

\added{\subsubsection{Gemma-3-12B-IT on SEEDBench}}
\added{\label{appendix_gemma3_seedbench_12B}}
\added{For comparison with the 4B model (Tables~\ref{tab:gemma_fliprates}--\ref{tab:gemma_drift_ans_open_cos}), we provide the corresponding embedding-invariance and drift-vs-control statistics for Gemma-3-12B-IT on SEEDBench (200 samples for drift analysis, $k\!=\!8$ control pairs per image). Compared with the 4B model, the 12B model shows \emph{larger} representation drift across all regimes --- consistent with the observation that scale does not stabilise internal representations.}

\myadd{At the label level, the 12B model is also \emph{less} robust than the 4B: the union vulnerability rises from $0.471$ (4B) to $0.671$ (12B) and the TextOverlay instance-flip rate from $0.203$ to $0.316$ (Table~\ref{tab:gemma_seedbench_12B_fliprates}), despite higher base accuracy ($0.47\!\to\!0.53$). This behavioural scale-decoupling mirrors Qwen3-VL and parallels the representation-drift growth reported below.}

\begin{table}[h]
\myadd{\centering
\small
\begin{tabular}{lcc}
\toprule
\textbf{Perturbation} & \textbf{$\mathrm{\AVg}$ (flip-rate)} & \textbf{$\mathrm{\Ve}$ (vulnerability)} \\
\midrule
Translation & 0.145 & 0.355 \\
Pad/Crop    & 0.154 & 0.384 \\
Scale       & 0.113 & 0.113 \\
Scale+Pad   & 0.178 & 0.235 \\
Rotation    & 0.259 & 0.341 \\
TextOverlay & 0.316 & 0.463 \\
BoxOverlay  & 0.148 & 0.170 \\
RandomText  & 0.198 & 0.283 \\
\midrule
Any (union) & 0.179 & 0.671 \\
\bottomrule
\end{tabular}
\caption{ Label invariance for \texttt{Gemma-3-12B-IT} on SEEDBench (log-likelihood MCQ scoring, 1{,}000 samples). The 12B model is less robust than the 4B (Table~\ref{tab:gemma_fliprates}) across every perturbation, despite higher base accuracy.}
\label{tab:gemma_seedbench_12B_fliprates}}
\end{table}

\begin{table*}[t]
\added{\centering
\resizebox{\linewidth}{!}{%
\small
\begin{tabular}{lccccc}
\toprule
\textbf{Type} &
\textbf{ctx\_open (cos / L2)} &
\textbf{ctx\_mcq (cos / L2)} &
\textbf{ans\_open (cos / L2)} &
\textbf{ans\_mcq (cos / L2)} &
\textbf{ans\_mcq\_free (cos / L2)} \\
\midrule
Translation  & 0.973 / 66.08  & 0.994 / 33.76 & 0.967 / 35.34 & 0.972 / 29.14 & 0.967 / 23.83 \\
Pad/Crop     & 0.972 / 69.59  & 0.992 / 39.66 & 0.966 / 37.41 & 0.971 / 30.81 & 0.968 / 23.88 \\
Scale        & 0.984 / 52.39  & 0.995 / 30.25 & 0.976 / 29.38 & 0.979 / 23.88 & 0.977 / 19.30 \\
Scale+Pad    & 0.963 / 79.17  & 0.990 / 46.95 & 0.956 / 43.16 & 0.965 / 35.38 & 0.960 / 26.92 \\
Rotation     & 0.925 / 115.42 & 0.974 / 75.43 & 0.926 / 60.64 & 0.939 / 51.33 & 0.943 / 34.26 \\
TextOverlay  & 0.918 / 125.45 & 0.937 / 116.90 & 0.887 / 80.06 & 0.914 / 63.82 & 0.934 / 37.19 \\
RandomText   & 0.954 / 90.19  & 0.979 / 67.00 & 0.942 / 51.65 & 0.956 / 41.78 & 0.952 / 30.17 \\
BoxOverlay   & 0.982 / 56.93  & 0.993 / 37.96 & 0.977 / 31.21 & 0.969 / 31.05 & 0.965 / 23.76 \\
\midrule
Any          & \textbf{0.963 / 78.12} & \textbf{0.985 / 50.88} & \textbf{0.954 / 43.57} & \textbf{0.962 / 36.21} & \textbf{0.961 / 26.55} \\
\bottomrule
\end{tabular}}
\caption{Embedding invariance for \texttt{Gemma-3-12B-IT} on SEEDBench (mean similarity to base across five representation regimes, 200 samples).}
\label{tab:gemma_seedbench_12B_embinv}}
\end{table*}

\begin{table}[h]
\added{\centering
\small
\begin{tabular}{lccc}
\toprule
\textbf{Perturbation} & \textbf{Pert. Drift} & \textbf{Ctrl. Drift} & \textbf{Cohen's $d$} \\
\midrule
Translation  & 0.033 & 0.314 & $-7.88$ \\
Pad/Crop     & 0.032 & 0.314 & $-8.10$ \\
Scale        & 0.023 & 0.314 & $-7.86$ \\
Scale+Pad    & 0.040 & 0.314 & $-6.67$ \\
Rotation     & 0.057 & 0.314 & $-5.98$ \\
TextOverlay  & 0.066 & 0.314 & $-5.74$ \\
RandomText   & 0.048 & 0.314 & $-6.11$ \\
BoxOverlay   & 0.035 & 0.314 & $-6.67$ \\
\bottomrule
\end{tabular}
\caption{Drift vs.\ control drift for \texttt{Gemma-3-12B-IT} on SEEDBench, \texttt{ans\_mcq\_free} regime (cosine distance). Compared with 4B (Table~\ref{tab:gemma_drift_ans_mcq_free_cos}), drift is larger across nearly every perturbation, including TextOverlay ($0.066$ vs.\ $0.042$).}
\label{tab:gemma_seedbench_12B_drift_ans_mcq_free}}
\end{table}

\begin{table}[h]
\added{\centering
\small
\begin{tabular}{lccc}
\toprule
\textbf{Perturbation} & \textbf{Pert. Drift} & \textbf{Ctrl. Drift} & \textbf{Cohen's $d$} \\
\midrule
Translation  & 0.033 & 0.350 & $-3.69$ \\
Pad/Crop     & 0.035 & 0.350 & $-3.66$ \\
Scale        & 0.024 & 0.350 & $-3.66$ \\
Scale+Pad    & 0.044 & 0.350 & $-3.30$ \\
Rotation     & 0.074 & 0.350 & $-2.87$ \\
TextOverlay  & 0.112 & 0.350 & $-2.35$ \\
RandomText   & 0.058 & 0.350 & $-3.05$ \\
BoxOverlay   & 0.023 & 0.350 & $-3.83$ \\
\bottomrule
\end{tabular}
\caption{Drift vs.\ control drift for \texttt{Gemma-3-12B-IT} on SEEDBench, \texttt{ans\_open} regime (cosine distance). Open-ended drift exceeds MCQ-conditioned drift, confirming MCQ analyses are a conservative lower bound for the 12B model as well.}
\label{tab:gemma_seedbench_12B_drift_ans_open}}
\end{table}

\added{\subsubsection{Gemma-3 on MMMU: 4B vs.\ 12B}}
\added{\label{appendix_gemma3_mmmu}}
\added{We extend the Gemma-3 analysis to MMMU (847 validation samples) for both Gemma-3-4B-IT and Gemma-3-12B-IT. Tables~\ref{tab:gemma_mmmu_fliprates}--\ref{tab:gemma_mmmu_drift_ans_open_12B} show that, consistent with our Qwen3-VL scaling results (\cref{appendix_scaling_seedbench}), \emph{larger Gemma models exhibit higher --- not lower --- perturbation sensitivity}: Gemma-3-12B-IT shows substantially higher flip rates than Gemma-3-4B-IT under nearly every perturbation, with TextOverlay rising from $\mathrm{\AVg}\!=\!20.2\%$ ($\mathrm{\Ve}\!=\!34.0\%$) at 4B to $\mathrm{\AVg}\!=\!31.8\%$ ($\mathrm{\Ve}\!=\!45.6\%$) at 12B, and the union vulnerability rising from $44.3\%$ to $54.0\%$. The overlay ordering (semantic $>$ random $>$ empty) is preserved across both scales. Embedding-regime and drift-vs-control results, on the same 600-sample subset used in \cref{appendix_gemma3}, are reported in Tables~\ref{tab:gemma_mmmu_embinv_4B}--\ref{tab:gemma_mmmu_drift_ans_open_12B}.}

\begin{table}[h]
\added{\centering
\small
\resizebox{\linewidth}{!}{%
\begin{tabular}{lcccc}
\toprule
\textbf{Perturbation} & \multicolumn{2}{c}{\textbf{Gemma-3-4B-IT}} & \multicolumn{2}{c}{\textbf{Gemma-3-12B-IT}} \\
\cmidrule(lr){2-3}\cmidrule(lr){4-5}
                  & $\mathrm{\AVg}$ & $\mathrm{\Ve}$ & $\mathrm{\AVg}$ & $\mathrm{\Ve}$ \\
\midrule
Translation       & 0.079 & 0.198 & 0.088 & 0.234 \\
Pad/Crop          & 0.083 & 0.214 & 0.103 & 0.273 \\
Scale             & 0.055 & 0.055 & 0.071 & 0.071 \\
Scale+Pad         & 0.091 & 0.122 & 0.117 & 0.150 \\
Rotation          & 0.129 & 0.181 & 0.159 & 0.218 \\
TextOverlay       & \textbf{0.202} & \textbf{0.340} & \textbf{0.318} & \textbf{0.456} \\
RandomText        & 0.097 & 0.144 & 0.116 & 0.171 \\
BoxOverlay        & 0.053 & 0.063 & 0.065 & 0.079 \\
\midrule
Any (union)       & \textbf{0.094} & \textbf{0.443} & \textbf{0.121} & \textbf{0.540} \\
\bottomrule
\end{tabular}}
\caption{Label invariance for Gemma-3 on MMMU (847 samples). The 12B model is consistently \emph{less} robust than the 4B model across every perturbation type.}
\label{tab:gemma_mmmu_fliprates}}
\end{table}

\begin{table}[h]
\added{\centering
\small
\begin{tabular}{lcccc}
\toprule
\textbf{Perturbation} & \textbf{R$\to$W} & \textbf{W$\to$R} & \textbf{R$\to$R} & \textbf{W$\to$W} \\
\midrule
Translation     & 133  & 152  & 1579 & 4912 \\
Pad/Crop        & 146  & 164  & 1566 & 4900 \\
Scale           & 14   & 10   & 200  & 623 \\
Scale+Pad       & 44   & 37   & 384  & 1229 \\
Rotation        & 58   & 52   & 370  & 1214 \\
TextOverlay     & 142  & 125  & 431  & 1676 \\
RandomText      & 63   & 58   & 510  & 1743 \\
BoxOverlay      & 37   & 26   & 536  & 1775 \\
\midrule
Any             & \textbf{637} & \textbf{624} & \textbf{5576} & \textbf{18072} \\
\bottomrule
\end{tabular}
\caption{Prediction correctness transitions for \texttt{Gemma-3-4B-IT} on MMMU (instance level). W$\to$R counts are comparable to R$\to$W, confirming bidirectional movement across decision boundaries.}
\label{tab:gemma_mmmu_transitions_4B}}
\end{table}

\begin{table}[h]
\added{\centering
\small
\begin{tabular}{lcccc}
\toprule
\textbf{Perturbation} & \textbf{R$\to$W} & \textbf{W$\to$R} & \textbf{R$\to$R} & \textbf{W$\to$W} \\
\midrule
Translation     & 202  & 144  & 1510 & 4920 \\
Pad/Crop        & 231  & 168  & 1481 & 4896 \\
Scale           & 19   & 15   & 195  & 618 \\
Scale+Pad       & 66   & 39   & 362  & 1227 \\
Rotation        & 87   & 60   & 341  & 1206 \\
TextOverlay     & 230  & 182  & 352  & 1610 \\
RandomText      & 94   & 60   & 488  & 1732 \\
BoxOverlay      & 44   & 35   & 538  & 1757 \\
\midrule
Any             & \textbf{973} & \textbf{703} & \textbf{5267} & \textbf{17966} \\
\bottomrule
\end{tabular}
\caption{Prediction correctness transitions for \texttt{Gemma-3-12B-IT} on MMMU. Both error-injection (R$\to$W) and error-correction (W$\to$R) counts grow with scale, consistent with sharper but more sensitive decision boundaries.}
\label{tab:gemma_mmmu_transitions_12B}}
\end{table}

\begin{table}[h]
\added{\centering
\small
\begin{tabular}{lcc}
\toprule
\textbf{Perturbation} & \textbf{$\Delta E_{\text{dir}}$ (all)} & \textbf{$\Delta E_{\text{dir}}$ (flips)} \\
\midrule
Translation  & $-1.08 \pm 14.69$  & $-3.02 \pm 14.09$ \\
Pad/Crop     & $-0.49 \pm 20.83$  & $-3.64 \pm 26.99$ \\
Scale        & $\phantom{-}0.18 \pm 9.70$  & $\phantom{-}1.15 \pm 6.58$ \\
Scale+Pad    & $-5.30 \pm 24.23$  & $-6.56 \pm 26.54$ \\
Rotation     & $\phantom{-}17.66 \pm 47.18$  & $\phantom{-}14.01 \pm 52.08$ \\
TextOverlay  & $\phantom{-}0.29 \pm 29.77$  & $-2.60 \pm 25.58$ \\
RandomText   & $\phantom{-}25.22 \pm 36.28$  & $\phantom{-}25.91 \pm 36.74$ \\
BoxOverlay   & $-6.21 \pm 20.51$  & $-9.46 \pm 24.19$ \\
\bottomrule
\end{tabular}
\caption{Vision-token Dirichlet energy change for \texttt{Gemma-3-4B-IT} on MMMU. Flip-inducing instances generally show larger absolute deviations, consistent with the SEEDBench pattern.}
\label{tab:gemma_mmmu_dirichlet_4B}}
\end{table}

\begin{table}[h]
\added{\centering
\small
\begin{tabular}{lcc}
\toprule
\textbf{Perturbation} & \textbf{$\Delta E_{\text{dir}}$ (all)} & \textbf{$\Delta E_{\text{dir}}$ (flips)} \\
\midrule
Translation  & $-1.08 \pm 14.69$  & $-1.47 \pm 13.87$ \\
Pad/Crop     & $-0.49 \pm 20.83$  & $\phantom{-}0.00 \pm 24.03$ \\
Scale        & $\phantom{-}0.18 \pm 9.70$  & $\phantom{-}1.95 \pm 9.75$ \\
Scale+Pad    & $-5.30 \pm 24.23$  & $-3.29 \pm 25.06$ \\
Rotation     & $\phantom{-}17.66 \pm 47.18$  & $\phantom{-}17.88 \pm 45.85$ \\
TextOverlay  & $\phantom{-}0.37 \pm 29.73$  & $\phantom{-}0.00 \pm 26.83$ \\
RandomText   & $\phantom{-}25.15 \pm 36.19$  & $\phantom{-}23.77 \pm 33.02$ \\
BoxOverlay   & $-6.22 \pm 20.50$  & $-4.75 \pm 18.90$ \\
\bottomrule
\end{tabular}
\caption{Vision-token Dirichlet energy change for \texttt{Gemma-3-12B-IT} on MMMU. Patterns are consistent with the 4B model and with SEEDBench results, showing structural reorganisation invariant to model scale.}
\label{tab:gemma_mmmu_dirichlet_12B}}
\end{table}

\added{\paragraph{Embedding invariance and drift on MMMU.}
Tables~\ref{tab:gemma_mmmu_embinv_4B}--\ref{tab:gemma_mmmu_drift_ans_open_12B} report embedding invariance across the five representation regimes and drift vs.\ control-drift statistics for both Gemma-3-4B-IT and Gemma-3-12B-IT on MMMU (600-sample drift subset, $k\!=\!32$ control pairs per image). All key patterns from SEEDBench (Tables~\ref{tab:gemma_embinv}--\ref{tab:gemma_drift_ans_open_cos}) replicate: \texttt{TextOverlay} induces the largest drift, MCQ conditioning stabilises context embeddings, and open-ended drift consistently exceeds MCQ-conditioned drift. The 12B model exhibits visibly larger drift than 4B across nearly every perturbation and regime (\eg, \texttt{TextOverlay} in \texttt{ctx\_open}: $0.950$ at 4B vs.\ $0.926$ at 12B), again confirming the scale-decoupling pattern.}

\begin{table*}[t]
\added{\centering
\resizebox{\linewidth}{!}{%
\small
\begin{tabular}{lccccc}
\toprule
\textbf{Type} &
\textbf{ctx\_open (cos / L2)} &
\textbf{ctx\_mcq (cos / L2)} &
\textbf{ans\_open (cos / L2)} &
\textbf{ans\_mcq (cos / L2)} &
\textbf{ans\_mcq\_free (cos / L2)} \\
\midrule
Translation  & 0.981 / 59.27 & 0.996 / 24.58 & 0.932 / 55.90 & 0.988 / 19.89 & 0.974 / 27.09 \\
Pad/Crop     & 0.980 / 61.91 & 0.995 / 27.41 & 0.930 / 57.07 & 0.986 / 21.45 & 0.972 / 27.96 \\
Scale        & 0.989 / 46.00 & 0.998 / 19.43 & 0.946 / 46.69 & 0.991 / 15.53 & 0.981 / 22.24 \\
Scale+Pad    & 0.978 / 65.68 & 0.995 / 29.43 & 0.928 / 58.98 & 0.985 / 23.01 & 0.972 / 28.31 \\
Rotation     & 0.962 / 85.29 & 0.989 / 41.88 & 0.903 / 71.18 & 0.978 / 30.07 & 0.961 / 35.26 \\
TextOverlay  & 0.950 / 99.20 & 0.985 / 50.32 & 0.881 / 79.27 & 0.969 / 36.60 & 0.961 / 35.32 \\
RandomText   & 0.973 / 69.62 & 0.994 / 30.82 & 0.917 / 62.23 & 0.982 / 25.16 & 0.970 / 29.31 \\
BoxOverlay   & 0.986 / 45.28 & 0.997 / 19.07 & 0.946 / 43.95 & 0.991 / 15.04 & 0.981 / 20.13 \\
\midrule
Any          & \textbf{0.976 / 65.20} & \textbf{0.994 / 29.20} & \textbf{0.925 / 58.85} & \textbf{0.985 / 22.70} & \textbf{0.972 / 28.13} \\
\bottomrule
\end{tabular}}
\caption{Embedding invariance for \texttt{Gemma-3-4B-IT} on MMMU across five representation regimes (847 samples).}
\label{tab:gemma_mmmu_embinv_4B}}
\end{table*}

\begin{table*}[t]
\added{\centering
\resizebox{\linewidth}{!}{%
\small
\begin{tabular}{lccccc}
\toprule
\textbf{Type} &
\textbf{ctx\_open (cos / L2)} &
\textbf{ctx\_mcq (cos / L2)} &
\textbf{ans\_open (cos / L2)} &
\textbf{ans\_mcq (cos / L2)} &
\textbf{ans\_mcq\_free (cos / L2)} \\
\midrule
Translation  & 0.976 / 72.48 & 0.995 / 26.47 & 0.938 / 45.41 & 0.985 / 15.86 & 0.969 / 19.53 \\
Pad/Crop     & 0.973 / 76.25 & 0.993 / 30.75 & 0.934 / 47.31 & 0.984 / 17.39 & 0.967 / 20.64 \\
Scale        & 0.984 / 56.72 & 0.997 / 20.97 & 0.949 / 38.11 & 0.988 / 12.77 & 0.975 / 16.41 \\
Scale+Pad    & 0.970 / 79.96 & 0.991 / 34.15 & 0.930 / 49.87 & 0.981 / 19.26 & 0.963 / 22.39 \\
Rotation     & 0.942 / 111.20 & 0.978 / 52.16 & 0.902 / 62.57 & 0.970 / 26.22 & 0.952 / 26.65 \\
TextOverlay  & 0.926 / 130.31 & 0.955 / 77.17 & 0.866 / 76.43 & 0.947 / 37.32 & 0.939 / 31.08 \\
RandomText   & 0.963 / 86.10 & 0.988 / 38.90 & 0.927 / 51.37 & 0.977 / 22.26 & 0.963 / 22.17 \\
BoxOverlay   & 0.982 / 54.18 & 0.996 / 21.19 & 0.952 / 35.78 & 0.987 / 12.82 & 0.978 / 14.62 \\
\midrule
Any          & \textbf{0.967 / 81.18} & \textbf{0.989 / 35.24} & \textbf{0.928 / 49.76} & \textbf{0.979 / 19.47} & \textbf{0.965 / 21.29} \\
\bottomrule
\end{tabular}}
\caption{Embedding invariance for \texttt{Gemma-3-12B-IT} on MMMU. Compared with the 4B model (Table~\ref{tab:gemma_mmmu_embinv_4B}), \texttt{ctx\_open} similarity is consistently lower across all perturbations, confirming that the larger model exhibits greater representation drift.}
\label{tab:gemma_mmmu_embinv_12B}}
\end{table*}

\begin{table}[h]
\added{\centering
\small
\begin{tabular}{lccc}
\toprule
\textbf{Perturbation} & \textbf{Pert. Drift} & \textbf{Ctrl. Drift} & \textbf{Cohen's $d$} \\
\midrule
Translation  & 0.026 & 0.422 & $-9.07$ \\
Pad/Crop     & 0.027 & 0.422 & $-9.02$ \\
Scale        & 0.019 & 0.422 & $-9.06$ \\
Scale+Pad    & 0.027 & 0.422 & $-8.69$ \\
Rotation     & 0.038 & 0.422 & $-7.85$ \\
TextOverlay  & 0.038 & 0.422 & $-7.94$ \\
RandomText   & 0.028 & 0.422 & $-8.52$ \\
BoxOverlay   & 0.018 & 0.422 & $-8.86$ \\
\bottomrule
\end{tabular}
\caption{Drift vs.\ control drift for \texttt{Gemma-3-4B-IT} on MMMU, \texttt{ans\_mcq\_free} regime (cosine distance).}
\label{tab:gemma_mmmu_drift_ans_mcq_free_4B}}
\end{table}

\begin{table}[h]
\added{\centering
\small
\begin{tabular}{lccc}
\toprule
\textbf{Perturbation} & \textbf{Pert. Drift} & \textbf{Ctrl. Drift} & \textbf{Cohen's $d$} \\
\midrule
Translation  & 0.031 & 0.479 & $-10.01$ \\
Pad/Crop     & 0.033 & 0.479 & $-9.80$ \\
Scale        & 0.025 & 0.479 & $-9.46$ \\
Scale+Pad    & 0.037 & 0.479 & $-8.66$ \\
Rotation     & 0.048 & 0.479 & $-8.29$ \\
TextOverlay  & 0.061 & 0.479 & $-7.54$ \\
RandomText   & 0.037 & 0.479 & $-8.84$ \\
BoxOverlay   & 0.022 & 0.479 & $-9.90$ \\
\bottomrule
\end{tabular}
\caption{Drift vs.\ control drift for \texttt{Gemma-3-12B-IT} on MMMU, \texttt{ans\_mcq\_free} regime (cosine distance). \texttt{TextOverlay} drift increases relative to 4B ($0.061$ vs.\ $0.038$), tracking the higher flip rate at 12B.}
\label{tab:gemma_mmmu_drift_ans_mcq_free_12B}}
\end{table}

\begin{table}[h]
\added{\centering
\small
\begin{tabular}{lccc}
\toprule
\textbf{Perturbation} & \textbf{Pert. Drift} & \textbf{Ctrl. Drift} & \textbf{Cohen's $d$} \\
\midrule
Translation  & 0.067 & 0.610 & $-8.69$ \\
Pad/Crop     & 0.069 & 0.610 & $-9.07$ \\
Scale        & 0.052 & 0.610 & $-8.27$ \\
Scale+Pad    & 0.071 & 0.610 & $-7.97$ \\
Rotation     & 0.096 & 0.610 & $-6.54$ \\
TextOverlay  & 0.118 & 0.610 & $-5.04$ \\
RandomText   & 0.082 & 0.610 & $-6.87$ \\
BoxOverlay   & 0.051 & 0.610 & $-8.01$ \\
\bottomrule
\end{tabular}
\caption{Drift vs.\ control drift for \texttt{Gemma-3-4B-IT} on MMMU, \texttt{ans\_open} regime (cosine distance). Open-ended drift is consistently larger than MCQ-conditioned drift, confirming the MCQ regime is a conservative lower bound.}
\label{tab:gemma_mmmu_drift_ans_open_4B}}
\end{table}

\begin{table}[h]
\added{\centering
\small
\begin{tabular}{lccc}
\toprule
\textbf{Perturbation} & \textbf{Pert. Drift} & \textbf{Ctrl. Drift} & \textbf{Cohen's $d$} \\
\midrule
Translation  & 0.062 & 0.601 & $-7.82$ \\
Pad/Crop     & 0.066 & 0.601 & $-7.72$ \\
Scale        & 0.051 & 0.601 & $-7.19$ \\
Scale+Pad    & 0.070 & 0.601 & $-6.84$ \\
Rotation     & 0.098 & 0.601 & $-6.01$ \\
TextOverlay  & 0.132 & 0.601 & $-4.51$ \\
RandomText   & 0.071 & 0.601 & $-6.72$ \\
BoxOverlay   & 0.046 & 0.601 & $-7.53$ \\
\bottomrule
\end{tabular}
\caption{Drift vs.\ control drift for \texttt{Gemma-3-12B-IT} on MMMU, \texttt{ans\_open} regime (cosine distance). \texttt{TextOverlay} drift rises to $0.132$ (Cohen's $d\!=\!-4.51$), the largest open-ended drift observed in the Gemma-3 evaluations.}
\label{tab:gemma_mmmu_drift_ans_open_12B}}
\end{table}

\section{Semantic Robustness and Error Asymmetry on POPE}
\label{appendix_POPE}
We analyze robustness on POPE (Adversarial) to examine the relationship between visual instability and hallucination.
Unlike multiple-choice reasoning tasks, POPE poses a binary decision problem with a highly asymmetric label space. In such settings, representation drift does not induce random label switching.
Instead, misaligned or unfamiliar visual representations tend to collapse predictions toward
the model’s language prior, which in POPE is strongly biased toward negative (“No”) responses.
This allows us to study how the same underlying representational instability
manifests differently depending on task geometry.

\myadd{\paragraph{POPE margin (flip margin).} For the binary POPE task we define the margin as the signed log-likelihood gap $s_{\mathrm{Yes}}-s_{\mathrm{No}}$: positive when the model favours ``Yes'' (object present) and negative when it favours ``No''. This margin is signed by construction and can therefore be negative---as in the average flip margin of ${\approx}-0.96$ reported in \cref{sec:exp_pope}---in contrast to the non-negative predicted-option margin of \cref{eq:margin} used for the multiple-choice tasks. A negative shift under perturbation thus quantifies movement toward the ``No'' language prior (the drift-to-prior effect), rather than a change in the argmax-based confidence gap.}

Having established that representation drift can both degrade reasoning performance
and, in some cases, suppress hallucinations, we now provide a deeper analysis of
\emph{semantic robustness} using the POPE benchmark.
Unlike SEEDBench and MMMU, POPE focuses on object existence verification and enables
fine-grained analysis of asymmetric error dynamics under perturbations.

All experiments in this section are conducted on the adversarial split of POPE using
\texttt{Qwen3-VL-2B} and \texttt{Qwen3-VL-8B} Instruct models.
Perturbations are applied exclusively to the image, while the semantic content of
the query remains unchanged.

\subsection{Overall Semantic Stability Under Perturbations}
\label{subsec:pope_overall}

\begin{figure*}[t]
\centering
\includegraphics[width=0.48\linewidth]{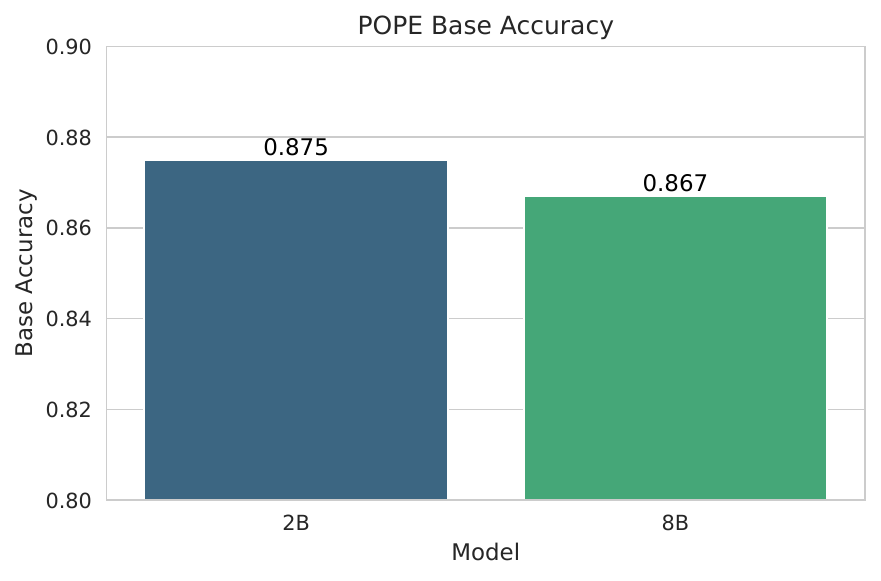} \hfill
\includegraphics[width=0.48\linewidth]{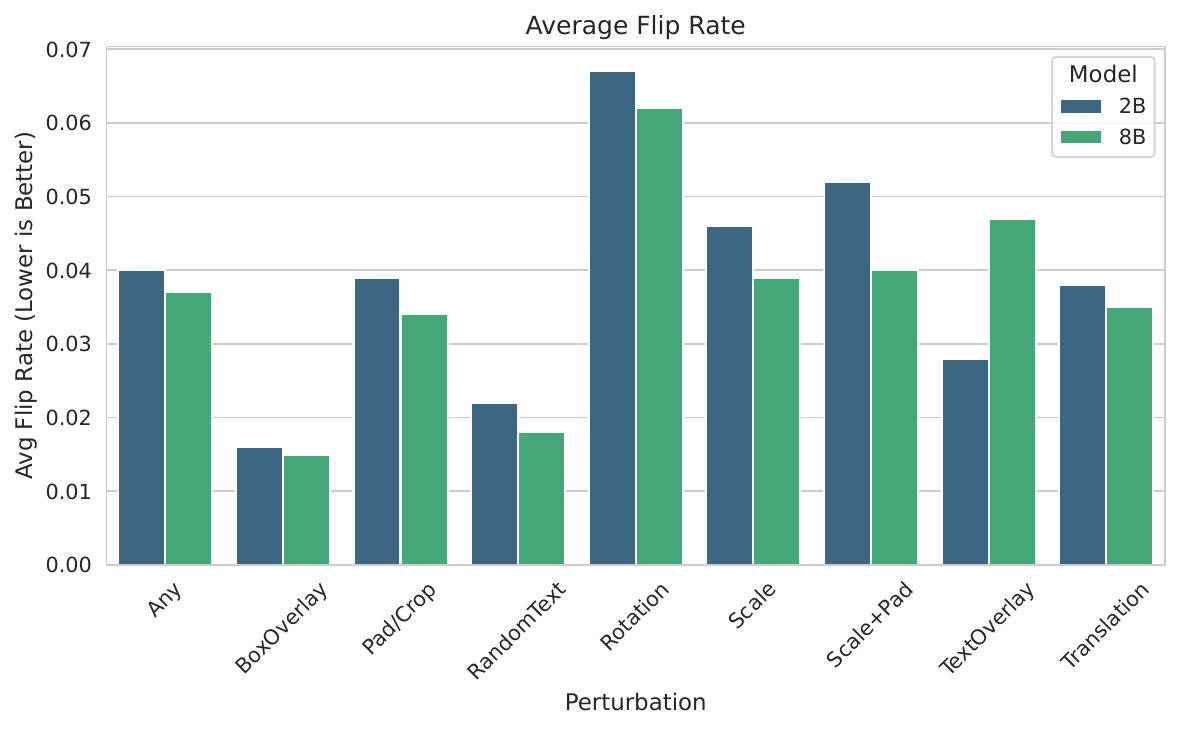}
\caption{
POPE adversarial split.
\textbf{Left}: base accuracy (higher is better).
\textbf{Right}: average flip rate under perturbations (lower is better).
Semantic robustness does not improve monotonically with model scale.
}
\label{fig:pope_overall}
\end{figure*}

Figure~\ref{fig:pope_overall} reports base accuracy and average flip rates across
perturbation types.
Although both models achieve high base accuracy on clean images, natural perturbations
induce substantial semantic instability.
Rotation consistently emerges as the most disruptive transformation, followed by
text overlays and scale-related perturbations.

Notably, increased model capacity does not guarantee improved semantic robustness:
the 8B model exhibits comparable or higher flip rates than the 2B model for several
perturbations, reinforcing the accuracy--robustness decoupling observed earlier.

\subsection{Asymmetric Error Dynamics}
\label{subsec:pope_asymmetry}

A key strength of POPE is its ability to distinguish asymmetric semantic errors.
We analyse four complementary transition types:
(i) true positives flipping to false negatives (TP $\rightarrow$ FN),
(ii) true negatives flipping to false positives (TN $\rightarrow$ FP),
(iii) correction of hallucinations (FP $\rightarrow$ TN),
and (iv) recovery of missed detections (FN $\rightarrow$ TP).

\begin{figure*}[t]
\centering
\includegraphics[width=0.48\linewidth]{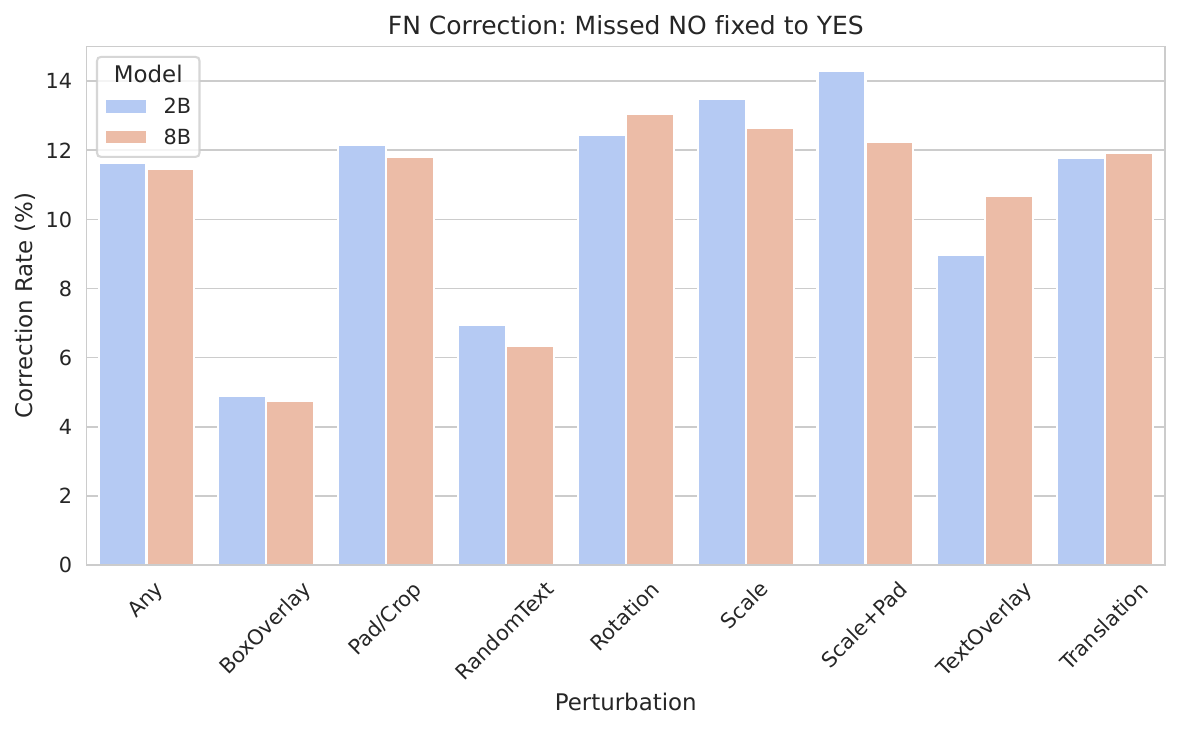}\hfill
\includegraphics[width=0.48\linewidth]{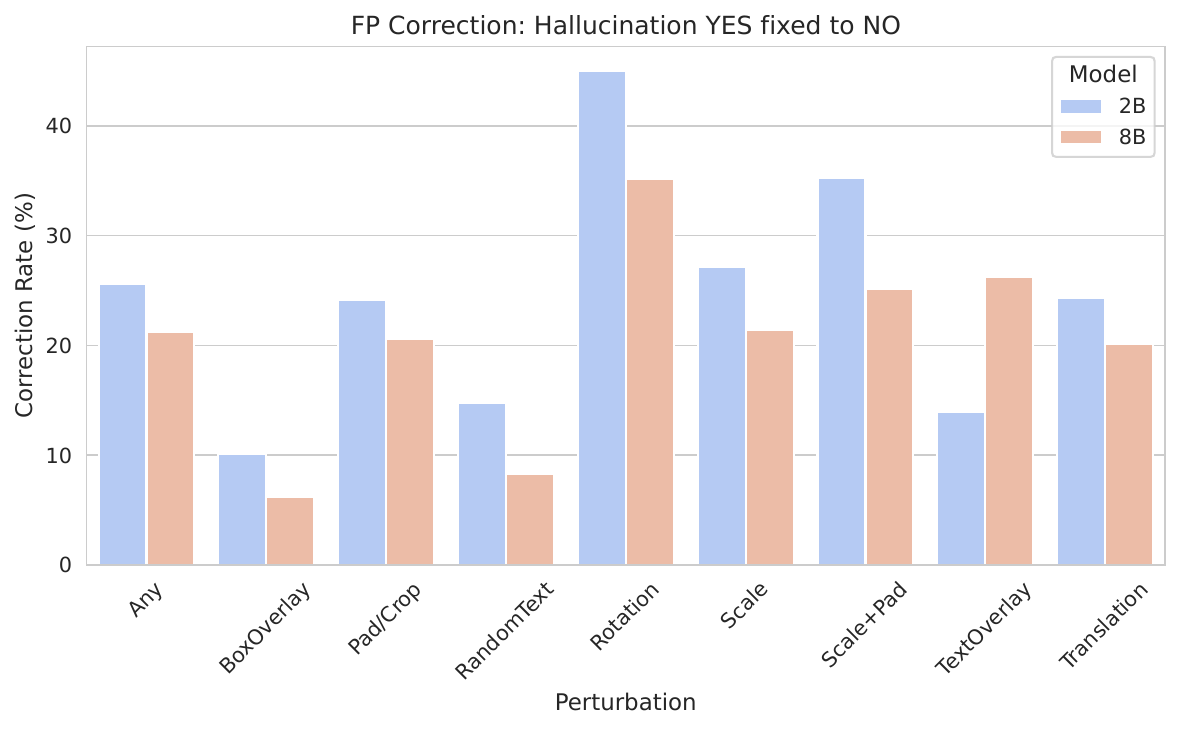}
\includegraphics[width=0.48\linewidth]{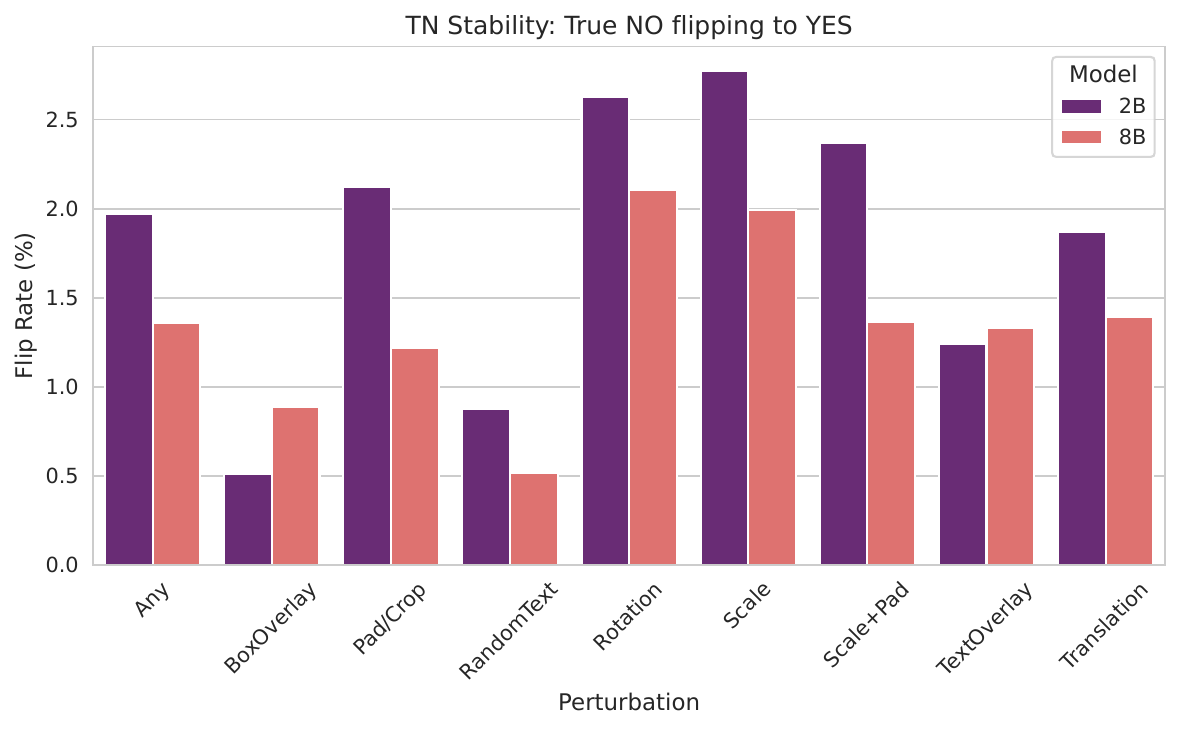} \hfill
\includegraphics[width=0.48\linewidth]{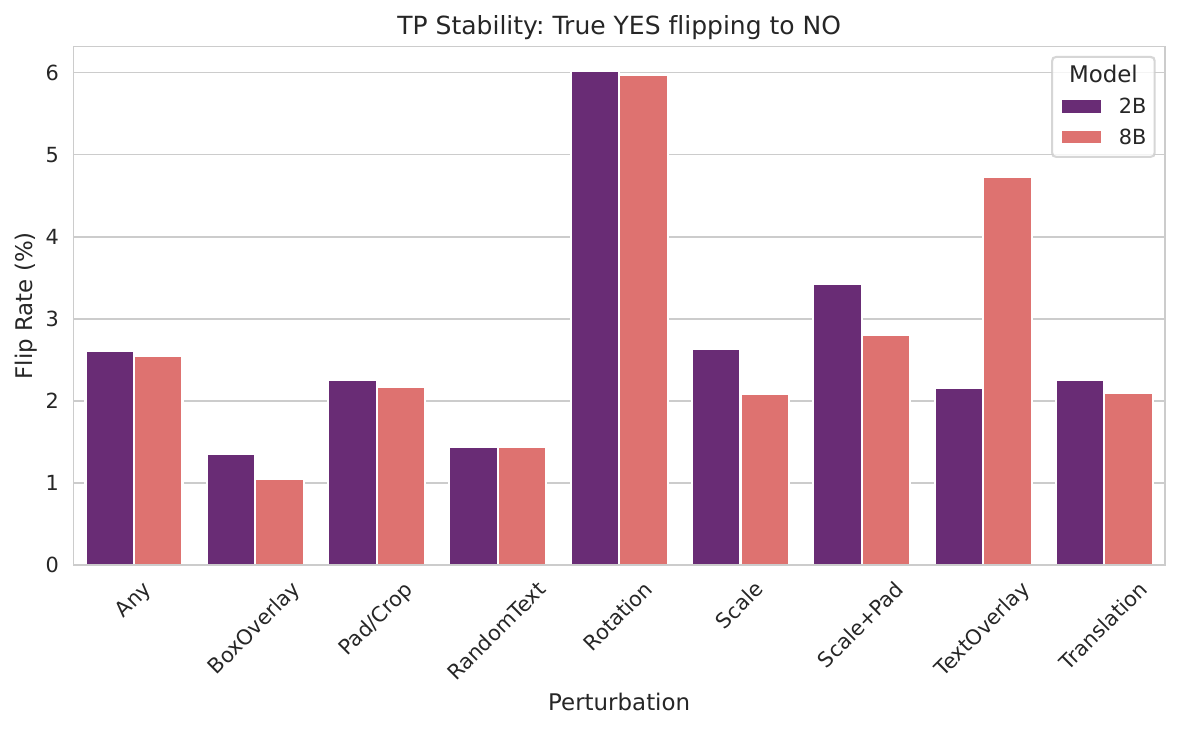}
\caption{
Asymmetric semantic error dynamics on POPE under natural perturbations.
Top: instability of correct predictions.
Bottom: perturbation-induced correction of semantic errors.
}
\label{fig:pope_error_dynamics}
\end{figure*}

Figure~\ref{fig:pope_error_dynamics} reveals pronounced asymmetry.
Perturbations such as rotation and scale predominantly increase TP $\rightarrow$ FN
errors, indicating that correct affirmative detections are fragile under visual
transformations.
In contrast, text overlays and crop-based perturbations more frequently induce
TN $\rightarrow$ FP transitions, increasing hallucination rates.

Importantly, some perturbations also \emph{reduce} hallucinations:
non-zero FP $\rightarrow$ TN correction rates indicate that perturbations can
accidentally suppress false positives, consistent with the drift-to-prior behaviour
identified earlier.

\subsection{Representation Drift and Smoothness Under Semantic Perturbations}
\label{subsec:pope_representation}

\begin{figure*}[t]
\centering
\includegraphics[width=0.48\linewidth]{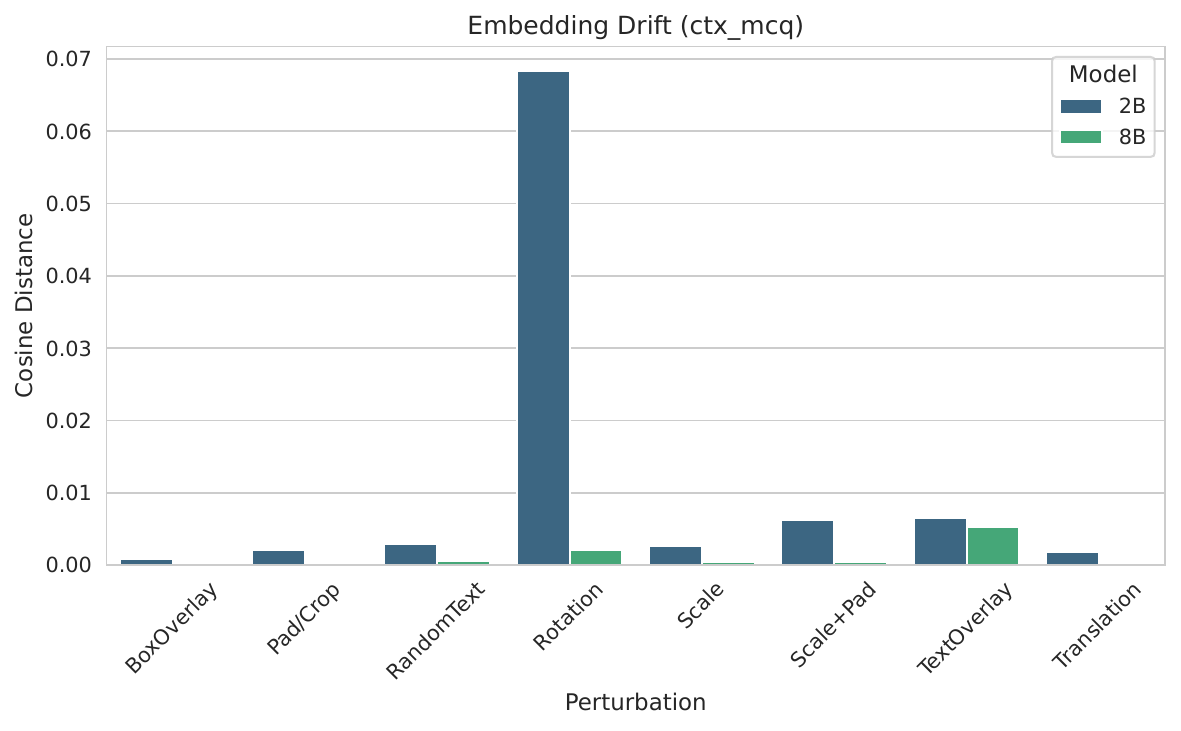}\hfill
\includegraphics[width=0.48\linewidth]{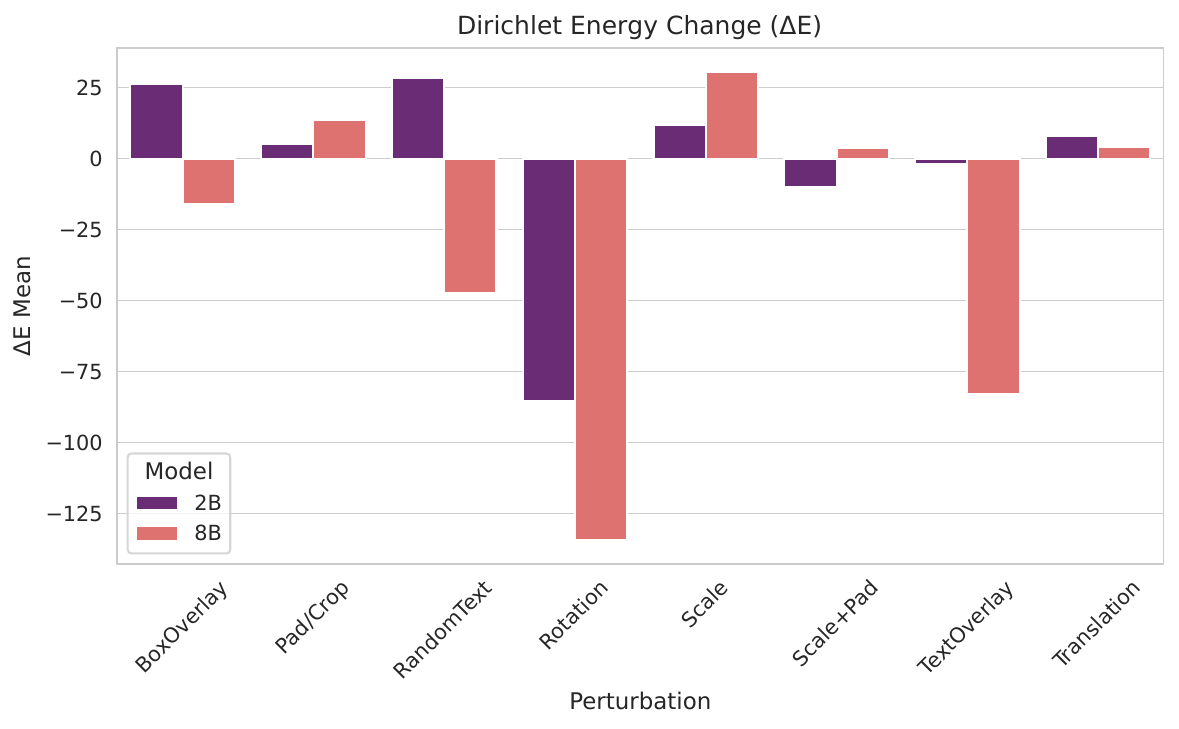}
\caption{
Representation-level effects on POPE.
\textbf{Left}: embedding drift.
\textbf{Right}: change in Dirichlet energy.
Semantic instability aligns with representation drift and structural disruption.
}
\label{fig:pope_representation}
\end{figure*}

To connect semantic instability with internal behaviour, we measure embedding drift
and changes in Dirichlet energy under perturbations.
Figure~\ref{fig:pope_representation} shows that perturbations inducing higher semantic
flip rates also exhibit larger embedding displacement and greater changes in the token
smoothness.

Rotation induces the largest embedding drift and the strongest decrease in Dirichlet
energy, aligning with its dominant impact on TP $\rightarrow$ FN errors.
Text overlays, while visually localised, produce consistent representation shifts,
explaining their disproportionate effect on hallucination dynamics.

These results extend our earlier analysis by showing that semantic robustness failures
are tightly coupled to both global representation drift and local structural
reorganization, even when high-level image semantics appear preserved.

\section{Vision-Token Smoothness and Dirichlet Energy}
\label{appendix_dirichlet}

Embedding drift captures \emph{global} movement in representation space, but it does not directly
characterise how visual information is organised \emph{locally} within the vision encoder.
To study structural stability at the token level, we analyse the \emph{Dirichlet energy} of vision
tokens arranged on their spatial grid.

This provides a complementary diagnostic of robustness:
embedding drift measures \emph{where} representations move in latent space,
whereas Dirichlet energy measures \emph{how} visual features are spatially structured and smoothed
across neighbouring tokens.
Together, these views allow us to distinguish global representation shifts from local structural
reorganization.

\subsection{Dirichlet Energy on Vision Tokens}
Following the general definition introduced in Section~\ref{methodology and experiment}, we instantiate the Dirichlet energy on the spatial grid of vision tokens to quantify
local structural organization within the vision encoder. Dirichlet energy quantifies spatial smoothness;
low values correspond to locally consistent token representations,
while high values indicate sharp spatial variation or token misalignment.
For each perturbation instance, we compute the change in Dirichlet energy
\begin{equation}
\Delta {E}_{\text{dir}} = {E}_{\text{dir}}(x') - {E}_{\text{dir}}(x),
\end{equation}
where $x$ is the base image and $x'$ its perturbed version.

\subsection{Dirichlet Energy Distributions}

\begin{figure*}[t]
\centering
\includegraphics[width=0.49\linewidth]{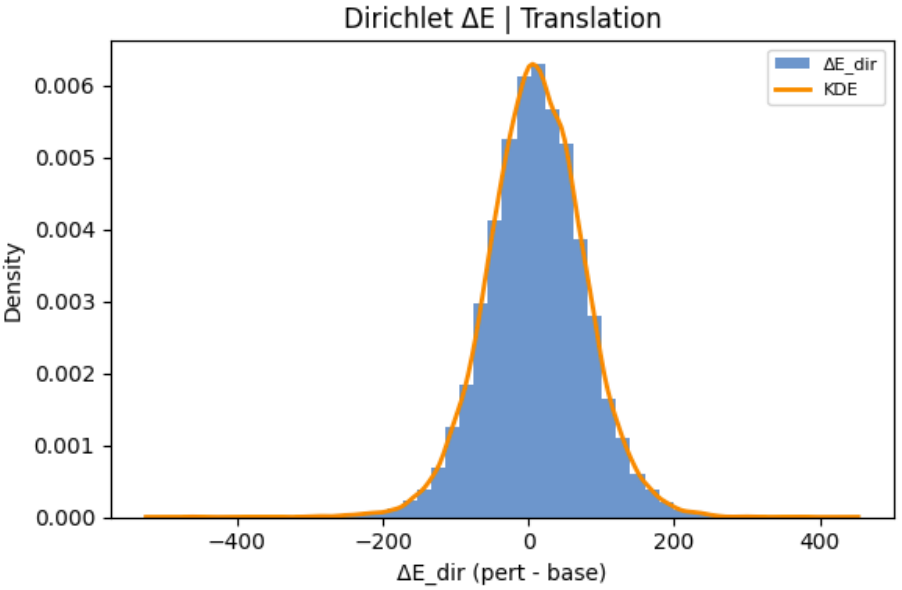} \hfill
\includegraphics[width=0.49\linewidth]{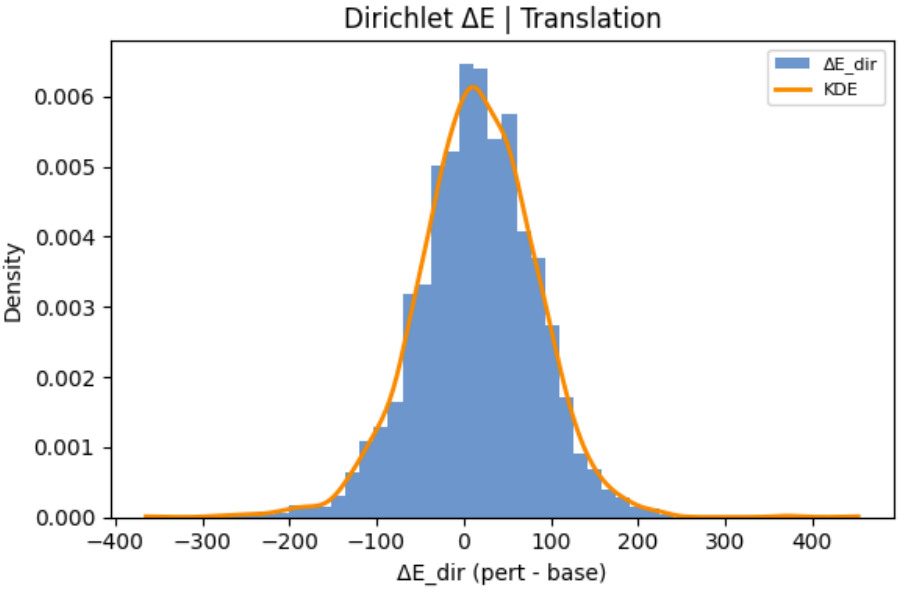}
\caption{
Dirichlet energy change $\Delta{E}_{\text{dir}}$ under Translation.
Left: all perturbation instances. Right: instances that induce label flips.
}
\label{fig:dirichlet_translation}
\end{figure*}

\begin{figure*}[t]
\centering
\includegraphics[width=0.48\linewidth]{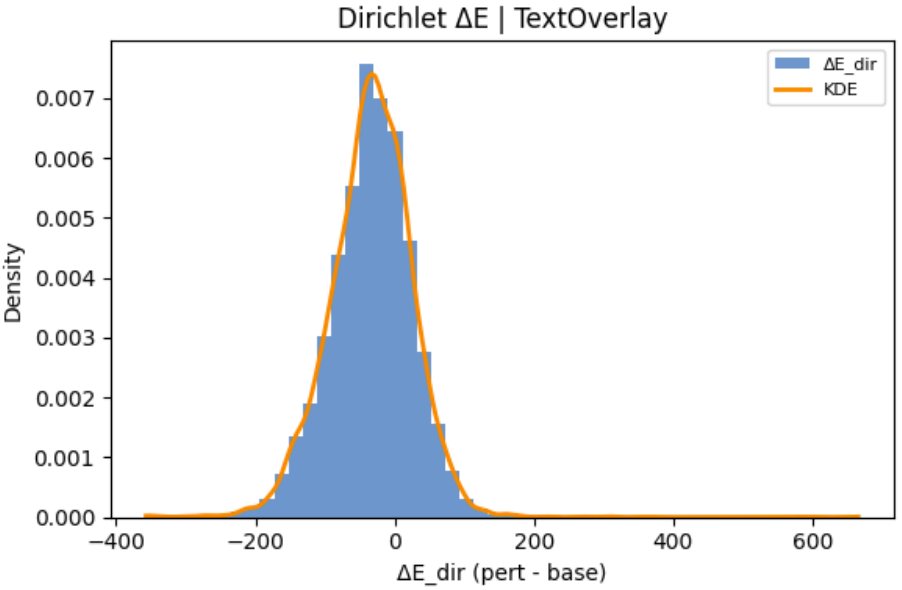} \hfill
\includegraphics[width=0.48\linewidth]{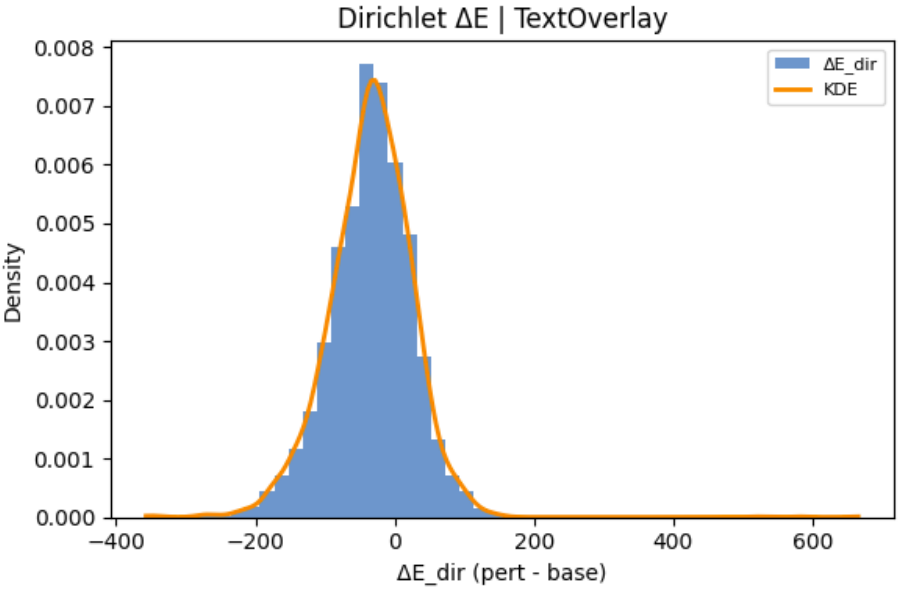}
\caption{
Dirichlet energy change $\Delta{E}_{\text{dir}}$ under TextOverlay.
Left: all perturbation instances. Right: instances that induce label flips.
}
\label{fig:dirichlet_textoverlay}
\end{figure*}

Figures~\ref{fig:dirichlet_translation} and~\ref{fig:dirichlet_textoverlay} show the distribution of
$\Delta {E}_{\text{dir}}$ for Translation and TextOverlay perturbations, respectively.
For each perturbation, we report distributions over all instances and over the subset of instances
that induce label flips.

\begin{table}[t]
\centering
\small
\begin{tabular}{lcc}
\toprule
Perturbation &
$\Delta {E}_{\text{dir}}$ (all) &
$\Delta {E}_{\text{dir}}$ (flips) \\
\midrule
Translation & $10.34 \pm 67.49$ & $15.26 \pm 68.56$ \\
Pad/Crop    & $8.65 \pm 70.13$  & $11.76 \pm 71.31$ \\
Scale       & $20.70 \pm 82.47$ & $21.02 \pm 88.59$ \\
Scale+Pad   & $9.86 \pm 82.29$  & $14.38 \pm 80.81$ \\
TextOverlay & $-33.87 \pm 60.14$& $-34.27 \pm 61.45$ \\
Rotation    & $-72.73 \pm 99.95$& $-66.76 \pm 101.21$ \\
\bottomrule
\end{tabular}
\caption{
Dirichlet energy change $\Delta {E}_{\text{dir}}$ (mean $\pm$ std).
}
\label{tab:dirichlet_summary}
\end{table}

Translation exhibits a small positive mean shift in $\Delta {E}_{\text{dir}}$
($10.34 \pm 67.49$), accompanied by substantial variance.
This indicates that translations do not uniformly smooth or disrupt spatial structure, but instead
introduce heterogeneous local misalignment.
Flip-inducing translation instances show slightly larger positive shifts, consistent with
phase-driven token misalignment rather than large-magnitude distortions. In contrast, TextOverlay produces a qualitatively different pattern.
The mean $\Delta {E}_{\text{dir}}$ is negative ($-33.87 \pm 60.14$), indicating a systematic
reorganization of local token neighbourhoods.
This shift occurs for both flip and non-flip instances, suggesting that overlays alter spatial
structure even when the final prediction remains unchanged.

Unlike geometric perturbations, which primarily redistribute spatial alignment, text overlays
introduce sharp edges and high-contrast strokes that interact directly with the patch grid.
The resulting effect is not simple displacement but a restructuring of local token relationships.

\subsection{Dirichlet Energy and Decision Instability}

Table~\ref{tab:dirichlet_summary} summarises mean $\Delta {E}_{\text{dir}}$ across all
perturbation types, reported separately for all instances and for flip-inducing instances. Across perturbation families, flip-inducing instances are associated with larger absolute Dirichlet deviations, indicating more pronounced spatial reorganization of vision tokens, rather than providing a standalone predictor of failure.

Dirichlet deviations than non-flip instances.
This indicates that perturbations that substantially reorganise local token structure, either
sharpening or smoothing relative to the base, are more likely to trigger decision boundary crossings.

Specifically, Dirichlet energy does not serve as a universal failure threshold; rather, it functions as a proxy for structural reorganization.
Label flips are associated with significant changes in the spatial relationships between vision
tokens, even when global embedding drift remains moderate.

\begin{figure*}[t]
\centering
\includegraphics[width=0.48\linewidth]{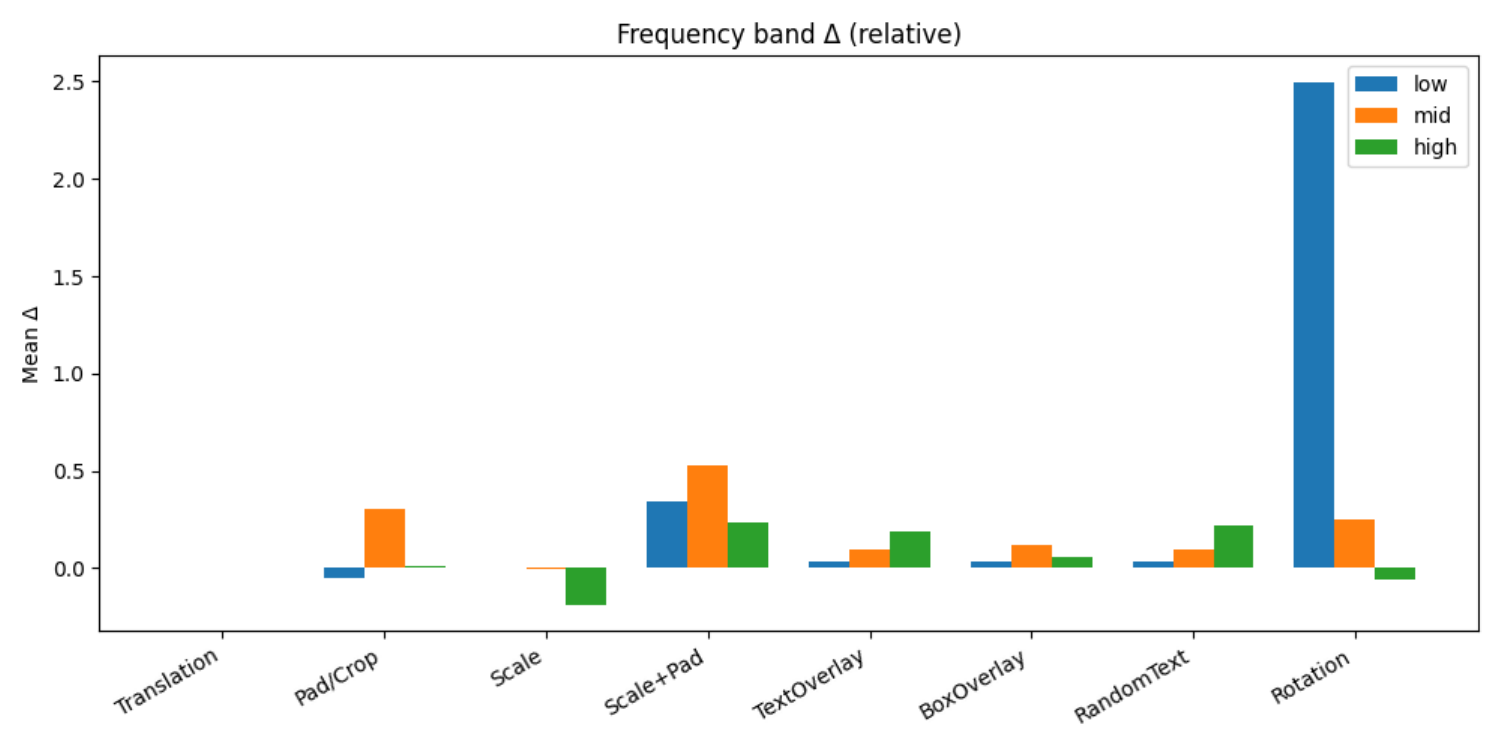}
\includegraphics[width=0.48\linewidth]{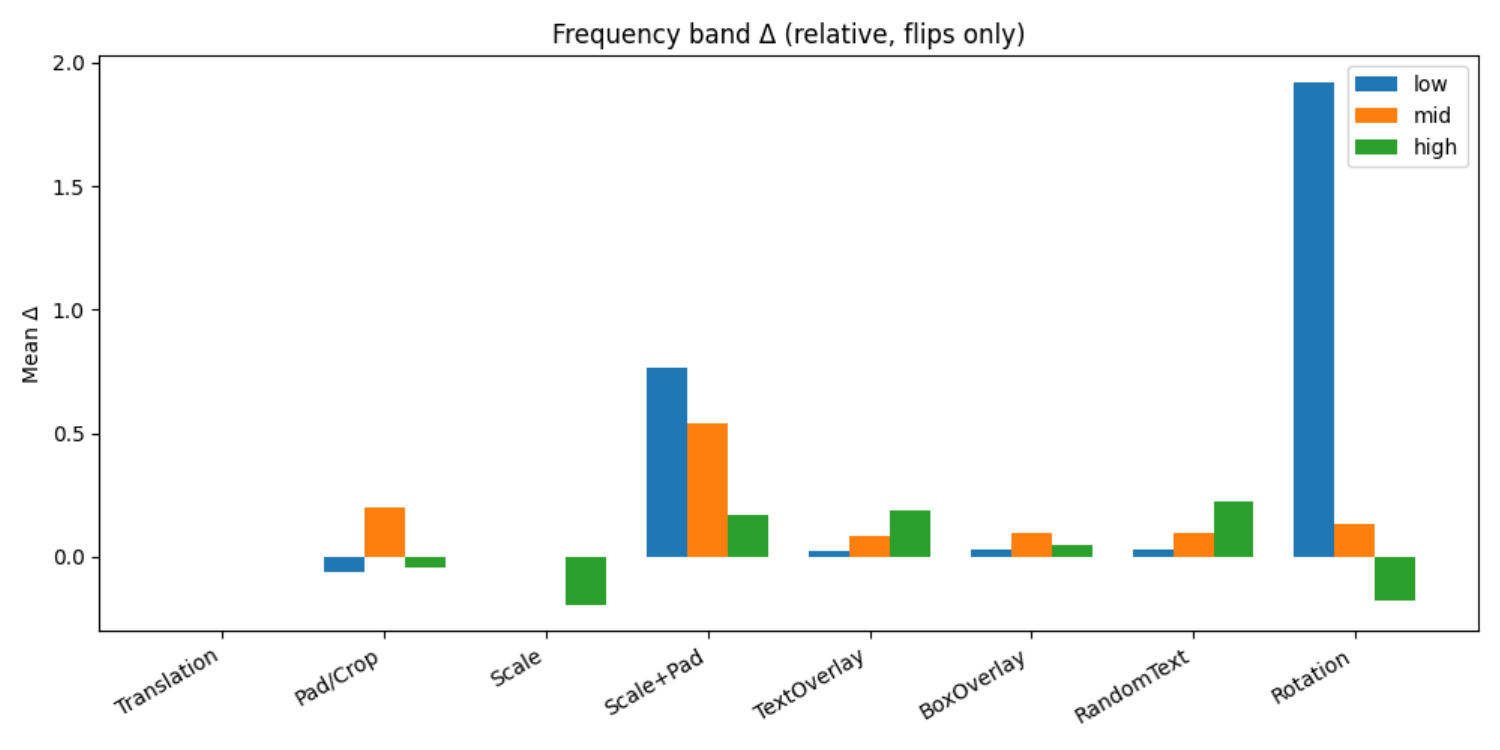}
\caption{
Frequency-band drift induced by perturbation variants.
\textbf{Left:} Average over all perturbations instances. \textbf{Right:} average over flip-inducing perturbation instances only.
}
\label{fig:overlay_band_energy}
\end{figure*}

Many natural perturbations are not defined in the frequency domain, yet they systematically
alter frequency content or phase relationships after interpolation, resampling, and discretization.
Thus we also examined correlations between Dirichlet energy change, embedding drift, and frequency-band energy
shifts. Across perturbation types, correlations between $\Delta {E}_{\text{dir}}$ and embedding
drift are modest but consistently non-zero, and are stronger for flip-inducing instances.
For example for flip inducing instances:
\begin{itemize}
    \item Translation: $\mathrm{corr}(\Delta {E}_{\text{dir}},\text{drift}) = 0.057$
    \item Scale: $\mathrm{corr}(\Delta {E}_{\text{dir}},\text{drift}) = 0.149$
    \item Rotation: $\mathrm{corr}(\Delta {E}_{\text{dir}},\text{drift}) = 0.065$
\end{itemize}
Importantly, Dirichlet energy is not intended as a superior predictor of failures compared to embedding drift.
Rather, it provides complementary structural evidence that global drift is accompanied by
localised reorganization of visual tokens, which is invisible to pooled embedding metrics. While these correlations do not imply direct causality, they indicate that structural token
reorganization and global representation drift are related but distinct phenomena.
Perturbation-specific frequency (Figure~\ref{fig:overlay_band_energy}) effects also align with Dirichlet trends, interpolation-based transformations amplify low-frequency components, while text overlays inject
broadband energy, consistent with both observed Dirichlet shifts and embedding drift.

\subsection{Structural Drift Complements Representation Drift}

Dirichlet energy provides a complementary diagnostic to embedding-based analyses:
\begin{itemize}
    \item Embedding drift captures \emph{where} representations move in latent space.
    \item Dirichlet energy captures \emph{how} spatial structure within vision tokens is reorganised.
\end{itemize}

Together, these results reinforce our central thesis:
robustness failures in VLMs arise from \emph{structural and spectral drift of vision tokens} that
misalign visual representations with language-conditioned decision boundaries.
This drift can accumulate even when output predictions remain unchanged, exposing a hidden reduction
in stability and decision margin.

\section{Detailed Frequency-Aware Robustness Suite}
\label{app:freq_suite}
This appendix details the experimental setup and full results for the frequency analysis summarised in Section~\ref{sec:freq_analysis}. This section is not intended as an independent robustness benchmark.
Rather, it serves as a controlled hypothesis test of the spectral-drift interpretation motivated by the natural perturbation and Dirichlet analyses (Sections~\ref{sec:embedding_drift}).
We move beyond geometric transformations and explicitly probe the role of \emph{frequency content} in VLM robustness.

Our objective is to disentangle two competing hypotheses:

\begin{itemize}
    \item \textbf{H1 (Low-frequency dominance).}
    VLM decisions rely primarily on low-frequency visual content; high-frequency perturbations should therefore have limited impact \citep{H11,H12,H13}.

    \item \textbf{H2 (Cross-frequency sensitivity and interaction).}
    Both low- and high-frequency components contribute to decision-making, and robustness failures arise from \emph{frequency-space drift} that misalign vision tokens with language-conditioned decision boundaries.
\end{itemize}

Unless stated otherwise, experiments in this section are conducted on a representative model
(\texttt{Qwen3-VL-2B}) and aggregated over 400 \textbf{SEEDBench} samples.

\subsection{Random Band-Limited Noise}
\label{subsec:freq_random_noise}

Given an image $x$, we sample i.i.d.\ Gaussian noise $\eta$ and project it into a frequency subspace using a radial Fourier mask:
low-pass (L), high-pass (H), or full-spectrum (A).
We construct perturbed inputs
\begin{equation}
x' = \mathrm{clip}\big(x + \epsilon \cdot \eta_{\text{band}}\big),
\end{equation}
where $\epsilon \in \{1/255, 2/255, 4/255, 8/255, 16/255\}$.
For each $\epsilon$, multiple trials are performed per image and statistics are averaged across all samples.

\begin{figure*}[t]
\centering
\begin{minipage}{0.48\linewidth}
    \centering
    \includegraphics[width=\linewidth]{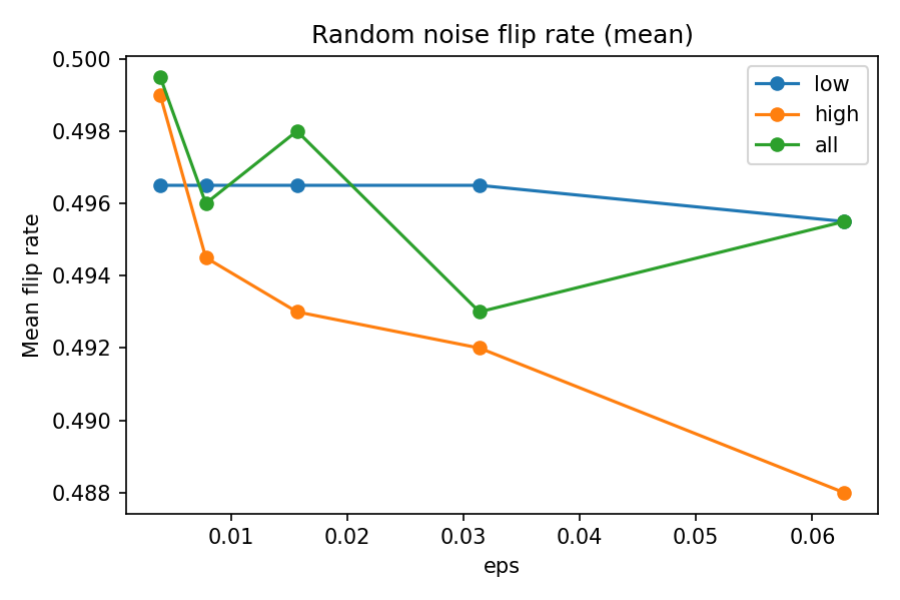}
    \caption{Mean flip rate under random band-limited noise, averaged over 400 samples. 
    Low-frequency (L), high-frequency (H), and full-spectrum (A) noise all induce comparable instability, rejecting a purely low-frequency bias explanation.}
    \label{fig:freq_rand_noise_flip}
\end{minipage}\hfill
\begin{minipage}{0.48\linewidth}
    \centering
    \includegraphics[width=0.75\linewidth]{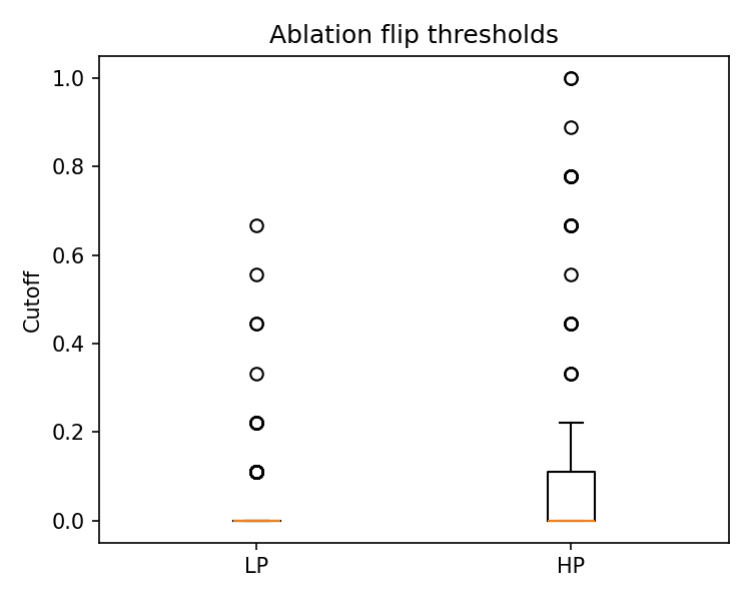}
    \caption{Distributions of flip thresholds under frequency ablation.
    Both low-pass and high-pass removal induce failures, supporting a
    cross-frequency reliance rather than a single-band bias.}
    \label{fig:freq_ablation_thresholds}
\end{minipage}
\end{figure*}

Figure~\ref{fig:freq_rand_noise_flip} reports mean flip rates as a function of $\epsilon$.
Flip rates remain high (approximately $50\%$) across all frequency bands, including low-frequency-only noise.
This directly contradicts a purely low-frequency dominance explanation and indicates sensitivity across the spectrum.

\begin{figure*}[t]
\centering
\includegraphics[width=0.48\linewidth]{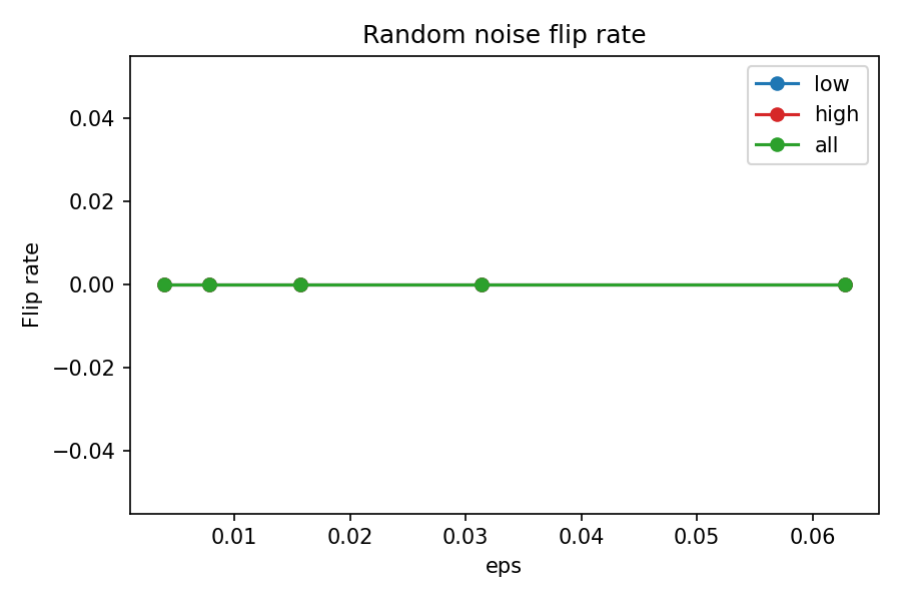}\hfill
\includegraphics[width=0.48\linewidth]{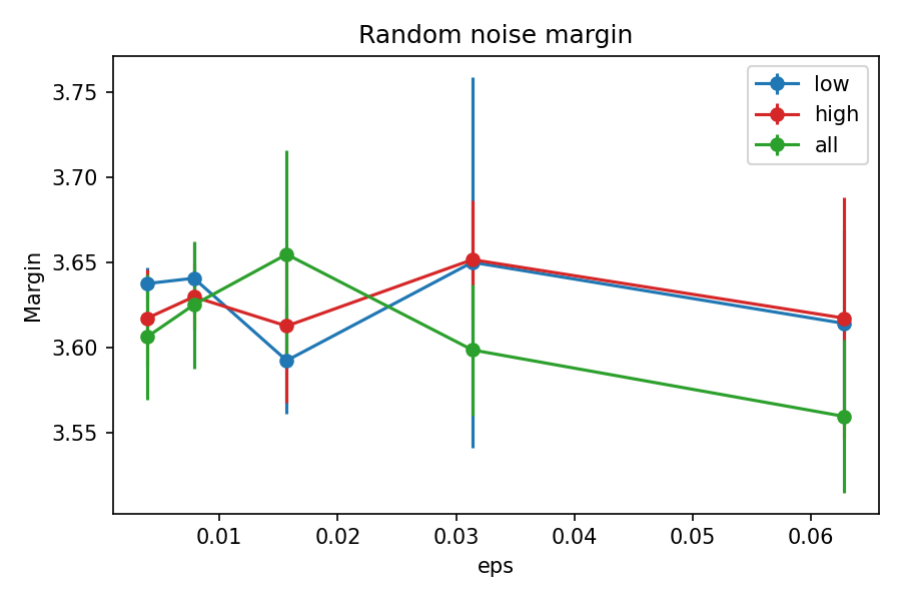}
\vspace{0.5em}
\includegraphics[width=0.48\linewidth]{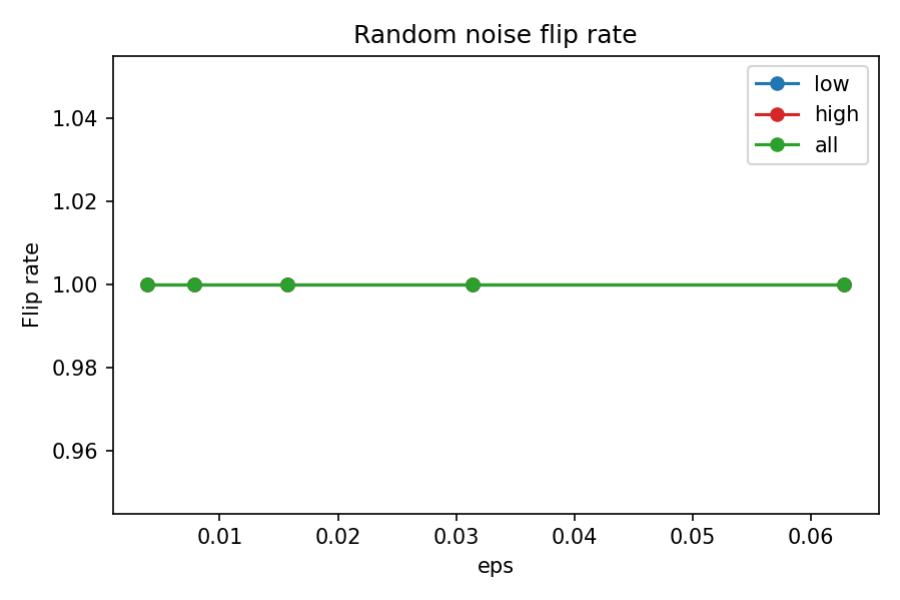}\hfill
\includegraphics[width=0.48\linewidth]{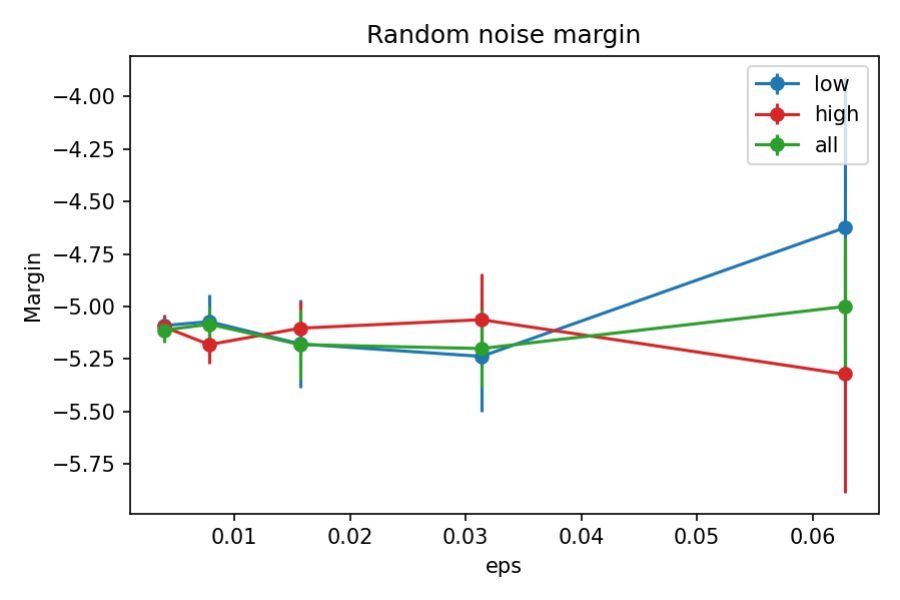}
\caption{
\textbf{Per-sample behaviour under random band-limited noise.}
Each row corresponds to one SEEDBench sample.
\textbf{Left column:} flip rate as a function of noise magnitude $\epsilon$.
\textbf{Right column:} corresponding margin evolution (base option minus strongest competitor).
Across both samples, margins degrade steadily with increasing $\epsilon$ even when predictions do not immediately flip,
revealing hidden representational instability that precedes output-level failures.
}
\label{fig:freq_rand_noise_margin}
\end{figure*}

Figure~\ref{fig:freq_rand_noise_margin} shows per-sample flip behaviour and corresponding margin evolution.
Even when predictions do not flip, increasing $\epsilon$ consistently erodes classification margins,
revealing representational instability that precedes output-level failure.





\subsection{Frequency Ablation (Reliance Test)}
\label{subsec:freq_ablation}

To assess frequency reliance directly, we perform controlled frequency ablations.
For a cutoff $c \in (0,1)$, we define a radial Fourier mask $M_c$ and construct:
\begin{itemize}
    \item \textbf{Low-pass keep:} retain frequencies below $c$ (remove high frequencies),
    \item \textbf{High-pass keep:} retain frequencies above $c$ (remove low frequencies).
\end{itemize}

For each image, we record the smallest cutoff $c$ at which the predicted answer flips, yielding per-sample flip thresholds.

Figure~\ref{fig:freq_ablation_thresholds} shows threshold distributions for low-pass and high-pass ablations.
Low-pass ablation induces flips at smaller cutoffs, indicating sensitivity to high-frequency removal.
However, high-pass ablation also produces failures across a broad range of cutoffs, demonstrating that low-frequency content alone is insufficient for robust decision-making.

\begin{figure*}[t]
\centering
\includegraphics[width=0.48\linewidth]{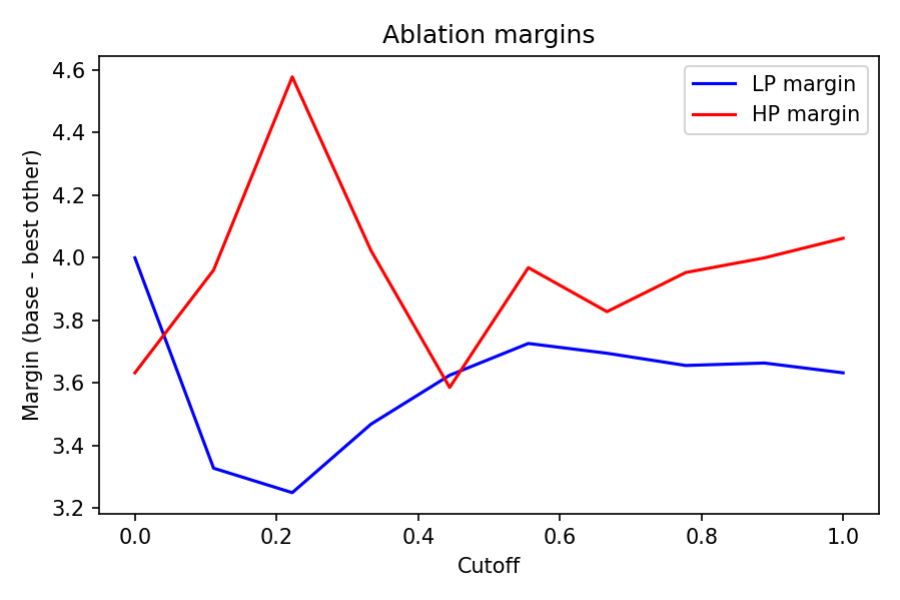}\hfill
\includegraphics[width=0.48\linewidth]{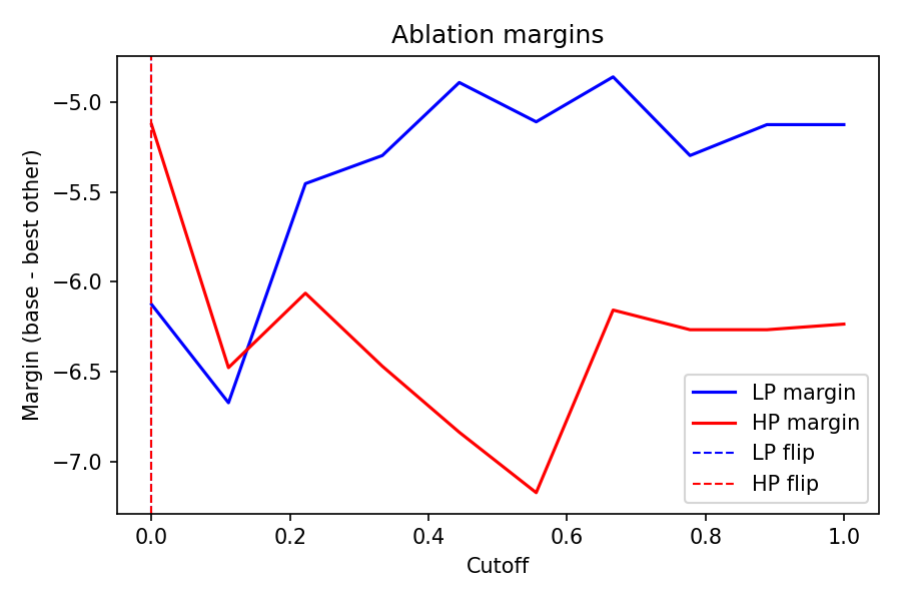}
\caption{
Frequency ablation margins for two representative samples. For each sample, we report the classification margin (base option minus strongest alternative) as a function of the frequency cutoff under low-pass keep (blue) and high-pass keep (red). In both cases, margins degrade substantially before any prediction flip occurs, revealing confidence erosion and representational drift induced by frequency removal. }
\label{fig:freq_ablation_margin}
\end{figure*}

Figure~\ref{fig:freq_ablation_margin} visualises margin evolution for representative samples.
Across both ablation types, margins degrade smoothly well before label flips occur,
indicating gradual confidence erosion rather than abrupt single-band failure.



\subsection{Frequency-Constrained Adversarial Attacks}
\label{subsec:freq_pgd}

We extend the analysis to adversarial perturbations using projected gradient descent (PGD) under an $\ell_\infty$ constraint.
At each iteration, perturbations are projected into a target frequency band:
\begin{equation}
\begin{split}
\delta_{t+1} = \Pi_{\|\delta\|_\infty \le \epsilon} \Big( & \mathrm{Proj}_{\text{band}} \big(\delta_t + \\
& \alpha \cdot \mathrm{sign}(\nabla_\delta \mathcal{L}(x+\delta_t))\big) \Big)
\end{split}
\end{equation}
following~\citep{madry2018towards}.

Table~\ref{tab:freq_pgd} reports attack success rates and prediction flip rates for pixel-space,
low-frequency, and high-frequency PGD attacks.
All three modes achieve high effectiveness, confirming that adversarial vulnerability is not confined to a single frequency band.

\begin{figure*}[t]
\centering
\includegraphics[width=0.32\linewidth]{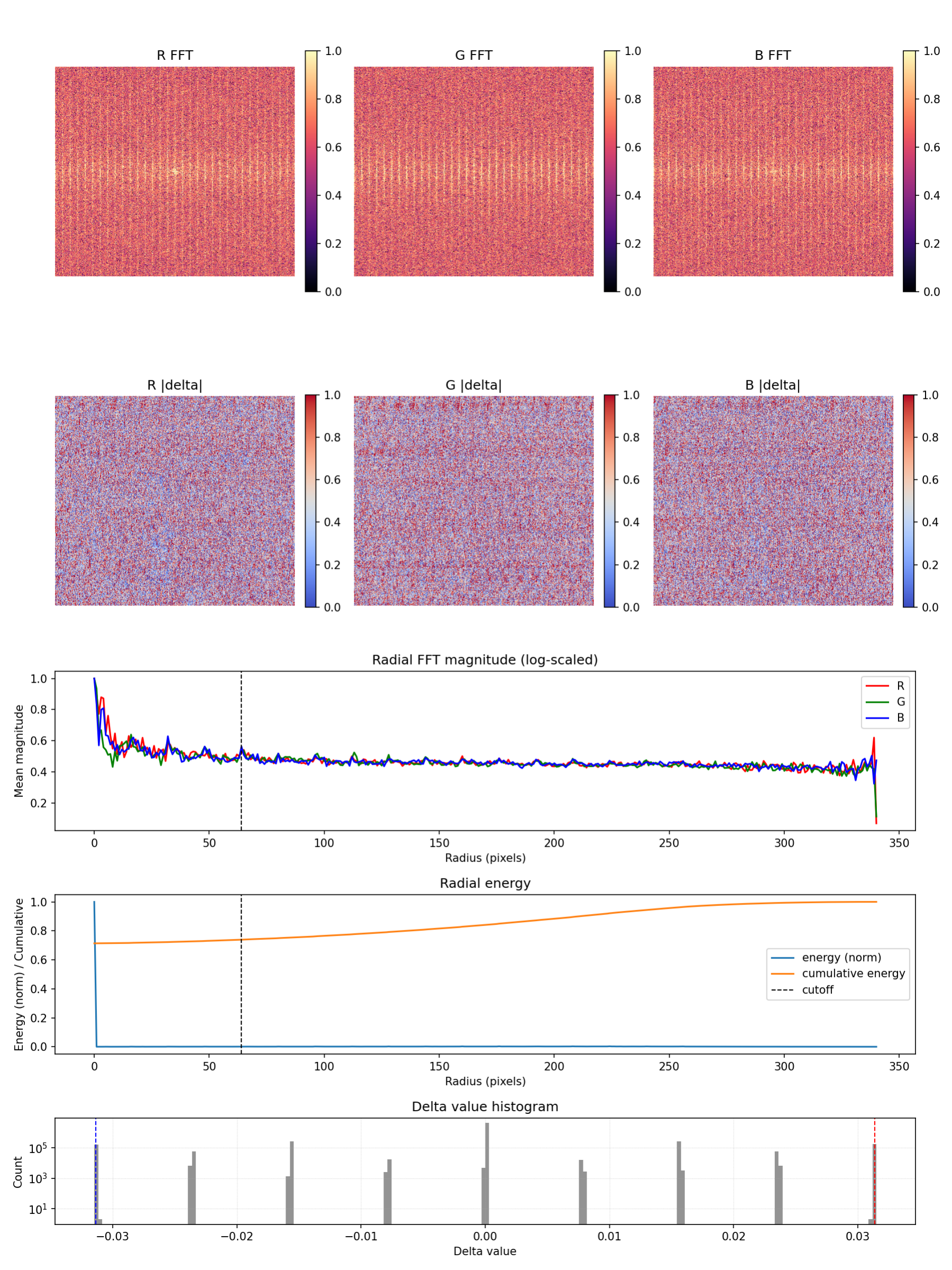}
\includegraphics[width=0.32\linewidth]{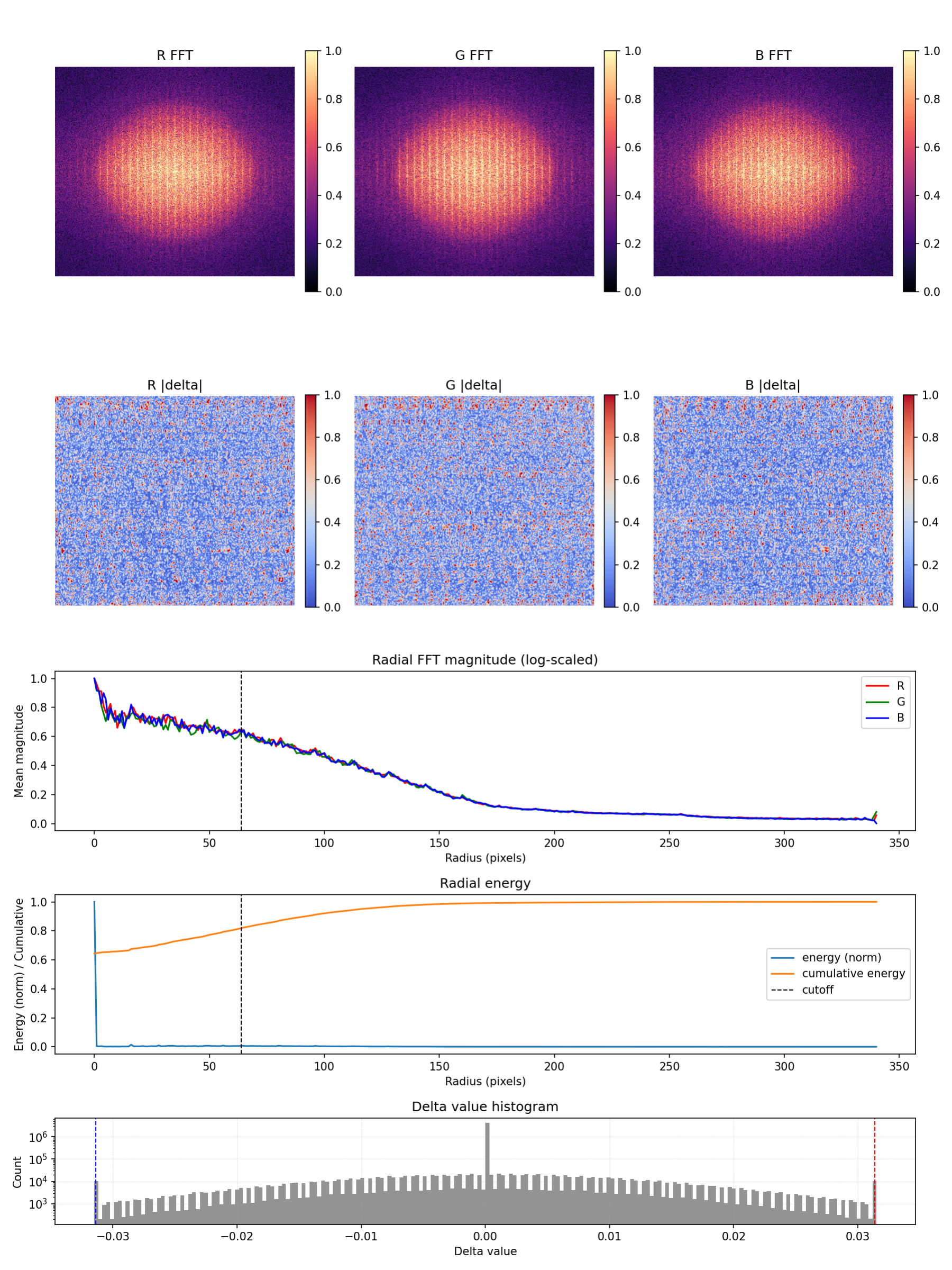}
\includegraphics[width=0.32\linewidth]{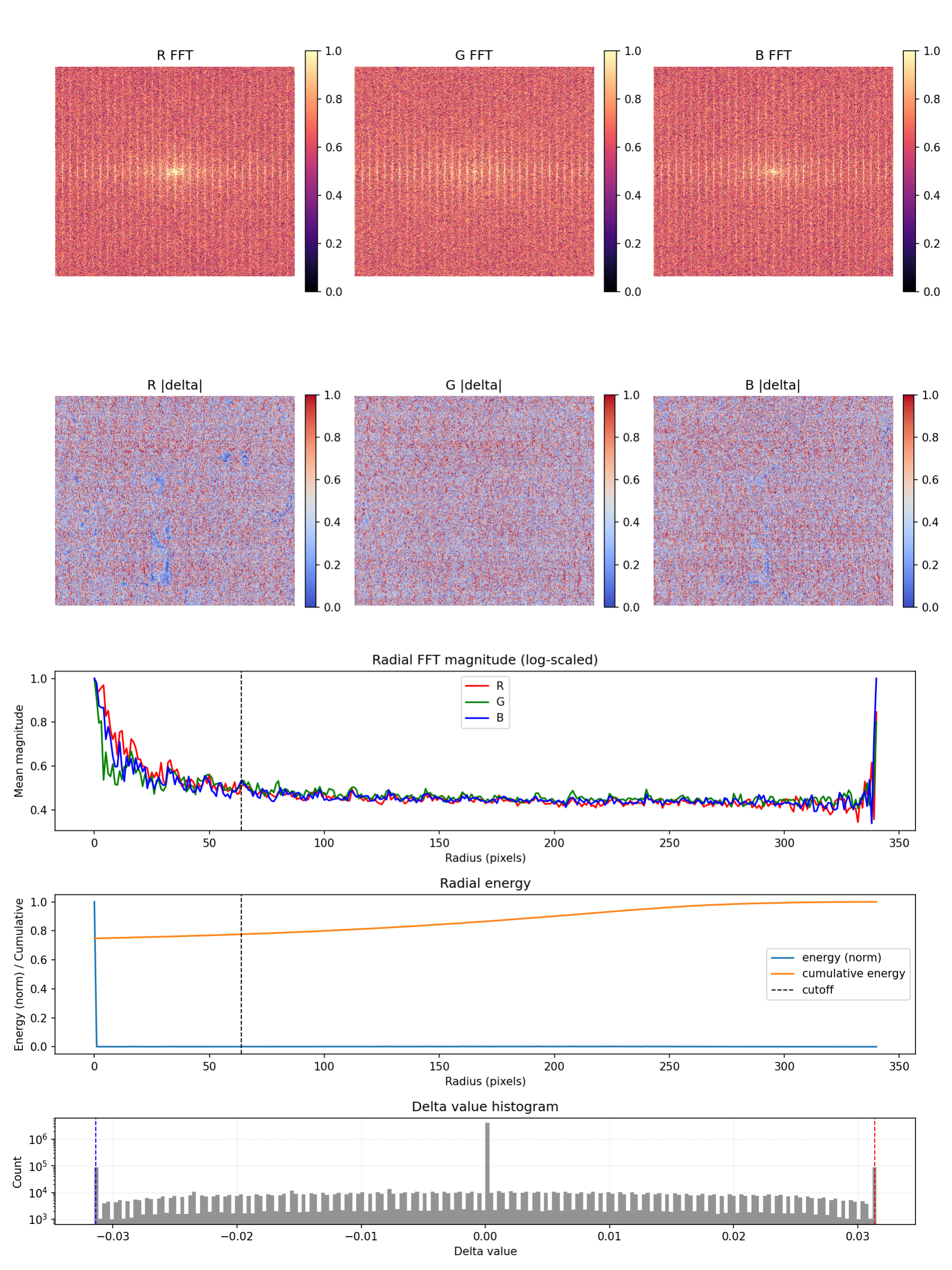}

\caption{
Frequency-domain structure of PGD perturbations.
\textbf{Left}: unconstrained pixel-space PGD.
\textbf{Middle}: low-frequency PGD.
\textbf{Right}: high-frequency PGD.
Radial energy profiles and masks confirm effective frequency isolation.
}
\label{fig:freq_pgd_fft}
\end{figure*}

Figure~\ref{fig:freq_pgd_fft} visualises the frequency-domain structure of the resulting perturbations.
Radial energy profiles confirm that frequency-constrained PGD successfully isolates distinct spectral bands while remaining effective.

\subsection{Natural Perturbations as Induced Spectral Drift}
\label{subsec:spectral_drift}

Taken together, these controlled experiments support a unifying interpretation:
many perturbations commonly regarded as meaning preserving induce robustness failures not by corrupting semantics,
but by causing \emph{spectral and phase drift}.

Translations primarily alter phase and interact with patch discretization;
crop and scale operations induce resampling and aliasing;
text overlays inject broadband, high-contrast edges.
The frequency-aware suite reproduces the same failure patterns observed under natural perturbations,
providing direct evidence that robustness failures cannot be attributed to single-band dominance,
but instead arise from sensitivity across frequency components and phase-induced drift.

\begin{table}[t]
\centering
\small
\begin{tabular}{lcc}
\toprule
Attack mode & ASR $\uparrow$ & PFR $\uparrow$ \\
\midrule
Pixel PGD          & 0.8125 & 0.9525 \\
Low-frequency PGD  & 0.7900 & 0.9550 \\
High-frequency PGD & 0.8225 & 0.9475 \\
\bottomrule
\end{tabular}
\caption{
Dataset-level adversarial robustness under frequency-constrained PGD
($\epsilon=8/255$, 400 samples).
Both frequency bands support effective attacks.
}
\label{tab:freq_pgd}
\end{table}

\end{document}